\documentclass[preprint,5p,times,twocolumn]{cas-dc}

\usepackage[numbers]{natbib}

\usepackage{algorithm}
\usepackage{algorithmic}

\usepackage{newfloat}
\usepackage{listings}

\usepackage{makecell}
\usepackage{colortbl}
\usepackage{multirow}
\usepackage{bbding}
\usepackage{bm}
\usepackage[caption=false]{subfig}

\usepackage{amsmath}
\usepackage{amssymb}
\definecolor{mygray}{gray}{.9}
\definecolor{myblue}{RGB}{93,80,180}
\definecolor{mygreen}{RGB}{93,173,85}

\def\tsc#1{\csdef{#1}{\textsc{\lowercase{#1}}\xspace}}
\tsc{WGM}
\tsc{QE}

\begin{document}
\let\WriteBookmarks\relax
\def\floatpagepagefraction{1}
\def\textpagefraction{.001}

\shorttitle{CD-Lamba: Boosting Remote Sensing Change Detection via a Cross-Temporal Locally Adaptive State Space Model}    

\shortauthors{Z.Wu et al.}  

\title [mode = title]{CD-Lamba: Boosting Remote Sensing Change Detection via a Cross-Temporal Locally Adaptive State Space Model}  



%

\author[1]{Zhenkai Wu}[type=editor,orcid=0009-0000-0613-0584]
\ead{zkwu@zju.edu.cn}
\ead[url]{https://scholar.google.com/citations?user=_xZs8PkAAAAJ&hl=zh-CN}
\fnmark[1]
\author[1, 2]{Xiaowen Ma}
\ead{xwma@zju.edu.cn}
\fnmark[1]
\author[1]{Rongrong Lian}
\ead{lianrr@zju.edu.cn}
\author[3]{Kai Zheng}
\ead{zhengkai1990@zju.edu.cn}
\author[3]{Mengting Ma}
\ead{mtma@zju.edu.cn}
\author[1]{Wei Zhang}
\ead{cstzhangwei@zju.edu.cn}
\cormark[1]
\author[4]{Siyang Song}
\ead{s.song@exeter.ac.uk}
\cormark[1]

\cortext[1]{Corresponding author}

\fntext[1]{Contributed equally}

\affiliation[1]{organization={School of Software Technology, Zhejiang University},
            addressline={}, 
            city={Hangzhou},
            postcode={310027}, 
            country={China}}
\affiliation[2]{organization={Noah's Ark Lab, Huawei},
            addressline={}, 
            city={Shanghai},
            postcode={201206}, 
            country={China}}
\affiliation[3]{organization={School of Computer Science and Technology, Zhejiang University},
            addressline={}, 
            city={Hangzhou},
            postcode={310027}, 
            country={China}}
\affiliation[4]{organization={HBUG Lab, University of Exeter},
            addressline={}, 
            country={UK}}





\begin{abstract}
Mamba, with its advantages of global perception and linear complexity, has been widely applied to identify changes of the target regions within the remote sensing (RS) images captured under complex scenarios and varied conditions. However, existing remote sensing change detection (RSCD) approaches based on Mamba frequently struggle to effectively perceive the inherent locality of change regions as they direct flatten and scan RS images (i.e., the features of the same region of changes are not distributed continuously within the sequence but are mixed with features from other regions throughout the sequence). In this paper, we propose a novel locally adaptive SSM-based approach, termed CD-Lamba, which effectively enhances the locality of change detection while maintaining global perception. Specifically, our CD-Lamba includes a Locally Adaptive State-Space Scan (LASS) strategy for locality enhancement, a Cross-Temporal State-Space Scan (CTSS) strategy for bi-temporal feature fusion, and a Window Shifting and Perception (WSP) mechanism to enhance interactions across segmented windows.
These strategies are integrated into a multi-scale Cross-Temporal Locally Adaptive State-Space Scan (CT-LASS) module to effectively highlight changes and refine changes' representations feature generation. CD-Lamba significantly enhances local-global spatio-temporal interactions in bi-temporal images, offering improved performance in RSCD tasks.
Extensive experimental results show that CD-Lamba achieves state-of-the-art performance on four benchmark datasets with a satisfactory efficiency-accuracy trade-off. Our code is publicly available at \href{https://github.com/xwmaxwma/rschange}{https://github.com/xwmaxwma/rschange}.
\end{abstract}


\begin{keywords}
Remote Sensing Change Detection \sep State Space Model \sep Adaptive Locality \sep Spatio-temporal Context
\end{keywords}

\maketitle

\section{Introduction}

Remote sensing (RS) change detection (CD) aims to compare two or more images describing the same geographical area but captured at different time stamps. It quantitatively and qualitatively assesses changes in geographical entities and environmental factors of interest \cite{singh1989review}. This technique is crucial for detecting and understanding surface activities and changes, which plays a significant role in various real-world applications such as urban expansion \cite{marin2014building}, deforestation \cite{de2020change}, land use \cite{hu2018automatic}, and damage assessment \cite{mahdavi2019polsar}.

Existing RSCD models aim to effectively aggregate semantic changes from bi-temporal image features~\cite{base_tr1, noman2024remote, rs16050804}, while suppressing non-interest changes caused by external factors (e.g., weather, lighting conditions, seasonal variations, and frequent irrelevant changes induced by human activities), and accurately capturing target changes. Although certain efforts have centered on designing effective bi-temporal feature fusion mechanisms to enhance spatio-temporal context interaction and perception, significant advancements also have been made by refining convolutional neural networks (CNNs) \cite{ifnet, snunet, dtcdscn, USSFCNet, AFCF3DNet} and transformers \cite{BIT, changeformer, SARASNet} to yield enhanced CD results. However, accurately modeling change-related spatio-temporal contexts while maintaining low computational complexity are critical challenges, especially for processing high-resolution RS images \cite{BIT,changeformer,  shi2020change, chen2020dasnet, stanet}. Specifically, CNN-based RSCD models typically suffer from limited contextual receptive fields due to the locality of convolution operations. Despite the attempts made to overcome this with deeper architectures, dilated convolutions, and attention mechanisms, they still struggle to capture dense long-range relationships \cite{de2020change, chen2020dasnet, zhang2020feature}. On the other hand, Transformer-based models, though capable of modeling global context through self-attention, face significant challenges in terms of computational complexity. This is primarily due to the quadratic computational cost of self-attention, which restricts their practical applicability \cite{noman2024remote, BIT}.

\begin{figure*}[h]
  \centering
  \includegraphics[width=\linewidth]{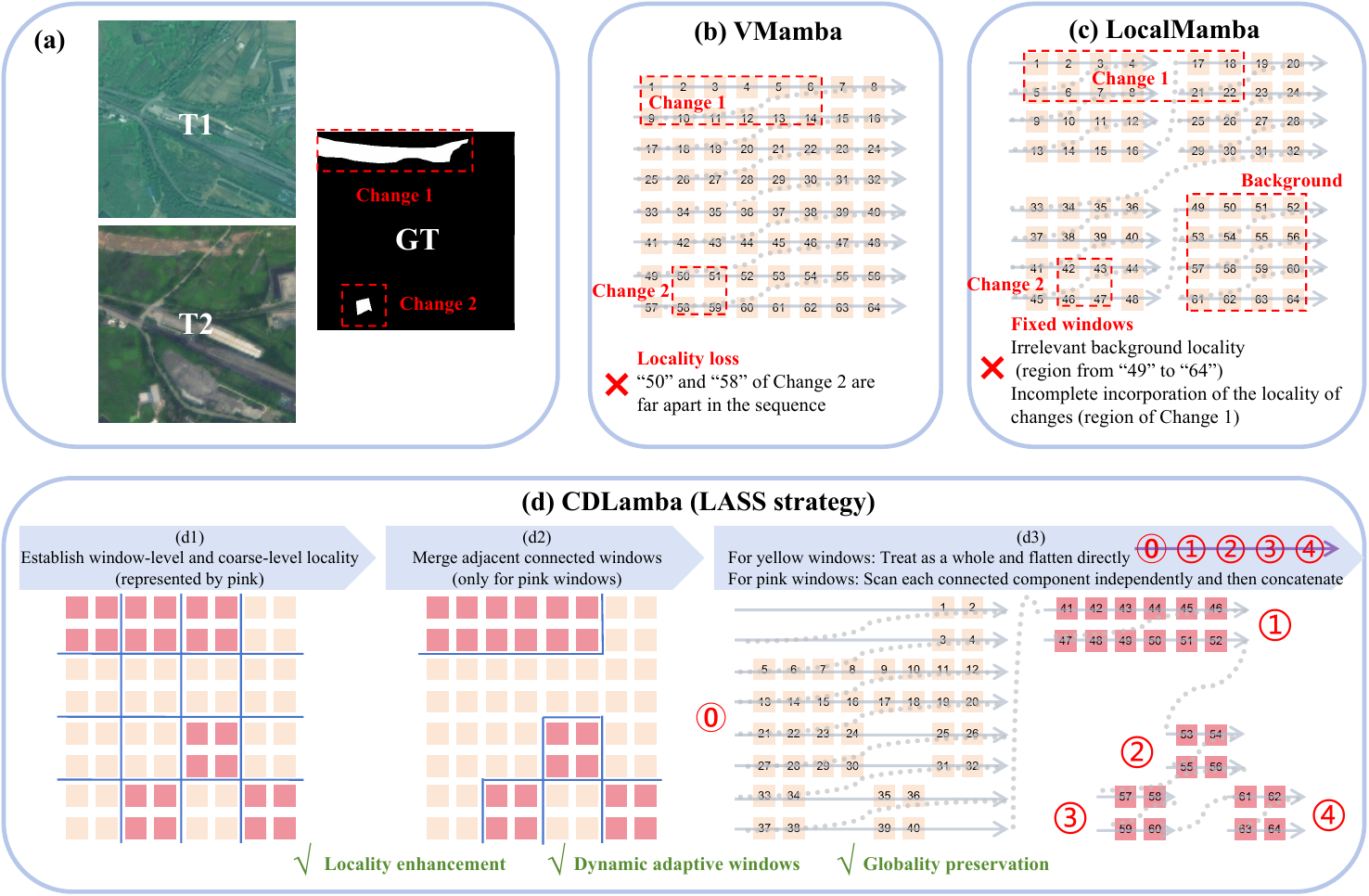}
  \caption{Comparisons of State-Space Scan (SS) strategies among VMamba \cite{liu2024vmamba}, LocalMamba \cite{huang2024localmamba} and CD-Lamba.
\textbf{(a)} An example of a pair of geographically co-registered remote sensing images along with the groundtruth of the change regions.
\textbf{(b)} The SS strategy in VMamba leads to locality loss, reducing the model's ability to capture local details.
\textbf{(c)} The SS strategy in LocalMamba is constrained by fixed windows, limiting its adaptability.
\textbf{(d)} The step-by-step process of our proposed Locally Adaptive State-Space Scan (LASS) strategy. Pink windows are selected based on the top-$k$ score map. The sequence is formed by first flattening all yellow pixels and then sequentially appending the flattened connected regions from the pink windows. 
In terms of reducing the loss of locality, for example, when comparing LocalMamba with VMamba, the gap between the two rows of features in Change 2 has been reduced from the original 6 pixels to just 2 pixels. Furthermore, when comparing our CD-Mamba with LocalMamba, this gap is reduced to 0 pixels.
}
  \label{fig:intro}
\end{figure*}

Inspired by the promising potential of state space models (SSMs) \cite{gu2021efficiently}, particularly the key model Mamba \cite{gu2023mamba}, for handling the challenge of balancing global awareness and linear computational complexity in a variety of computer vision tasks \cite{zhu2024vision, liu2024vmamba, behrouz2024graph, liu2024swin, pei2024efficientvmamba, zhao2024supervised}, a large number of studies have been devoted to adapting and extending SSMs for RSCD tasks. For instance, RSMamba \cite{rsmamba} introduces an omnidirectional scanning strategy tailored for the non-directionality of remote sensing images, while ChangeMamba \cite{changemamba} proposes a temporal cross-scanning strategy to accommodate the shift in RSCD tasks from single-image inputs to bi-temporal inputs. These innovations bring a fresh perspective to RSCD by incorporating global awareness, which is crucial for identifying change regions within a large volume of background features. However, these extensions inherit a significant limitation from VMamba's image-flattening approach \cite{liu2024vmamba}, which causes a loss of locality in regions of change (i.e., in a sequence, adjacent rows of a certain change may be interspersed with features from many other regions, causing the features of the change to become dispersed and disrupting its spatial coherence), as illustrated in Fig. \ref{fig:intro}(b). As a result, it is essential to address this flaw for preserving spatial coherence, and consequently enhance the performance of SSM-based RSCD approaches. As shown in Fig. \ref{fig:intro}(c), a recent LocalMamba \cite{huang2024localmamba} introduces locality by dividing the image into quarters and scanning each section individually. As a result, the features of the same target are distributed more closely within the sequence.  However, this fixed-window strategy inevitably suffer from two crucial issues: 1) it introduces irrelevant background locality, such as the locality of the sequence from "49" to "64", which may mislead the differentiation of change areas; and 2) it results in incomplete incorporation of the locality of changes, such as the Change 1 region being truncated in the middle and split across two separate scanning sequences.

\begin{table*}[t]
    \setlength{\tabcolsep}{8pt}
	\begin{center}
		\caption{
		Comparisons of properties among typical RSCD methods. The proposed CD-Lamba achieves an effective global perception capability by introducing the state space model. In particular, CD-Lamba offers linear computational complexity and low memory consumption, while ensuring superior locality of changes compared to CNN-based, transformer-based, and other SSM-based RSCD methods.}
		\label{table:1}
            \begin{tabular}{c||c||c||ccc}
		\Xhline{1.2pt}
            \rowcolor{mygray}
		     & & &\multicolumn{3}{c}{SSM-based}\\
            \rowcolor{mygray}
			\multicolumn{1}{c||}{\multirow{-2}{*}{Properties}}
		     & \multicolumn{1}{c||}{\multirow{-2}{*}{CNN-based}}  & \multicolumn{1}{c||}{\multirow{-2}{*}{Transformer-based}} & RSMamba \cite{rsmamba} & ChangeMamba \cite{changemamba} & CD-Lamba \\		
                \hline \hline
                Global perception capability & \XSolidBrush &\CheckmarkBold  & \CheckmarkBold & \CheckmarkBold  & \CheckmarkBold  \\
                Computational complexity & $\bm{\mathcal{O}(n)}$ &  
                $\bm{\mathcal{O}(n^2)}$&$\bm{\mathcal{O}(n)}$ &$\bm{\mathcal{O}(n)}$ &$\bm{\mathcal{O}(n)}$   \\
                Low memory consumption & \CheckmarkBold & \XSolidBrush &  \CheckmarkBold  & \CheckmarkBold & \CheckmarkBold   \\
                Locality of changes & \CheckmarkBold  &  \XSolidBrush &   \XSolidBrush & \XSolidBrush  &\CheckmarkBold    \\
            \hline
		\end{tabular}
	\end{center}
\end{table*}

In this paper, we propose a novel CD-Lamba which is the first work that not only inherits the strong local representation capability from the local scanning mechanism of LocalMamba \cite{huang2024localmamba} but also improve the locality enhancement of Mamba for the RSCD task. 
Specifically, our CD-Lamba consists of a innovately proposed multi-scale Cross-Temporal Locally Adaptive State-Space Scan (CT-LASS) module, as well as a Siamese backbone for modulating multi-scale feature generation, and lightweight change detector for deriving the final change mask. As illustrated in Fig. \ref{fig:intro}(d), our novel CT-LASS module is designed to enhance the locality of changes while preserving the global context. This is achieved by introducing a novel Locally Adaptive State-Space Scan (LASS) strategy, which employs dynamic adaptive windows, including the merging of adjacent connected windows, to accommodate varying sizes and shapes of change regions. We also perform Cross-Temporal State-Space Scan (CTSS) strategy to enhance cross-fusion of the bi-temporal locality in a pixel-wise way. To mitigate the loss of continuous perception caused by cropping operations conducted at feature map boundaries, we draw inspiration from the shifted window design of the Swin Transformer \cite{liu2021swin} (illustrated in Fig. \ref{fig:intro}(d1)). We further enhance this concept by implementing a Window Shifting and Perception (WSP) mechanism in the CT-LASS module, enabling connections in all directions rather than isolating each window. Moreover, unlike previous architectures, our CD-Lamba models spatio-temporal context information simultaneously. Each level of the weight-sharing Siamese backbone combines the output from the preceding level with bi-temporal change features refined by the CT-LASS module. These well-regulated change feature representations allow a lightweight change detector to effectively and efficiently generate the change detection map. Our main contributions are:
\begin{itemize}
     \item We propose a novel CD-Lamba which, for the first time, integrates a novel locality adaptive enhancement strategy into SSM-based RSCD. It overcomes regular Mamba's limitations in local perception, as shown in Table \ref{table:1}.
    
    
    \item We design a Locally Adaptive State-Space Scan strategy to enhance bi-temporal locality while preserving global context, complemented by a Cross-Temporal State-Space Scan strategy designed for pixel-wise cross-fusion.
    
    \item We develop a Window Shifting and Perception mechanism to improve interactions across segmented windows, addressing the discontinuity in perception caused by cropping operations at feature map boundaries.

    \item Experiments conducted on four RSCD benchmark datasets reveal that our proposed CD-Lambda model outperforms previous SSM-based approaches, achieving notable improvements in F1-score by 2.43\%, 3.28\%, 5.72\%, and 8.06\% on the WHU-CD, SYSU-CD, DSIFN-CD, and CLCD datasets compared to ChangeMamba \cite{changemamba}, respectively.

\end{itemize}


\section{Related Work}
\subsection{State Space Model}

State Space Models (SSMs) have become pivotal in deep learning for handling long-range dependencies in sequential data \cite{gu2021efficiently, gu2021combining, smith2022simplified}. Traditional models like RNNs \cite{arjovsky2016unitary}, CNNs \cite{bai2018empirical}, and Transformers \cite{katharopoulos2020transformers} struggle with prolonged interrelations. The HiPPO initialization \cite{gu2020hippo} enhanced SSMs' capability to capture extended dependencies, and the Linear State Space Layer (LSSL) \cite{gu2021combining, voelker2019legendre} addressed continuous-time memorization but was limited by computational demands. The S4 model \cite{gu2021efficiently} improved speed and memory efficiency, outperforming LSSL.

Subsequent advancements included complex-diagonal structures \cite{gu2022parameterization, gupta2022diagonal}, multiple-input multiple-output support \cite{smith2022simplified}, and diagonal plus low-rank operations \cite{hasani2022liquid} to enhance generalization across tasks. These strategies have expanded to large representation models \cite{fu2022hungry, ma2022mega, mehta2022long}. SSMs' success in computer vision began with S4ND \cite{nguyen2022s4nd}, modeling 1D, 2D, and 3D visual data but struggled with image adaptation. To address this challenge, a multitude of methods based on the selection mechanism proposed by Mamba \cite{gu2023mamba} have been emerging recently. Vim \cite{zhu2024vision} leverages a bidirectional state space to model the global visual context of dependency data without being biased towards specific images. Meanwhile, Vmamba \cite{liu2024vmamba} pioneers a four-way scanning module to tackle directionality sensitivity issues arising from differences between one-dimensional sequences and multi-channel images.  

However, unlike the one-dimensional sequences in natural language, two-dimensional data like images exhibit strong locality, where each pixel is often closely related to its surrounding pixels. Flattening images into one-dimensional sequences by row or column disrupts this inherent locality, weakening the model's ability to capture fine-grained details. To address this issue, LocalMamba \cite{huang2024localmamba} introduces a Windowed Selective Scan mechanism that effectively captures local dependencies within images while preserving global context. Nevertheless, LocalMamba divides images into four fixed windows, and this rigid partitioning can introduce unnecessary background locality and often causes objects near window boundaries to be cut off, scattering their locality across multiple windows. This limitation highlights the need for a more flexible and adaptive local window scanning strategy to better preserve and capture detailed locality.

\subsection{Remote Sensing Change Detection}

Traditional RSCD methods can be divided into algebraic-based \cite{singh1986change, todd1977urban, sun2022structured}, transformation-based \cite{celik2009unsupervised, saha2019unsupervised, crist1985tm}, and classification-based \cite{suthaharan2016support, seo2018fusion} methods. However, the effectiveness of these methods often depends heavily on empirically designed and handcrafted feature qualities.

CNN-based RSCD methods can be classified into three categories based on the fusion stages of bi-temporal information. Early-term fusion methods \cite{de2020change, lebedev2018change, peng2019end} first concatenate the bi-temporal images into a single input, which is then directly connected to the semantic segmentation network. Mid-term fusion methods \cite{zhang2022swinsunet, liu2021building, li2022remote, zheng2021change, song2022remote, chen2021remote} combine bi-temporal features extracted from neural networks and generate change maps based on the fused features. Late-term fusion methods \cite{liu2019temporal, nemoto2017building} first classify the bi-temporal images separately and then compare their classification results to obtain change areas. To increase the receptive field size, existing methods include using deeper CNN models \cite{de2020change, chen2020dasnet, zhang2020feature}, employing dilated convolutions \cite{zhang2018triplet}, and applying attention mechanisms \cite{liu2020building, zhang2020deeply, peng2020optical}. However, most of them still struggle to adequately simulate dense long-range relationships between image features \cite{liu2020building,zhang2020deeply, peng2020optical}.

Transformer-based RSCD methods uses self-attention for global feature dependency modeling. ChangeFormer \cite{changeformer} applies Transformer directly to RSCD tasks but faces high computational complexity. Mainstream methods combine CNN and Transformer to balance local and global information modeling with efficiency. BIT \cite{BIT} uses a Transformer encoder to model context from semantic labels derived from convolutional features. ICIF-Net \cite{ICIF} leverages intra-scale cross interaction and inter-scale feature fusion for CNN and Transformer integration. SARASNet \cite{SARASNet} focuses on interaction between bi-temporal images, using relation perception, scale perception, and cross transformation for better scene change detection. However, these methods often amplify discrepancies in bi-temporal features with non-lightweight detectors, reducing model efficacy, especially for high-resolution RS images.

SSM-based RSCD methods have improved upon the SS2D strategy, i.e., the 2D-selective-scan strategy, originally introduced in VMamba \cite{liu2024vmamba}, by refining the scanning mechanism to better align with the unique characteristics of bi-temporal remote sensing images. Specifically, RSMamba \cite{rsmamba} introduces an omnidirectional scanning strategy designed to address the non-directional nature of remote sensing data. Meanwhile, ChangeMamba \cite{changemamba} adopts a temporal cross-scanning strategy to handle the transition of RSCD tasks from single-image inputs to bi-temporal inputs, effectively incorporating the time dimension. These advancements offer a new perspective on RSCD by integrating global context, which is essential for accurately identifying change regions amidst extensive background features. However, both methods share a significant limitation: their scanning strategies directly flatten images using VMamba, leading to the loss of locality in critical change regions.

\begin{figure*}[h]
  \centering
  \includegraphics[width=\linewidth]{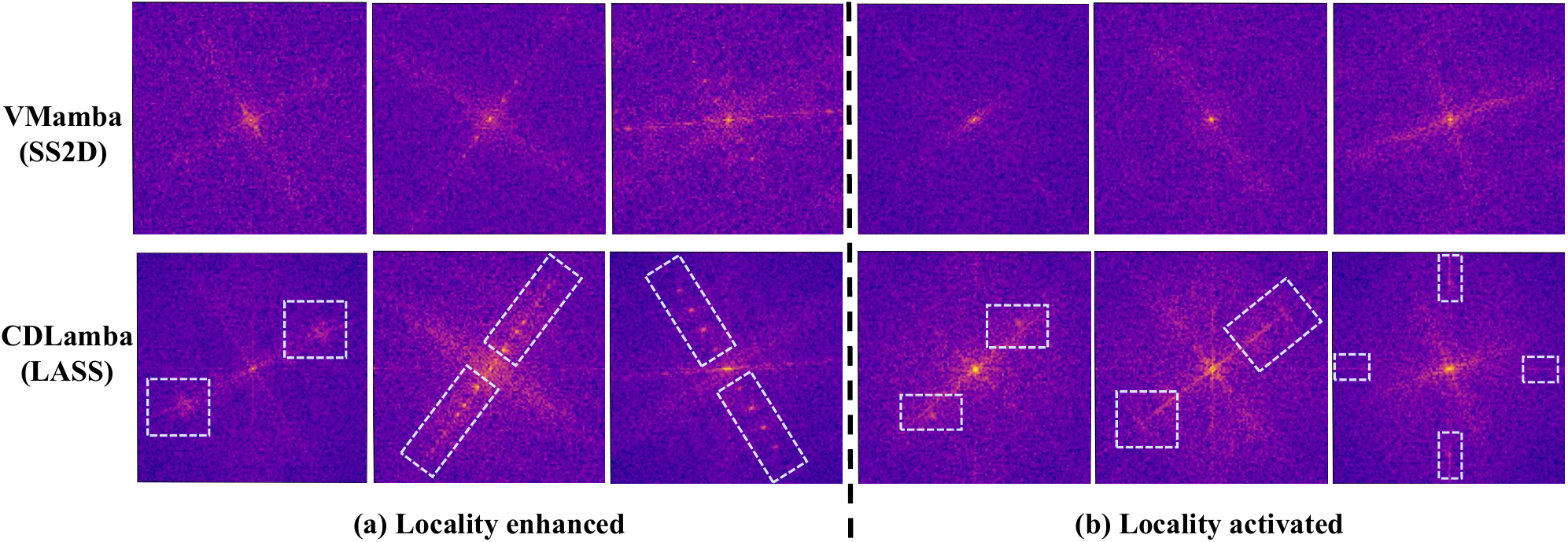}
  \caption{Spectral analysis between SS2D in VMamba \cite{liu2024vmamba} and our proposed LASS in CD-Lamba. Low-frequency global features are closer to the center, while high-frequency local features are farther from the center. The brighter the pixel, the greater the energy. Therefore, the figure demonstrates that LASS not only \textbf{(a) enhances the locality} recognized by SS2D but also \textbf{(b) activates the locality} that SS2D cannot recognize.}
  \label{fig:pinpu}
\end{figure*}

\begin{figure*}[h]
  \centering
  \includegraphics[width=\linewidth]{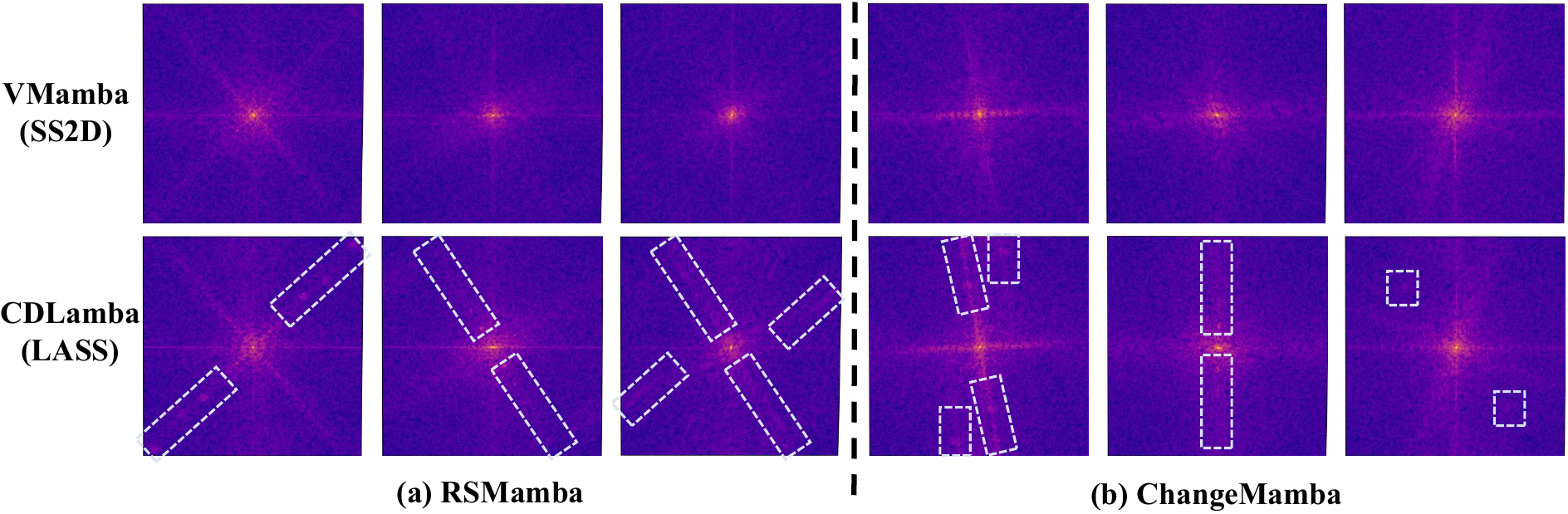}
  \caption{More spectral analysis of typical SSM-based RSCD methods (e.g., RSMamba \cite{rsmamba} and ChangeMamba \cite{changemamba}). The comparison focuses on replacing their selective scan strategies with either the SS2D approach in VMamba \cite{liu2024vmamba} or our proposed LASS in CD-Lamba. Low-frequency global features are closer to the center, while high-frequency local features are farther from the center. The brighter the pixel, the greater the energy. Therefore, the figure demonstrates that LASS not only enhances the locality recognized by SS2D but also activates the locality that SS2D cannot recognize.}
  \label{fig:pinpu1}
\end{figure*}

\section{Preliminaries and Findings}

To facilitate the understanding of the working mechanism of our proposed CD-Lamba, we first introduce the preliminary concepts related to state space models (SSMs), including their continuous and discrete formulations, as well as the efficient computation of SSMs based on the selective scanning mechanism. Additionally, we provide our findings on the effectiveness of Mamba's scanning strategy for extracting RSCD-related cues from RS images.

\subsection{Preliminaries}

\begin{figure*}[h]
  \centering
  \includegraphics[width=\linewidth]{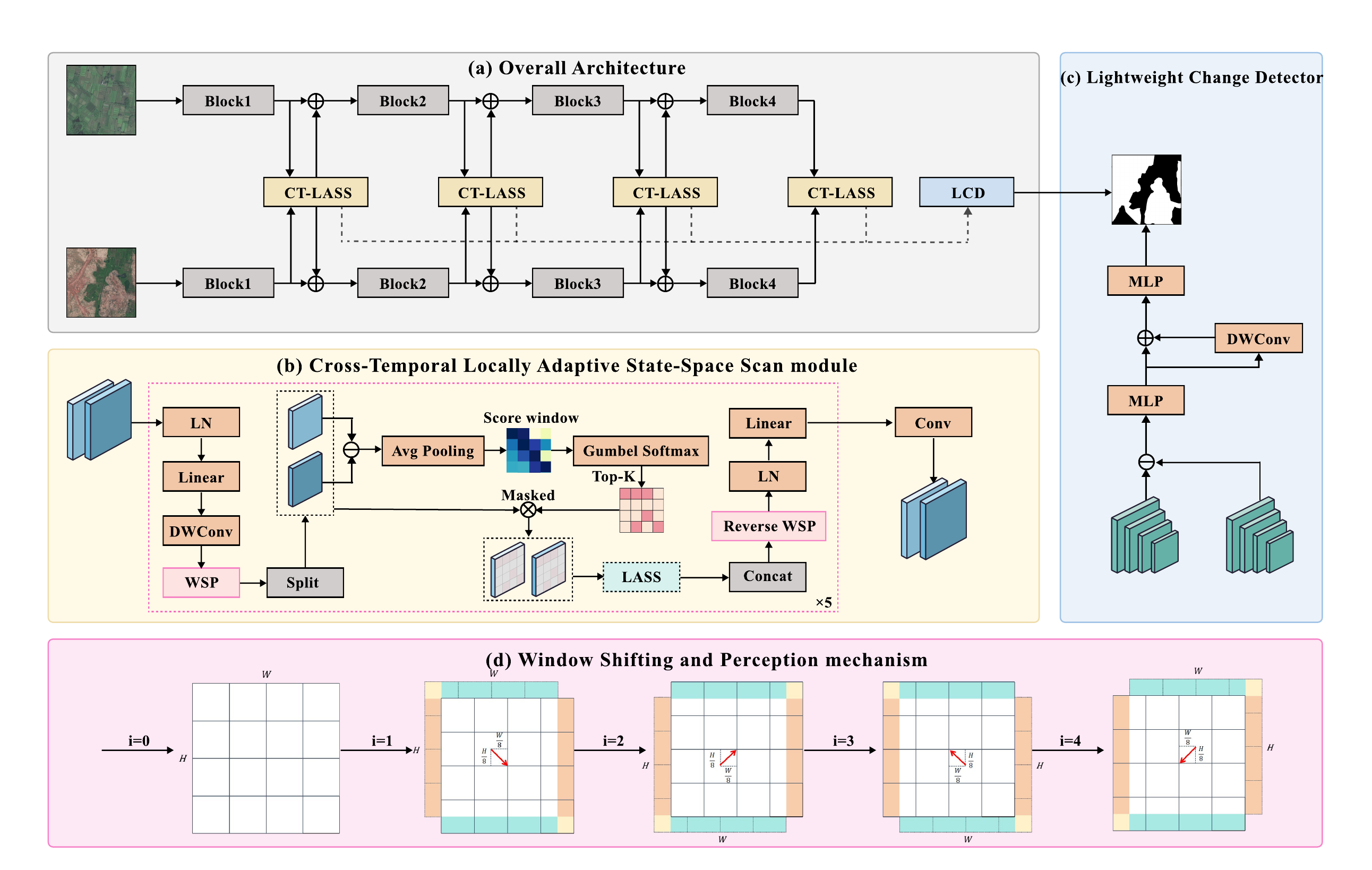}
  \caption{\textbf{(a)} The overall architecture of CD-Lambda consists of a Siamese backbone, \textbf{(b)} a Cross-Temporal Locally Adaptive State-Space Scan (CT-LASS) module, and \textbf{(c)} a lightweight change detector (LCD). In the CT-LASS module, we propose a novel Locally Adaptive State-Space Scan (LASS) strategy, where the score window is generated by average pooling the differences of bi-temporal features. The CT-LASS module efficiently models the global-local spatio-temporal context, enabling the modulation of feature generation based on bi-temporal feature flows and selectively enhancing changes of interest. \textbf{(d)} Additionally, we introduce a Window Shifting and Perception (WSP) mechanism, performing CT-LASS five times at each scale with shifts of 1/8 length in four directions, ensuring sufficient interaction between bi-temporal features.}
  \label{fig:net}
\end{figure*}

\subsubsection{State Space Models (S4)} State space models (SSMs) are deep learning models that have been widely-used in sequencial data analysis \cite{gu2021efficiently}.These models originate from control systems, where they are designed to describe the state representation of sequences at each time step and predict the next state based on the input. Specifically, they transform an input sequence 
\( x(t) \in \mathbb{R}^{L \times D} \) 
into an output sequence 
\( y(t) \in \mathbb{R}^{L \times D} \) 
by leveraging a learnable potential state 
\( h(t) \in \mathbb{R}^{N \times D} \). 
The process can be mathematically expressed as follows:

\begin{equation}
\label{eq1}
\left\{ \begin{array}{l}
	h'(t) = \bm{A} h(t)+\bm{B} x(t),\\
	y(t) = \bm{C}h(t) + \bm{D} x(t),\\
\end{array} \right. 
\end{equation}
where $\bm{A}\in \mathbb{R}^{N \times N}$, $\bm{B}, \bm{C}\in \mathbb{R}^{N \times D}$ denote the learnable parameters, $\bm{D}$ denotes the bias, and $N$ is the state size. 

\subsubsection{Discretization} The original SSMs are belong to the continuous time system family, as deep learning models typically process discrete data (e.g., matrices and vectors). Therefore, discretization is necessary to align the model with the sampling frequency of the input signal, thereby enhancing computational efficiency in computer systems \cite{gu2021combining}. Following the previous work \cite{gupta2022diagonal}, given the sampling time scale $\bm{\Delta}$, the above continuous system is discretized through zero-order hold rule, thus converting the continuous time parameters ($\bm{A}$, $\bm{B}$) to their discrete counterparts ($\overline{\bm{A}}$, $\overline{\bm{B}}$):
\begin{equation}
\left\{ \begin{array}{l}
    \overline{\bm{A}} = e^{\bm{\Delta}\bm{A}},\\
    \overline{\bm{B}} = (\bm{\Delta}\bm{A})^{-1}(e^{\bm{\Delta}\bm{A}}-\bm{I})\cdot \bm{\Delta}\bm{B},\\
    h_0=\overline{\bm{B}} x_0,
\end{array} \right. 
\end{equation}

For simplicity, we omit the constant bias $\bm{D}$. The discretized formulation of $\bm{A}$ and $\bm{B}$ in Eq. \eqref{eq1} can be expanded as:
\begin{equation}
\label{eq3}
\left\{ \begin{array}{l}
    h_t = \overline{\bm{A}} h_{t-1}+\overline{\bm{B}} x_t,\\
    y_t = \bm{C}h_t,
\end{array} \right. 
\end{equation}
where $\overline{\bm{A}}\in \mathbb{R}^{N \times N}$, $\overline{\bm{B}}\in \mathbb{R}^{N \times D}$. To improve the computational efficiency, the iterative process of Eq. \eqref{eq3} can be performed by the parallel computing mode of global convolution \cite{gu2023mamba} as:
\begin{equation}
\begin{aligned}
    y &= x \circledast \overline{\bm{K}},   \\
   \text{with} \quad  \overline{\bm{K}} &= (\bm{C}\overline{\bm{B}},\bm{C}\overline{\bm{A}\bm{B}},\cdots, \bm{C}\overline{\bm{A}}^{L-1}\overline{\bm{B}}),
\end{aligned}
\end{equation}
where $\circledast$ denotes the convolution operation, and $\overline{\bm{K}}\in \mathbb{R}^{L} have $ serves as the kernel of the SSMs.

\subsubsection{Selective State Space Models (S6)} Conventional SSMs (i.e., S4) have been implemented to capture sequence context under linear time complexity, despite the fact that they are constrained by static parameterization and cannot perform content-based reasoning \cite{gu2023mamba}. To address this problem, selective spatial state models (i.e., S6 or Mamba \cite{gu2023mamba}) have been proposed, which enable matrices ($\bm{A}$, $\bm{B}$, and $\bm{C}$) to vary dynamically with the input, making the process data-driven. This allows the model to selectively propagate or forget information along the sequence length based on the characteristics of the current token. In S6, the parameters $\bm{B}$, $\bm{C}$, and $\Delta$ are computed directly from the input sequence $x(t)$, thus enabling sequence-aware parameterization.

\subsection{Findings}
\label{finding}

As shown in Fig. \ref{fig:intro}(b), previous SSM-based RSCD approaches primarily employ the SS2D scanning strategy from VMamba, which flattens the feature map row by row or column by column for scanning. Conceptually, it is straightforward to assume that this strategy inevitably compromises the inherent locality of the image. To validate our hypothesis, as shown in Fig. \ref{fig:pinpu}, we replace the LASS in our CD-Lamba with SS2D and compare the information differences in the resulting output feature maps. To ensure a more rigorous and comprehensive analysis, we also conduct similar comparative experiments on RSMamba and ChangeMamba, as illustrated in Fig. \ref{fig:pinpu1}.

To validate the locality enhancement capability of the LASS strategy, we perform a spectral analysis, as spectrograms provide an intuitive way to observe the distribution of global and local information in an image. Specifically, frequencies closer to the center represent higher frequencies (fine-grained details), while those farther from the center indicate lower frequencies (coarse-grained structures). As shown in Fig. \ref{fig:pinpu} and Fig. \ref{fig:pinpu1}, we plot the spectrogram of the shallowest feature map output by the SSM-based module in these models, with a size of $(\frac{H}{4}, \frac{W}{4})$. This feature map contains the richest locality, making it easier to observe how the model perceives the inherent locality of the image. From Fig. \ref{fig:pinpu} and Fig. \ref{fig:pinpu1}, it is evident that our LASS strategy not only enhances the local features already identified by SS2D but also activates local features that SS2D fails to recognize, effectively improving the model's ability to perceive the inherent locality of images.

\section{Methodology}

In this section, we describe in detail how Mamba can be modified to capture RS images' inherent locality while maintaining its global awareness, making it suitable for the RSCD tasks. Specifically, Sec. \ref{oa} first presents the overall architecture of CD-Lamba, and then we elaborate on its specific components, including the Siamese Backbone (Sec. \ref{backbone}), the proposed state-space scan strategies (Sec. \ref{strategy}), the Cross-Temporal Locally Adaptive State-Space Scan (CT-LASS) module (Sec. \ref{ctlass}), and the Lightweight Change Detector (LCD) (Sec. \ref{lcd}). Finally, Sec. \ref{4.4} describes the loss functions used for training our CD-Lamba.

\begin{figure*}[h]
  \centering
  \includegraphics[width=\linewidth]{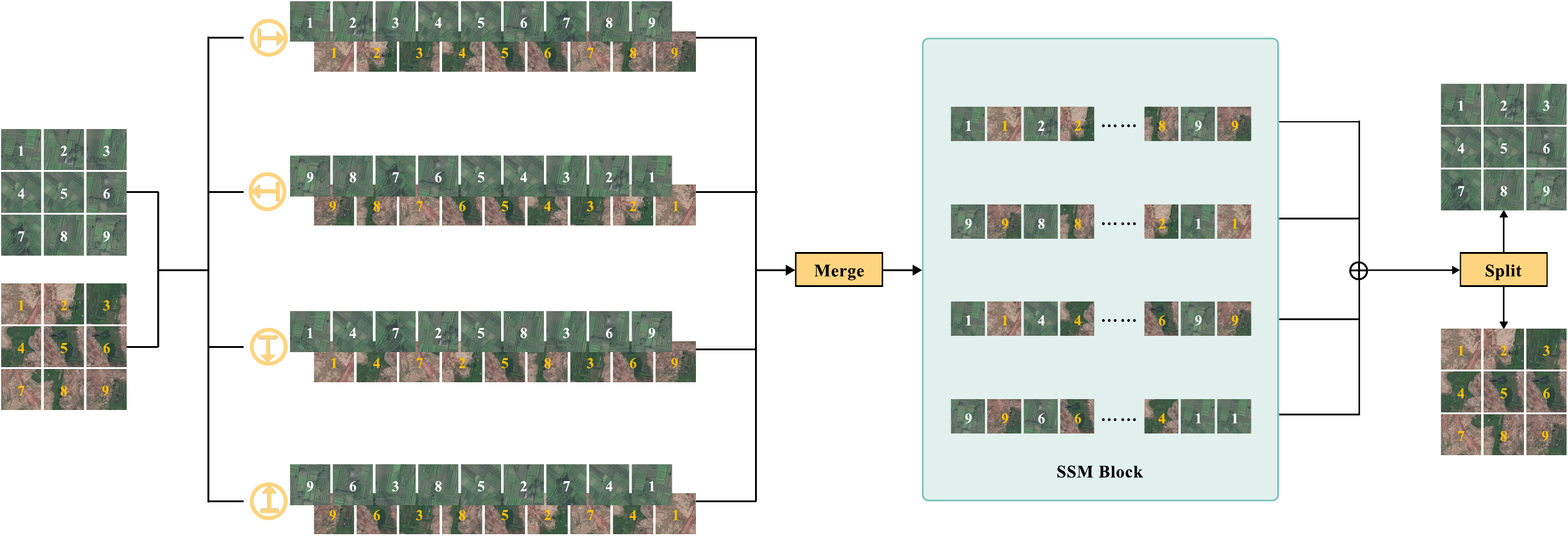}
  \caption{Illustration of Cross-Temporal State-Space Scan (CTSS) strategy with bi-temporal selective scan strategy. We first unfold the bi-temporal windows into one-dimensional sequences along each of the four directions. For the sequence of bi-temporal features obtained in each direction, we perform cross-scanning for bi-temporal features at the same location. Finally, the output features from the four directions are merged to construct the final feature windows.}
\label{fig:scan}
\end{figure*}

\subsection{Overall Architecture}
\label{oa}

As outlined in Section \ref{finding}, current SSM-based RSCD methods struggle to preserve both local and global characteristics of the RS image. To address this challenge, we propose our novel CD-Lamba, which is specifically designed to enhance the locality perception of the scanning strategy. As illustrated in Fig. \ref{fig:net}. given the input images $\mathcal{T}_1$ and $\mathcal{T}_2$, bi-temporal features $\mathcal{F}_1$ and $\mathcal{F}_2$ are extracted through a pair of weight-shared backbones. Notably, each level of the weight-sharing Siamese backbone integrates the output of the previous level with bi-temporal features modulated by CT-LASS modules. Here, the LASS strategy (explained in Fig. \ref{fig:intro}(d)) serves as the core component of the CT-LASS modules, where the score window, generated based on the difference between the bi-temporal features, enables the effective and simultaneous capture of both local and global contexts. CT-LASS is executed five times at each scale with shifts of 1/8 length in four directions. As a result, after modulation of the bi-temporal features, our LCD module is employed to produce the final change maps, ensuring the efficiency and effectiveness of CD-Lamba.

\subsection{Siamese Backbone}
\label{backbone}

We follow \cite{rsmamba} to utilize a Siamese backbone network for producing multi-scale bi-temporal features, where SeaFormer-L \cite{seaformer} is employed. The bi-temporal feature outputs $\mathcal{F}_1^i$ and $\mathcal{F}_2^i$ at each stage will be modulated by CT-LASS modules, as detailed in Section \ref{ctlass}. Subsequently, the bi-temporal inputs $\varGamma_1^i$ and $\varGamma_2^i$ at each stage are obtained by adding the modulated features to the output of the previous stage. The process is denoted as:
\begin{equation}
    \varGamma _{j}^{i}=\mathcal{F}_{j}^{i-1}+\mathrm{CT\_LASS}\left( \mathcal{F}^{i-1} \right). 
\end{equation}
This strategy enriches the bi-temporal feature with the local-global spatio-temporal context generated by SSMs, which underscores the superiority of SSMs in RSCD tasks.

\setlength{\tabcolsep}{5pt}
\begin{table*}[t]
	\centering
		\caption{
		Comparison of performance for RSCD on WHU-CD and SYSU-CD datasets. Highest scores are in bold. All scores are in percentage. Flops are counted with image size of $256 \times 256 \times 3$.
		}
		\label{table:2}
            \begin{tabular}{l||cc||ccccc||ccccc}
		\Xhline{1.2pt}
            \rowcolor{mygray}
		     & & &\multicolumn{5}{c||}{WHU-CD} &\multicolumn{5}{c}{SYSU-CD}\\
            \rowcolor{mygray}
			\multicolumn{1}{c||}{\multirow{-2}{*}{Method}}
               & \multirow{-2}{*}{Params (M)}& \multirow{-2}{*}{Flops (G)} &F1 &Pre. &Rec. &IoU &OA  &F1 &Pre. &Rec. &IoU &OA \\			
                \hline \hline
                FC-EF \cite{fc-siam} &  1.10 & 1.55 &72.01 &77.69 &67.10 &56.26 &92.07 & 72.57 & 72.28& 72.85&56.94 & 87.01 \\
   FC-Siam-Di \cite{fc-siam} & 1.35 & 4.25  &58.81 &47.33 &77.66 &41.66 &95.63 & 64.08& 84.94& 51.45& 47.15&86.40\\
   FC-Siam-Conc \cite{fc-siam} & 1.55 & 4.86  &66.63 &60.88 &73.58 &49.95 &97.04 & 76.89& 83.03& 71.62& 62.45& 89.35 \\
   IFNet \cite{ifnet} & 50.71  &  41.18  &83.40 &\bf96.91 &73.19 &71.52 &98.83& 76.53& 79.59& 73.58& 61.91& 89.17\\
   DTCDSCN \cite{dtcdscn} & 41.07  &  14.42  &71.95 &63.92 &82.30 &56.19 &97.42 & 80.11& 83.19& 77.25& 66.82& 90.96\\
   BIT \cite{BIT}& 11.89  & 8.71  &83.98 &86.64 &81.48 &72.39 &98.75& 79.24& 84.89& 74.29 & 65.62& 90.82 \\
   SNUNet \cite{snunet} &  12.03  & 46.70  &83.50 &85.60 &81.49 &71.67 &98.71 & 77.43& 81.93& 73.39& 63.17& 89.91 \\
   ChangeStar \cite{changestar} & 16.77 & 68.52  &87.01 &88.78 &85.31 &77.00 &98.70& -  &  - &-  & - & -\\
   LGPNet \cite{lgpnet} & 70.99  &  125.79  &79.75 &89.68 &71.81 &66.33 &98.33& 77.63&80.97   & 74.56 &63.44  &89.87\\
   ChangeFormer \cite{changeformer} & 41.03 & 202.86  &81.82 &87.25 &77.03 &69.24 &94.80& 78.21& 79.37& 77.08& 64.22& 89.87\\
   SARASNet \cite{SARASNet} & 56.89 & 139.9  & 89.55 & 88.68 & 90.44  & 81.08 & 99.05 & 79.24& \bf86.35& 73.21& 65.62& 91.08 \\
   USSFC-Net \cite{USSFCNet} & 1.52 & 3.17  & 88.93 & 91.56 & 86.43 & 80.06 & 99.01& 81.05 & 81.05 & 81.05 & 68.13 & 91.06  \\
   AFCF3D-Net \cite{AFCF3DNet} & 17.64 & 31.58 & 92.07& 93.56& 90.62& 85.30 &99.28 & 82.15& 84.22& 80.18& 69.71& \bf91.78\\
   RS-Mamba \cite{rsmamba} &51.95 & 22.82  & 91.50& 93.21& 89.85 &84.33 &99.23& 75.70 & 78.20 &73.35 & 60.90&  88.89\\
   ChangeMamba \cite{changemamba} & 48.56 &38.49 & 90.08  & 92.94 & 87.38 & 81.95 & 99.12 &  79.38& 82.43&76.54 &65.80 &90.62\\
   			\hline
   Ours & 28.74 & 15.26  & \bf92.51 & 93.45 &\bf91.59 & \bf86.07 & \bf99.32 & \bf82.66 & 81.66 &\bf83.68 &\bf70.44 &91.72 \\
			\hline
		\end{tabular}
\end{table*}

\subsection{Proposed State-Space Scan Strategies}
\label{strategy}

\subsubsection{Locally Adaptive State-Space Scan Strategy}
\label{LASSstategy}
To address the challenge highlighted in Section \ref{finding}, where existing SSM-based RSCD methods face difficulties in preserving both the local and global characteristics of RS images, we propose the Cross-Temporal Locally Adaptive State-Space Scan (LASS) strategy. As shown in Fig. \ref{fig:intro}(d), the LASS strategy is completed in three main steps. Firstly, to coarsely identify regions in the feature map $\varPhi \in \mathbb{R}^{H \times W}$ that are rich in locality, we apply ($\frac{1}{4}$, $\frac{1}{4}$) average pooling to construct a score window, where Gumbel Softmax is applied to introduce a differentiable approximation for discrete selection when identifying the top-\(k\) windows with the highest scores as: 
\begin{equation}
\label{s1}
\mathrm{Score}_{4\times 4}=\sigma \left( \mathrm{Avg\_pooling}\left( \varPhi \right) \right), 
\end{equation}
where \( \sigma \) denotes the Gumbel Softmax.

Secondly, we identify and merge connected components within these top-\(k\) windows to adapt to the varying shapes and sizes of local change regions as: 
\begin{equation}
\label{s2}
\mathrm{Loc\_wins}=\mathrm{Merge_\uparrow}\left(\mathrm{Top}\_k\left( \mathrm{Score}_{4\times 4} \right) \right),
\end{equation}
where \( \mathrm{Merge_\uparrow} \) represents the operation that merges connected components into \( k' \) connected windows, followed by an upsampling process. The matrix \( \mathrm{Loc\_wins} \in \mathbb{R}^{H \times W} \) consists of the top-\( k \) windows, renumbered by connected windows, which are assigned values from the set \( \{1,2,\dots,k'\} \), while the non-top-\( k \) windows are assigned a fixed value of 0.

The final step aims to maintain global awareness while enhancing local perception. Specifically, windows outside the top-\(k\) are treated as a single entity and flattened into a sequence for scanning (as shown by the sequence $S_0$ "\textcircled{0}" in Fig. \ref{fig:intro}(d3)). Meanwhile, the top-\(k\) windows are individually flattened and scanned sequentially (represented by sequences $S_i$ "\textcircled{1}," "\textcircled{2}," "\textcircled{3}," and "\textcircled{4}" in Fig. \ref{fig:intro}(d3)). Finally, the final sequence \( S \) is concatenated into a unified order: 
"\(\textcircled{0} - \textcircled{1} - \textcircled{2} - \textcircled{3} - \textcircled{4}\)" 
and then fed into Mamba (\( S_6 \)) to learn the intrinsic relationships. This step can be formulated as:
\begin{equation}
\label{s3}
\left\{ \begin{array}{l}
	S_0=\mathrm{Flatten}\left(  \mathrm{Loc\_wins}_0 \odot \varPhi \right),\\
	S_i=\mathrm{Flatten}\left( \mathrm{Loc\_wins}_i  \odot \varPhi \right) ,\ i=1,2,\cdots ,k',\\
    S=\bigoplus_{i=0}^k{S_i},
\end{array} \right. 
\end{equation}
where \( \mathrm{Loc\_wins}_i \) represents the portion assigned number \( i \in \{ 0,1,\dots, k' \} \), \( \odot \) denotes element-wise multiplication, and \( \bigoplus \) represents sequence concatenation.

\subsubsection{Cross-Temporal State-Space Scan Strategy}

To adapt the LASS strategy to the bi-temporal input (pre- and post-temporal images) of RSCD tasks, we introduce the Cross-Temporal State-Space Scan (CTSS) strategy within the CT-LASS module. For each pair of feature windows of size \( H_f \times W_f \), created according to Section \ref{LASSstategy}, which generate bi-temporal sequences, we employ a pixel-level cross-scan strategy. As illustrated in Fig. \ref{fig:scan}, we adhere to the principle of avoiding to increase additional space and computational complexity. We define \( N = H_f \times W_f \). 

In this sense, we first scan the bi-temporal windows of pre- and post-temporal features into four sequences of length \( N \) to satisfy the disorderliness of images, as described in VMamba \cite{liu2024vmamba}. Subsequently, we cross-concatenate the corresponding sequences pixel by pixel to form four new sequences of length \( 2N \), ensuring the full alignment of bi-temporal information at each position. By feeding these sequences into S6 \cite{liu2024vmamba}, any pixel in the bi-temporal features integrates information from all other pixels across different directions and temporal states. Subsequently, the four sequences are reshaped back into four bi-temporal windows of size \( H_f \times W_f \), followed by pixel-wise addition of these four windows into a final window. In the implementation, we first expand one dimension at the end of each sequence $S_a$ and $S_b$ to be merged. Then, the merging operation is performed along this new dimension. Finally, the last two dimensions are flattened to produce the pixel-by-pixel merged sequence $S_{merge}$ as:
\begin{equation}
\label{merge}
\left\{ \begin{array}{l}
    S'_a = \mathrm{Unsqueeze}_{-1}\left(S_a\right),\\
    S'_b = \mathrm{Unsqueeze}_{-1}\left(S_b\right),\\	
    S_{merge} = \mathrm{Flatten}_{-1,-2}\left(S'_a \oplus_{-1} S'_b\right), \\
\end{array} \right. 
\end{equation}
where \(-1\) and \(-2\) represent operations performed on the last and second-to-last dimensions, respectively, and \( \oplus \) represents the operation of concatenation. 

In summary, our Cross-Temporal State-Space Scan strategy ensures that the selective scan mechanism directly engages with both the spatial and temporal dimensions, leveraging its full potential to capture the complex dynamics in the bi-temporal windows.

\subsubsection{Window Shifting and Perception Mechanism}

Due to the LASS strategy splitting feature maps into $4 \times 4$ windows, information loss is inevitable at the boundaries of these split feature windows. To mitigate and compensate for this loss, we draw inspiration from the shifted window designed in Swin-Transformer \cite{liu2021swin}, based on which we propose a novel Window Shifting and Perception (WSP) mechanism, as illustrated in Fig. 3(d). Specifically, we apply LASS five times to the feature map at each spatial scale, where the first LASS operation is performed directly on the original feature map, while the rest four operations cyclically shifts the feature map by 1/8 of its length along the diagonal directions of 45°, 135°, 225°, and 315°, respectively, as:
\begin{equation}
\label{shiftsize}
\mathrm{Shift\_size}=\left\{ \begin{array}{l}
	\left( 0,0 \right) ,\ i=0,\\
	\left( \frac{H}{8},\frac{W}{8} \right) ,\ i=1,\\
	\left( \frac{H}{8},-\frac{W}{8} \right) ,\ i=2,\\
	\left( -\frac{H}{8},-\frac{W}{8} \right) ,\ i=3,\\
	\left( -\frac{H}{8},\frac{W}{8} \right) ,\ i=4.\\
\end{array} \right. 
\end{equation}

\setlength{\tabcolsep}{5pt}
\begin{table*}[t]
	\begin{center}
		\caption{
		Comparison of performance for RSCD on DSIFN-CD and CLCD datasets. Highest scores are in bold. All scores are in percentage. Flops are counted with image size of $256 \times 256 \times 3$.
		}
		\label{table:3}
            \begin{tabular}{l||cc||ccccc||ccccc}
		\Xhline{1.2pt}
            \rowcolor{mygray}
		     & & &\multicolumn{5}{c||}{DSIFN-CD} &\multicolumn{5}{c}{CLCD}\\
            \rowcolor{mygray}
			\multicolumn{1}{c||}{\multirow{-2}{*}{Method}}
               & \multirow{-2}{*}{Params (M)}& \multirow{-2}{*}{Flops (G)} &F1 &Pre. &Rec. &IoU &OA  &F1 &Pre. &Rec. &IoU &OA \\			
                \hline \hline
                FC-EF \cite{fc-siam} &  1.10 & 1.55  &  59.71  & 61.80  &   57.75 &  42.56 & 86.77 &48.64 &73.34 &36.29 &32.14 &94.30   \\
   FC-Siam-Di \cite{fc-siam} & 1.35 & 4.25   &  62.95 & 68.44 & 58.27 & 45.93  & 88.35 &44.10 &72.97 &31.60 &28.29 &94.04 \\
   FC-Siam-Conc \cite{fc-siam} & 1.55 & 4.86 &  60.88  & 59.08 &  62.80 &  43.76 & 86.30 &54.48&68.21&45.22&37.35&94.35\\
   IFNet \cite{ifnet} & 50.71  &  41.18  & 60.10 &  67.86 &   53.94 & 42.96 & 87.83 &48.65&49.96&47.41&32.14&92.55\\
   DTCDSCN \cite{dtcdscn} & 41.07  &  14.42  &  63.72 & 53.87 & 77.99 & 46.76 & 84.91 &60.13&62.98&57.53&42.99&94.32\\
   BIT \cite{BIT}& 11.89  & 8.71  &  69.26 &  68.36 & 70.18 & 52.97 & 89.41 &67.10&73.07&62.04&50.49&95.47\\
    SNUNet \cite{snunet} &  12.03  & 46.70   & 66.18  &  60.60 & 72.89 & 49.45 &  87.34 &66.70&73.76 &60.88&50.04&95.48\\
   ChangeStar \cite{changestar} & 16.77 & 68.52   &-  &  - &-  & - & - &60.75&62.23&59.34&43.63&94.3\\
   LGPNet \cite{lgpnet} & 70.99  &  125.79  & 63.20 & 49.96& \bf85.97 &46.19  & 82.99  &63.03&70.54&56.96&46.01&95.03  \\
   ChangeFormer \cite{changeformer} & 41.03 & 202.86  & 64.52 & 57.90 & 72.85 & 47.63 & 87.72 &58.44&65.00&53.07&41.28&94.38\\
   SARASNet \cite{SARASNet} & 56.89 & 139.9    & 67.49 & 68.18  & 66.81 & 50.93 & 88.96 & 74.70 & 76.68 & 72.83 & 59.62  & 96.33\\
   USSFC-Net \cite{USSFCNet} & 1.52 & 3.17 & 65.00 & 62.03 & 68.28 & 48.15 & 87.65 & 63.04 & 64.83 & 61.34 & 46.03 &  94.42 \\
   AFCF3D-Net \cite{AFCF3DNet} & 17.64 & 31.58  &  71.44 & 67.21  & 76.25 & 55.57 & 89.64 &76.92& \bf84.20& 70.79& 62.49 &\bf96.84 \\
   RS-Mamba \cite{rsmamba} &51.95 & 22.82  &  64.79 & 53.58  & 81.93 &47.92 &84.87 &71.27& 72.95& 69.67& \bf72.95& 95.82\\
   ChangeMamba \cite{changemamba} & 48.56 &38.49 & 65.91  & \bf70.60  & 61.80&49.15 &  89.14& 70.00 & 75.39&65.33 & 53.85 &95.83 \\
   			\hline
   Ours & 28.74 & 15.26 & \bf71.66 &67.88& 75.89& \bf55.84& \bf89.80 & \bf78.06 & 79.20 &\bf76.96  &64.02 &  96.78		\\
			\hline
		\end{tabular}
	\end{center}
\end{table*}

\subsection{Cross-Temporal Locally Adaptive State-Space Scan Module}
\label{ctlass}

We integrate our LASS strategy, CTSS strategy, and WSP mechanism (proposed in Section \ref{strategy}) into a Cross-Temporal Locally Adaptive State-Space Scan (CT-LASS) module to effectively extract and enhance bi-temporal locality from pre- and post-temporal features.

Initially, bi-temporal features $\mathcal{F}_1$ and $\mathcal{F}_2$ learned from siamese backbone are concatenated as $F_c$ along the channel dimension, which is then passed through a layer normalization step, followed by a linear transformation to adjust the dimensionality. Subsequently, depthwise convolution is applied to further extract spatial features as:
\begin{equation}
    \mathcal{F}_c=\mathcal{F}_1\oplus \mathcal{F}_2,
\end{equation}
\begin{equation}
    \mathcal{F}_{c}^{'}=\mathrm{DWConv}\left( \mathrm{Linear}\left( \mathrm{LN}\left( \mathcal{F}_c \right) \right) \right). 
\end{equation}
where $\oplus$ denotes channel-wise concatenation.

\begin{figure*}[t]
\centering
\captionsetup[subfloat]{labelsep=none,format=plain,labelformat=empty}
\subfloat[$T_1$]{
\begin{minipage}[t]{0.096\linewidth}
\includegraphics[width=1\linewidth]{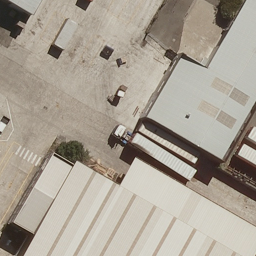}\vspace{2pt}
\includegraphics[width=1\linewidth]{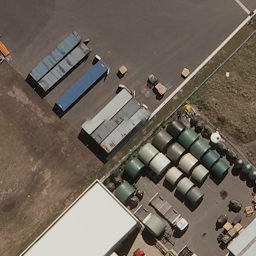}\vspace{2pt}
\includegraphics[width=1\linewidth]{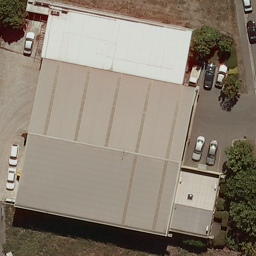}\vspace{2pt}
\includegraphics[width=1\linewidth]{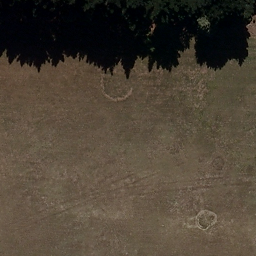}\vspace{2pt}
\end{minipage}}
\subfloat[$T_2$]{
\begin{minipage}[t]{0.096\linewidth}
\includegraphics[width=1\linewidth]{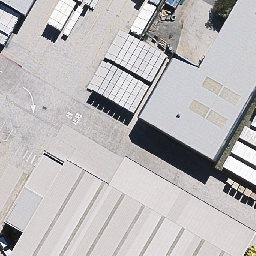}\vspace{2pt}
\includegraphics[width=1\linewidth]{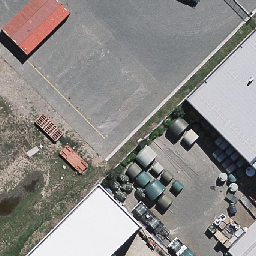}\vspace{2pt}
\includegraphics[width=1\linewidth]{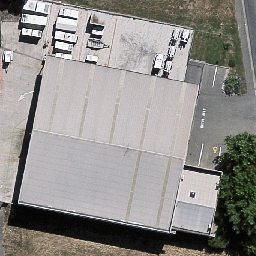}\vspace{2pt}
\includegraphics[width=1\linewidth]{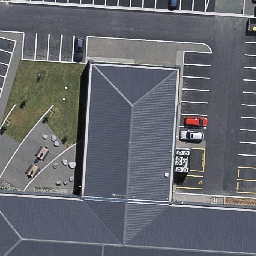}\vspace{2pt}
\end{minipage}}
\subfloat[GT]{
\begin{minipage}[t]{0.096\linewidth}
\includegraphics[width=1\linewidth]{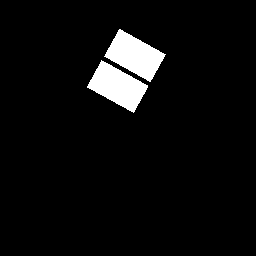}\vspace{2pt}
\includegraphics[width=1\linewidth]{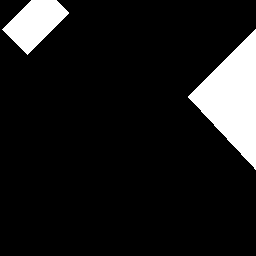}\vspace{2pt}
\includegraphics[width=1\linewidth]{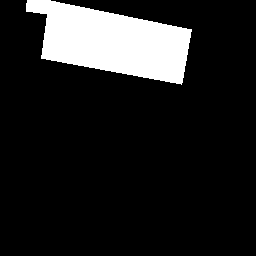}\vspace{2pt}
\includegraphics[width=1\linewidth]{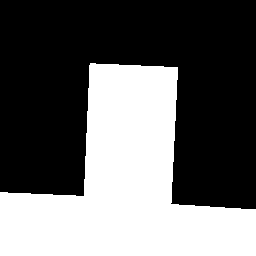}\vspace{2pt}
\end{minipage}}
\subfloat[SNUNet]{
\begin{minipage}[t]{0.096\linewidth}
\includegraphics[width=1\linewidth]{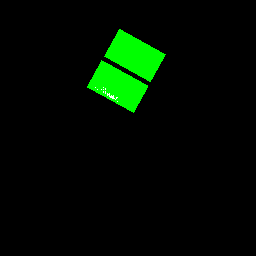}\vspace{2pt}
\includegraphics[width=1\linewidth]{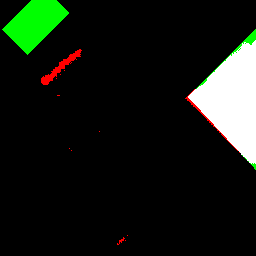}\vspace{2pt}
\includegraphics[width=1\linewidth]{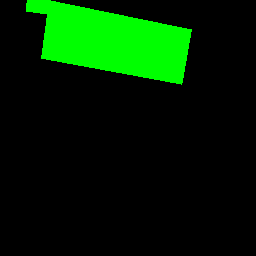}\vspace{2pt}
\includegraphics[width=1\linewidth]{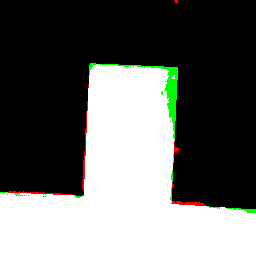}\vspace{2pt}
\end{minipage}}
\subfloat[BIT]{
\begin{minipage}[t]{0.096\linewidth}
\includegraphics[width=1\linewidth]{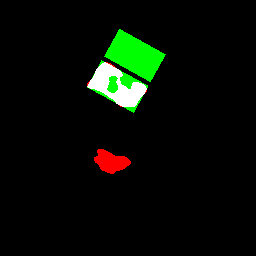}\vspace{2pt}
\includegraphics[width=1\linewidth]{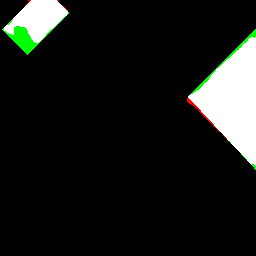}\vspace{2pt}
\includegraphics[width=1\linewidth]{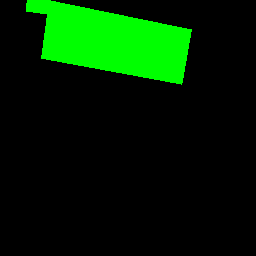}\vspace{2pt}
\includegraphics[width=1\linewidth]{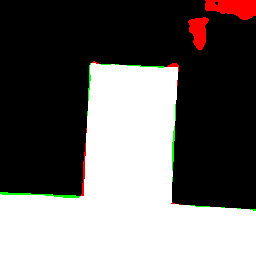}\vspace{2pt}
\end{minipage}}
\subfloat[SARASNet]{
\begin{minipage}[t]{0.096\linewidth}
\includegraphics[width=1\linewidth]{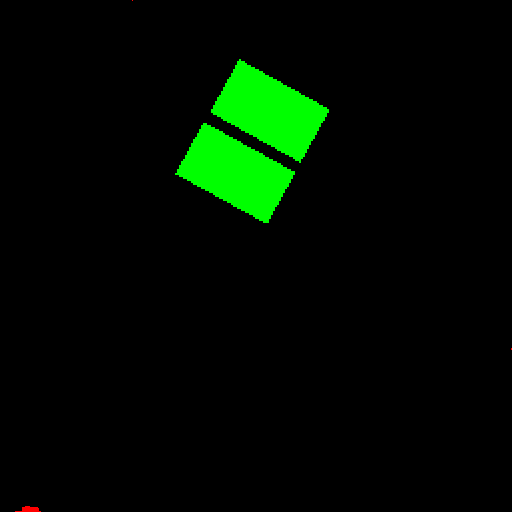}\vspace{2pt}
\includegraphics[width=1\linewidth]{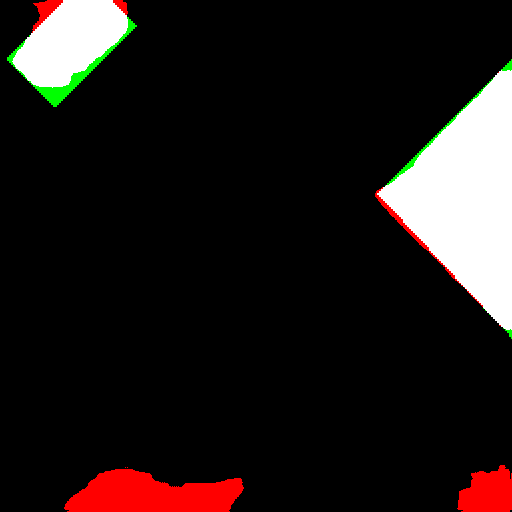}\vspace{2pt}
\includegraphics[width=1\linewidth]{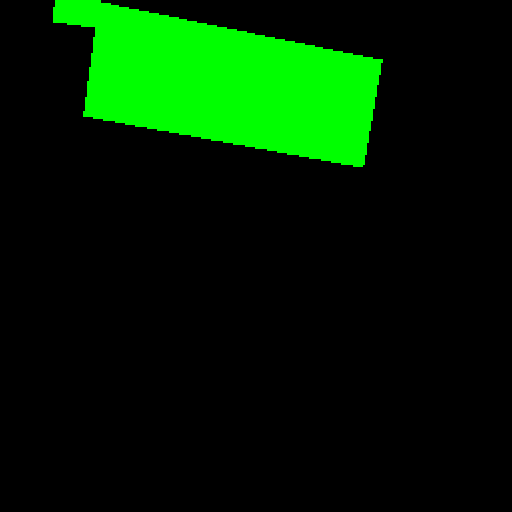}\vspace{2pt}
\includegraphics[width=1\linewidth]{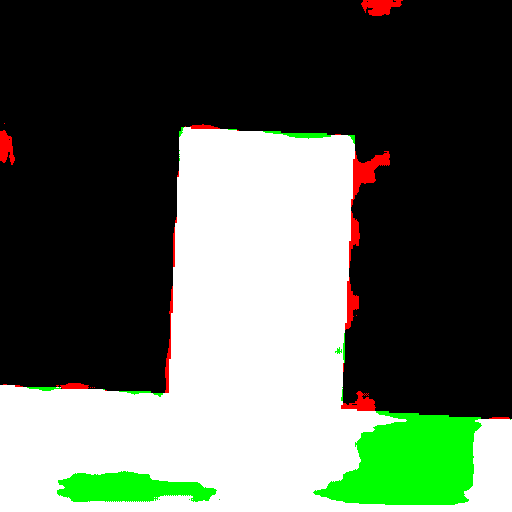}\vspace{2pt}
\end{minipage}}
\subfloat[AFCD3DNet]{
\begin{minipage}[t]{0.096\linewidth}
\includegraphics[width=1\linewidth]{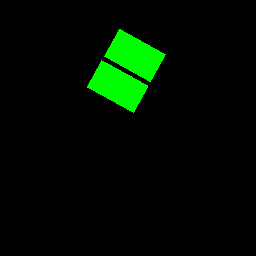}\vspace{2pt}
\includegraphics[width=1\linewidth]{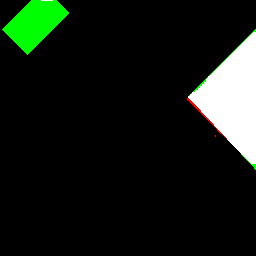}\vspace{2pt}
\includegraphics[width=1\linewidth]{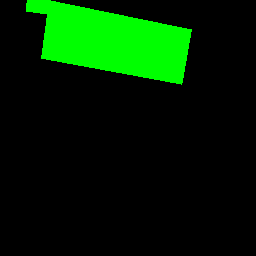}\vspace{2pt}
\includegraphics[width=1\linewidth]{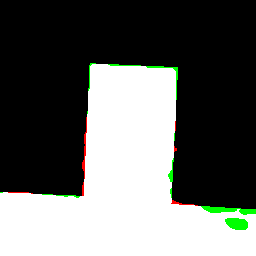}\vspace{2pt}
\end{minipage}}
\subfloat[RSMamba]{
\begin{minipage}[t]{0.096\linewidth}
\includegraphics[width=1\linewidth]{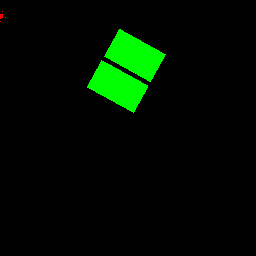}\vspace{2pt}
\includegraphics[width=1\linewidth]{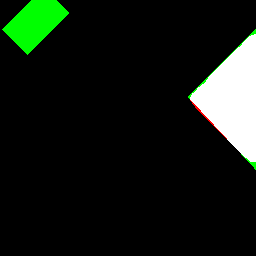}\vspace{2pt}
\includegraphics[width=1\linewidth]{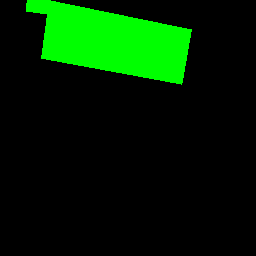}\vspace{2pt}
\includegraphics[width=1\linewidth]{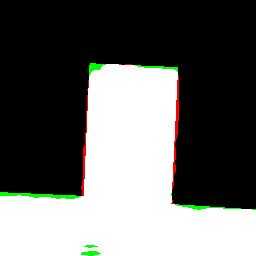}\vspace{2pt}
\end{minipage}}
\subfloat[ChangeMamba]{
\begin{minipage}[t]{0.096\linewidth}
\includegraphics[width=1\linewidth]{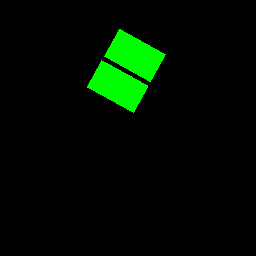}\vspace{2pt}
\includegraphics[width=1\linewidth]{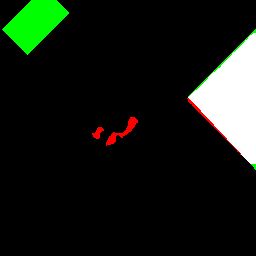}\vspace{2pt}
\includegraphics[width=1\linewidth]{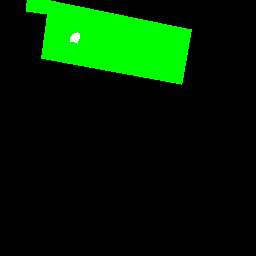}\vspace{2pt}
\includegraphics[width=1\linewidth]{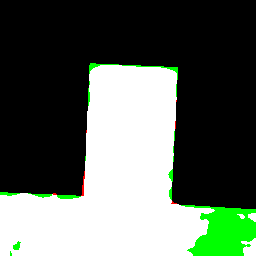}\vspace{2pt}
\end{minipage}}
\subfloat[CDLamba]{
\begin{minipage}[t]{0.096\linewidth}
\includegraphics[width=1\linewidth]{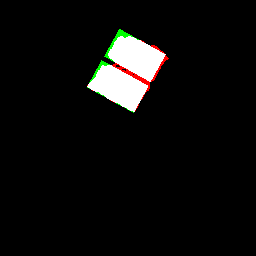}\vspace{2pt}
\includegraphics[width=1\linewidth]{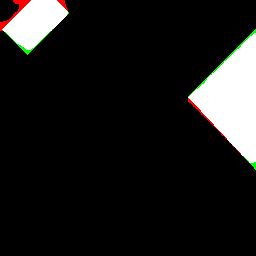}\vspace{2pt}
\includegraphics[width=1\linewidth]{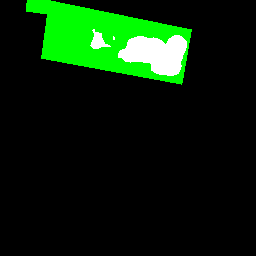}\vspace{2pt}
\includegraphics[width=1\linewidth]{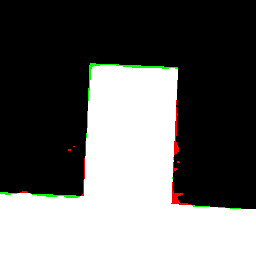}\vspace{2pt}
\end{minipage}}
\caption{Example results output from RSCD methods on test sets from WHU-CD dataset. Pixels are colored differently for better visualization (i.e., white for true positive, black for true negative, red for false positive, and green for false negative).}
\label{fig:whucd}
\end{figure*}

\begin{figure*}[t]
\centering
\captionsetup[subfloat]{labelsep=none,format=plain,labelformat=empty}
\subfloat[$T_1$]{
\begin{minipage}[t]{0.096\linewidth}
\includegraphics[width=1\linewidth]{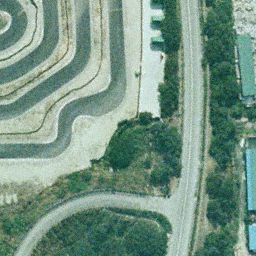}\vspace{2pt}
\includegraphics[width=1\linewidth]{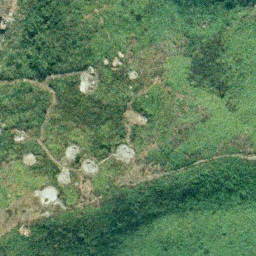}\vspace{2pt}
\includegraphics[width=1\linewidth]{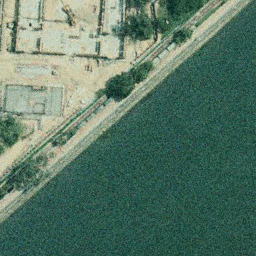}\vspace{2pt}
\includegraphics[width=1\linewidth]{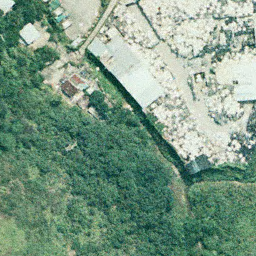}\vspace{2pt}
\end{minipage}}
\subfloat[$T_2$]{
\begin{minipage}[t]{0.096\linewidth}
\includegraphics[width=1\linewidth]{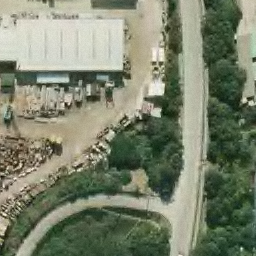}\vspace{2pt}
\includegraphics[width=1\linewidth]{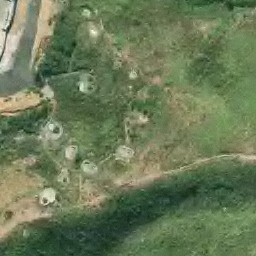}\vspace{2pt}
\includegraphics[width=1\linewidth]{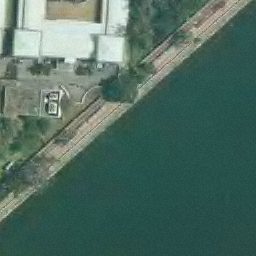}\vspace{2pt}
\includegraphics[width=1\linewidth]{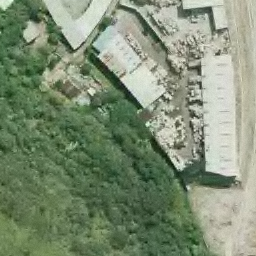}\vspace{2pt}
\end{minipage}}
\subfloat[GT]{
\begin{minipage}[t]{0.096\linewidth}
\includegraphics[width=1\linewidth]{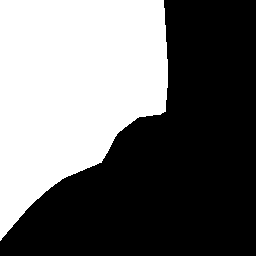}\vspace{2pt}
\includegraphics[width=1\linewidth]{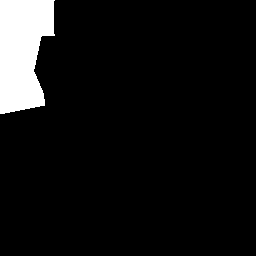}\vspace{2pt}
\includegraphics[width=1\linewidth]{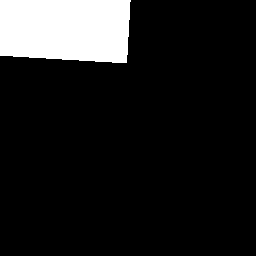}\vspace{2pt}
\includegraphics[width=1\linewidth]{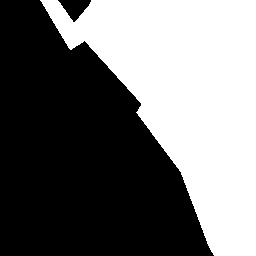}\vspace{2pt}
\end{minipage}}
\subfloat[SNUNet]{
\begin{minipage}[t]{0.096\linewidth}
\includegraphics[width=1\linewidth]{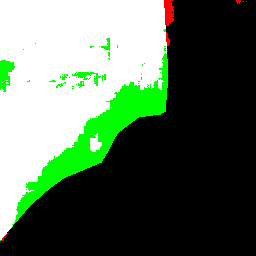}\vspace{2pt}
\includegraphics[width=1\linewidth]{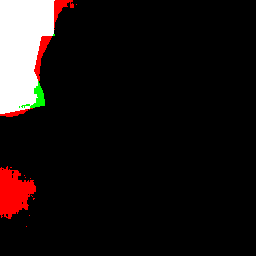}\vspace{2pt}
\includegraphics[width=1\linewidth]{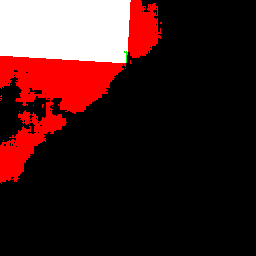}\vspace{2pt}
\includegraphics[width=1\linewidth]{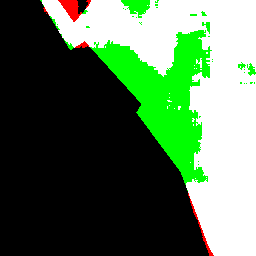}\vspace{2pt}
\end{minipage}}
\subfloat[BIT]{
\begin{minipage}[t]{0.096\linewidth}
\includegraphics[width=1\linewidth]{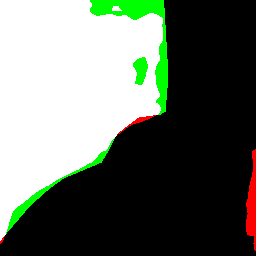}\vspace{2pt}
\includegraphics[width=1\linewidth]{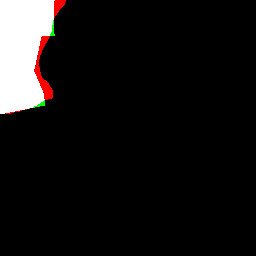}\vspace{2pt}
\includegraphics[width=1\linewidth]{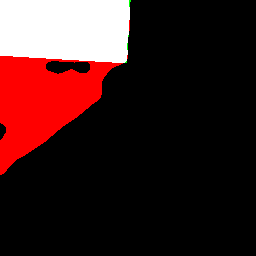}\vspace{2pt}
\includegraphics[width=1\linewidth]{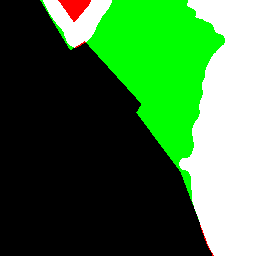}\vspace{2pt}
\end{minipage}}
\subfloat[SARASNet]{
\begin{minipage}[t]{0.096\linewidth}
\includegraphics[width=1\linewidth]{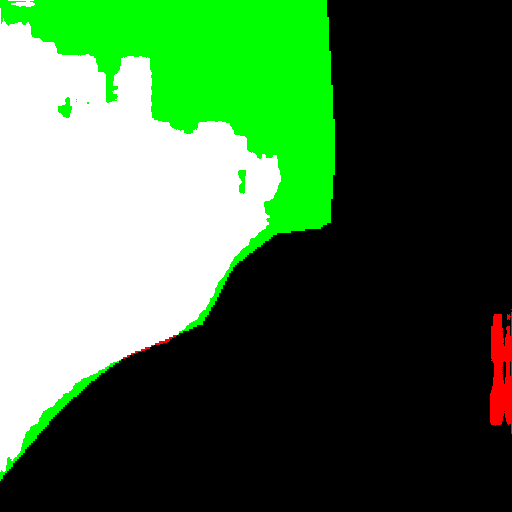}\vspace{2pt}
\includegraphics[width=1\linewidth]{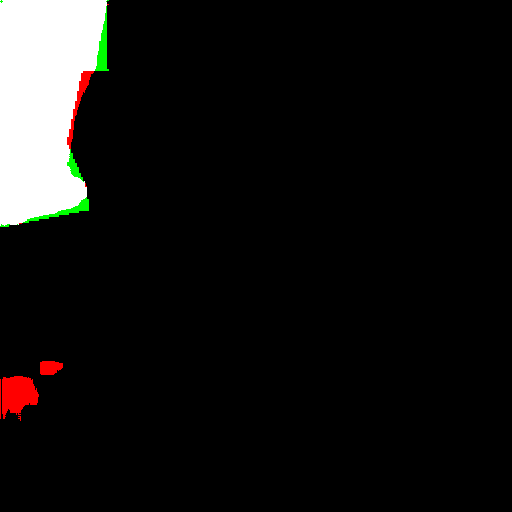}\vspace{2pt}
\includegraphics[width=1\linewidth]{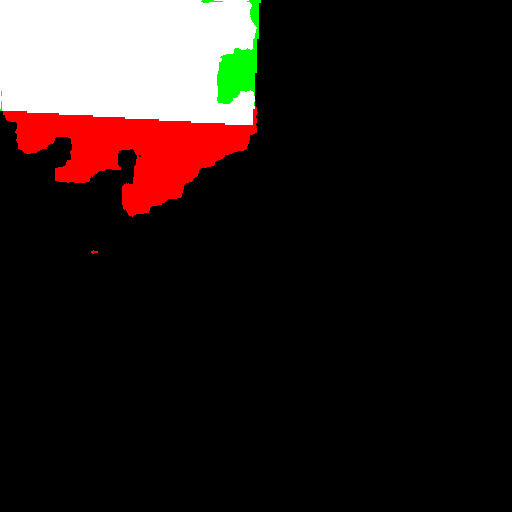}\vspace{2pt}
\includegraphics[width=1\linewidth]{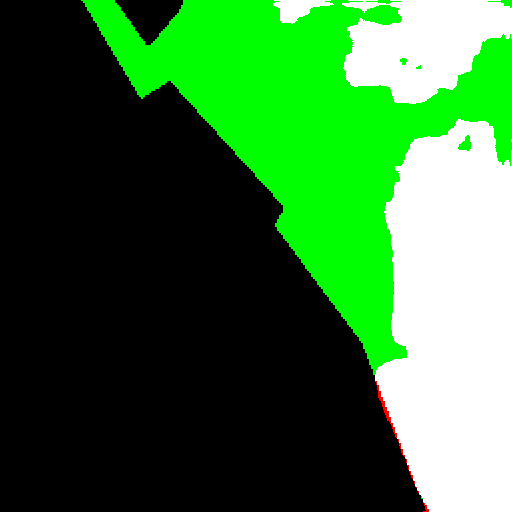}\vspace{2pt}
\end{minipage}}
\subfloat[AFCD3DNet]{
\begin{minipage}[t]{0.096\linewidth}
\includegraphics[width=1\linewidth]{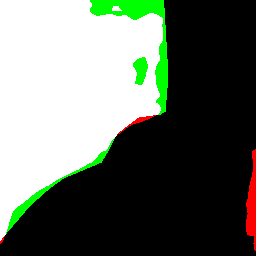}\vspace{2pt}
\includegraphics[width=1\linewidth]{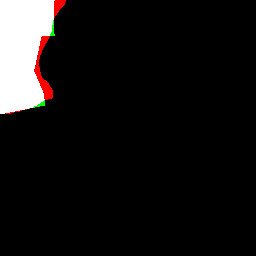}\vspace{2pt}
\includegraphics[width=1\linewidth]{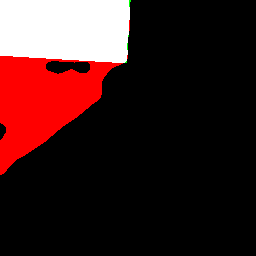}\vspace{2pt}
\includegraphics[width=1\linewidth]{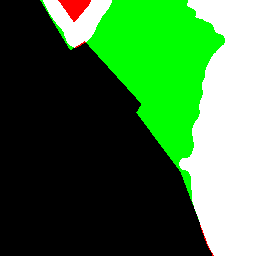}\vspace{2pt}
\end{minipage}}
\subfloat[RSMamba]{
\begin{minipage}[t]{0.096\linewidth}
\includegraphics[width=1\linewidth]{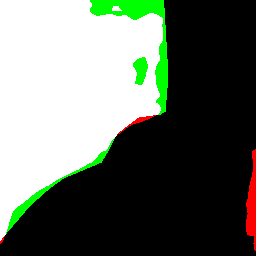}\vspace{2pt}
\includegraphics[width=1\linewidth]{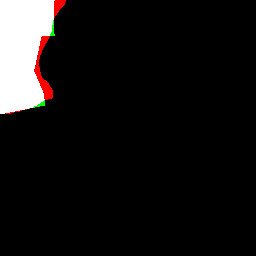}\vspace{2pt}
\includegraphics[width=1\linewidth]{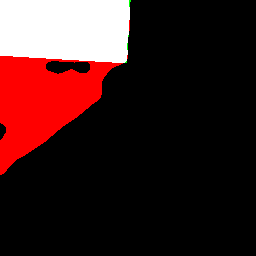}\vspace{2pt}
\includegraphics[width=1\linewidth]{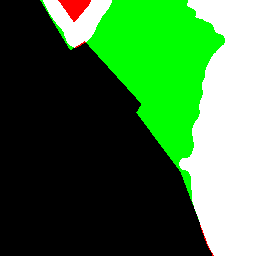}\vspace{2pt}
\end{minipage}}
\subfloat[ChangeMamba]{
\begin{minipage}[t]{0.096\linewidth}
\includegraphics[width=1\linewidth]{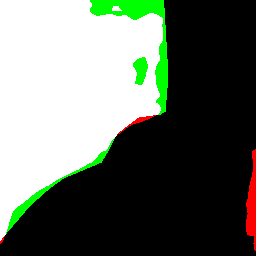}\vspace{2pt}
\includegraphics[width=1\linewidth]{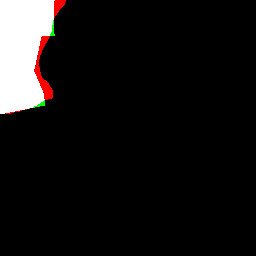}\vspace{2pt}
\includegraphics[width=1\linewidth]{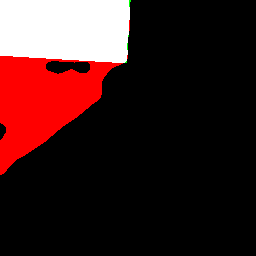}\vspace{2pt}
\includegraphics[width=1\linewidth]{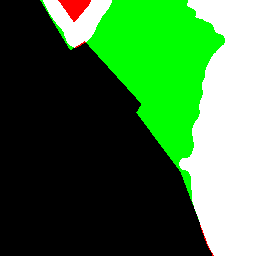}\vspace{2pt}
\end{minipage}}
\subfloat[CDLamba]{
\begin{minipage}[t]{0.096\linewidth}
\includegraphics[width=1\linewidth]{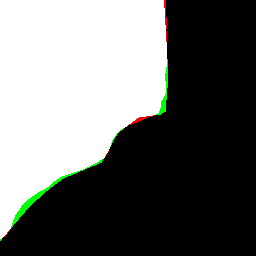}\vspace{2pt}
\includegraphics[width=1\linewidth]{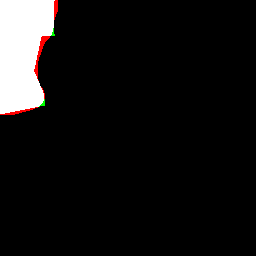}\vspace{2pt}
\includegraphics[width=1\linewidth]{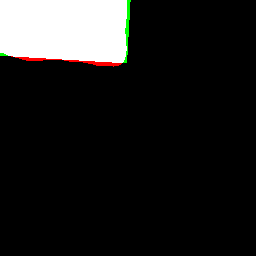}\vspace{2pt}
\includegraphics[width=1\linewidth]{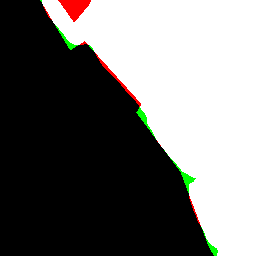}\vspace{2pt}
\end{minipage}}
\caption{Example results output from RSCD methods on test sets from SYSU-CD dataset. Pixels are colored differently for better visualization (i.e., white for true positive, black for true negative, red for false positive, and green for false negative).}
\label{fig:sysucd}
\end{figure*}

To facilitate interactions between adjacent windows and reduce spatial information loss at the boundaries of windows, the feature map \( \mathcal{F}_{c}^{'} \) is shifted along one direction by 1/8 of its total length using a Window Shift and Padding (WSP) mechanism.
\begin{equation}
    \mathcal{F}_{c}^{s}=\mathrm{WSP}\left( \mathcal{F}_{c}^{'} \right). 
\end{equation}

The bi-temporal features are subsequently split from \( \mathcal{F}_{c}^{s} \) along the channel dimension into \( \mathcal{F}_{c}^{s1} \) and \( \mathcal{F}_{c}^{s2} \), and processed using the LASS strategy. In this step, a score window is generated by applying an average pooling operation with a ($\frac{1}{4}$, $\frac{1}{4}$) kernel to the difference $\varPhi$ of the bi-temporal features. 
\begin{equation}
\label{varphi}
\varPhi =\mathcal{F}_{c}^{s1}\ominus \mathcal{F}_{c}^{s2},
\end{equation}
where $\ominus$ denotes element-wise subtraction.

Then \( \mathrm{Loc\_wins}_i \) is calculated by Eqs. \eqref{s1} and \eqref{s2}. The bi-temporal sequences \( S_a \) and \( S_b \) are then calculated using Eqs. \eqref{sa} and \eqref{sb}.

\begin{equation}
\label{sa}
\left\{ \begin{array}{l}
	S_0^a=\mathrm{Flatten}\left(  \mathrm{Loc\_wins}_0 \odot \mathcal{F}_{c}^{s1} \right),\\
	S_i^a=\mathrm{Flatten}\left( \mathrm{Loc\_wins}_i  \odot \mathcal{F}_{c}^{s1} \right) ,\ i=1,2,\cdots ,k',\\
    S_a=\bigoplus_{i=0}^k{S_i^a},
\end{array} \right. 
\end{equation}

\begin{equation}
\label{sb}
\left\{ \begin{array}{l}
	S_0^b=\mathrm{Flatten}\left(  \mathrm{Loc\_wins}_0 \odot \mathcal{F}_{c}^{s2} \right),\\
	S_i^b=\mathrm{Flatten}\left( \mathrm{Loc\_wins}_i  \odot \mathcal{F}_{c}^{s2} \right) ,\ i=1,2,\cdots ,k',\\
    S_b=\bigoplus_{i=0}^k{S_i^b},
\end{array} \right. 
\end{equation}
where \( \odot \) denotes element-wise multiplication and \( \bigoplus \) represents sequence concatenation.

Following the above \( S_a \) and \( S_b \), our CTSS strategy is employed to scan the bi-temporal sequences in the LASS process as: 
\begin{equation}
S_{merge}=\mathrm{CTSS}\left( S_a,\ S_b \right). 
\end{equation}

After processing \( S_{merge} \) with the CTSS strategy, the original bi-temporal feature map dimensions are restored, resulting in \( \hat{\mathcal{F}}_{c}^{s1} \) and \( \hat{\mathcal{F}}_{c}^{s2} \). The processed bi-temporal features are then concatenated along the channel dimension, and a Reverse WSP operation is applied to restore the original pixel positions. Finally, the flow passes through another layer normalization and linear transformation.
\begin{equation}
\hat{\mathcal{F}}_c=\hat{\mathcal{F}}_{c}^{s1}\oplus \hat{\mathcal{F}}_{c}^{s2},
\end{equation}
\begin{equation}
\mathring{\mathcal{F}}=\mathrm{Linear}\left( \mathrm{LN}\left( \mathrm{Reverse\_WSP}\left( \hat{\mathcal{F}}_c \right) \right) \right), 
\end{equation}
where $\oplus$ denotes channel-wise concatenation.

This process is repeated five times at each scale, with the WSP shift direction alternating in each iteration. The outputs $\mathring{\mathcal{F}}_i$ of these five iterations are then aggregated using a 1×1 convolution layer to produce the final bi-temporal output $\mathcal{F}_o$.
\begin{equation}
\mathcal{F}_o=\mathrm{Conv}\left( \mathring{\mathcal{F}}_0 \oplus \mathring{\mathcal{F}}_1 \oplus\mathring{\mathcal{F}}_2 \oplus\mathring{\mathcal{F}}_3 \oplus\mathring{\mathcal{F}}_4 \right), 
\end{equation}
where $\oplus$ denotes channel-wise concatenation.

Finally, the outputted bi-temporal features \( \mathcal{F}_1^o \) and \( \mathcal{F}_2^o \) are obtained by splitting \( \mathcal{F}_o \) along the channel dimension.

\begin{figure*}[t]
\centering
\captionsetup[subfloat]{labelsep=none,format=plain,labelformat=empty}
\subfloat[$T_1$]{
\begin{minipage}[t]{0.096\linewidth}
\includegraphics[width=1\linewidth]{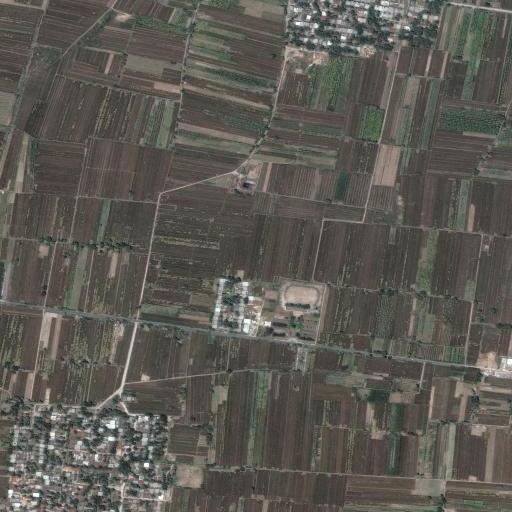}\vspace{2pt}
\includegraphics[width=1\linewidth]{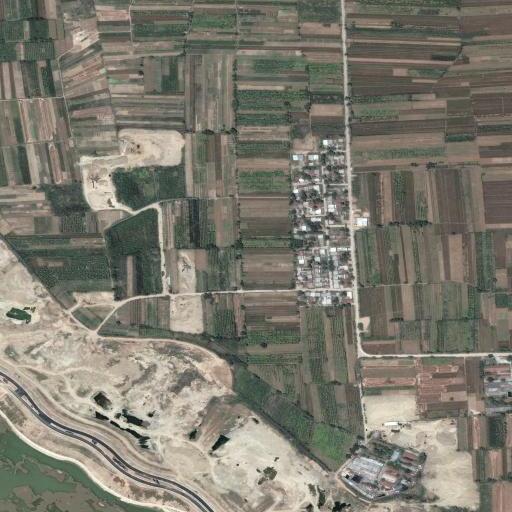}\vspace{2pt}
\includegraphics[width=1\linewidth]{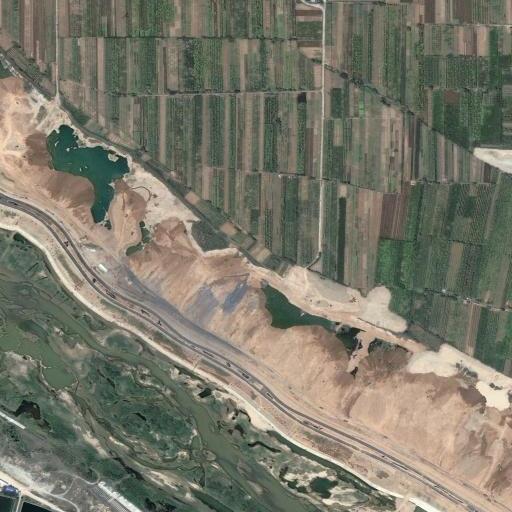}\vspace{2pt}
\includegraphics[width=1\linewidth]{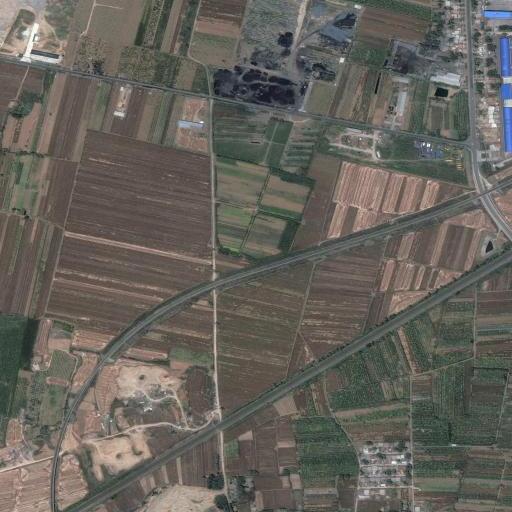}\vspace{2pt}
\end{minipage}}
\subfloat[$T_2$]{
\begin{minipage}[t]{0.096\linewidth}
\includegraphics[width=1\linewidth]{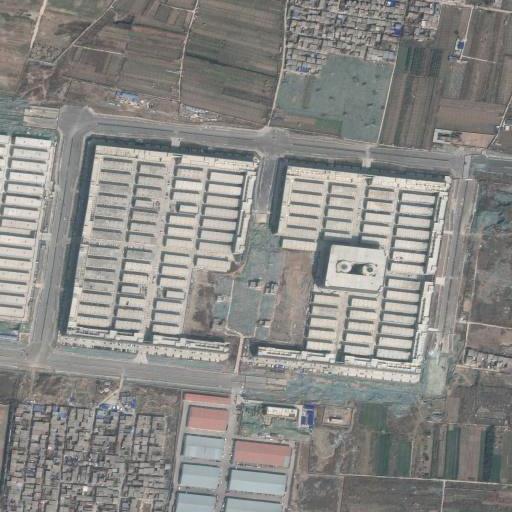}\vspace{2pt}
\includegraphics[width=1\linewidth]{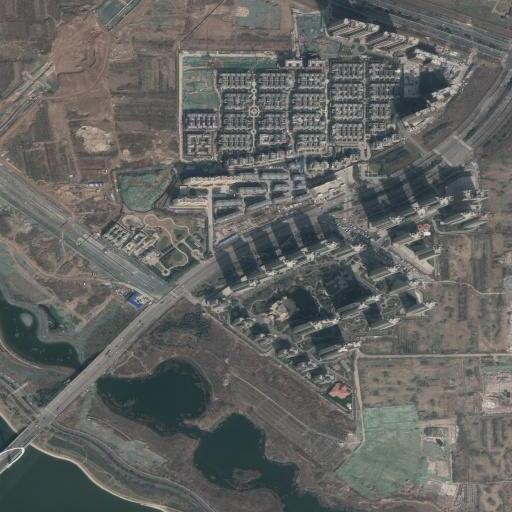}\vspace{2pt}
\includegraphics[width=1\linewidth]{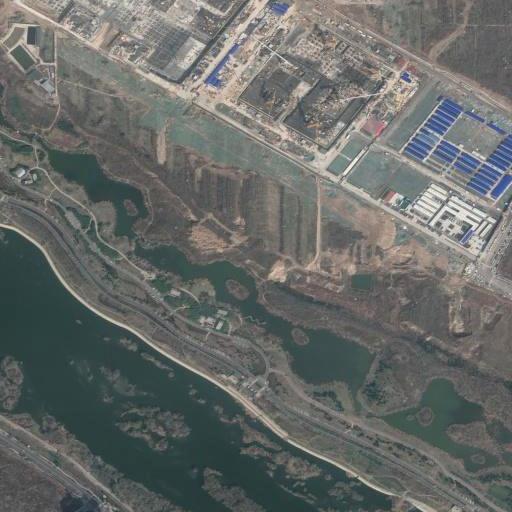}\vspace{2pt}
\includegraphics[width=1\linewidth]{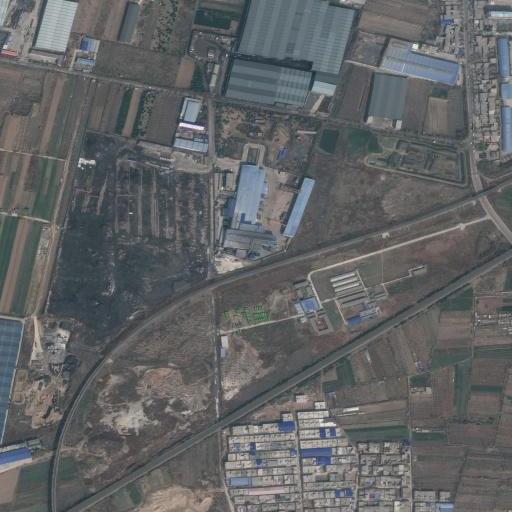}\vspace{2pt}
\end{minipage}}
\subfloat[GT]{
\begin{minipage}[t]{0.096\linewidth}
\includegraphics[width=1\linewidth]{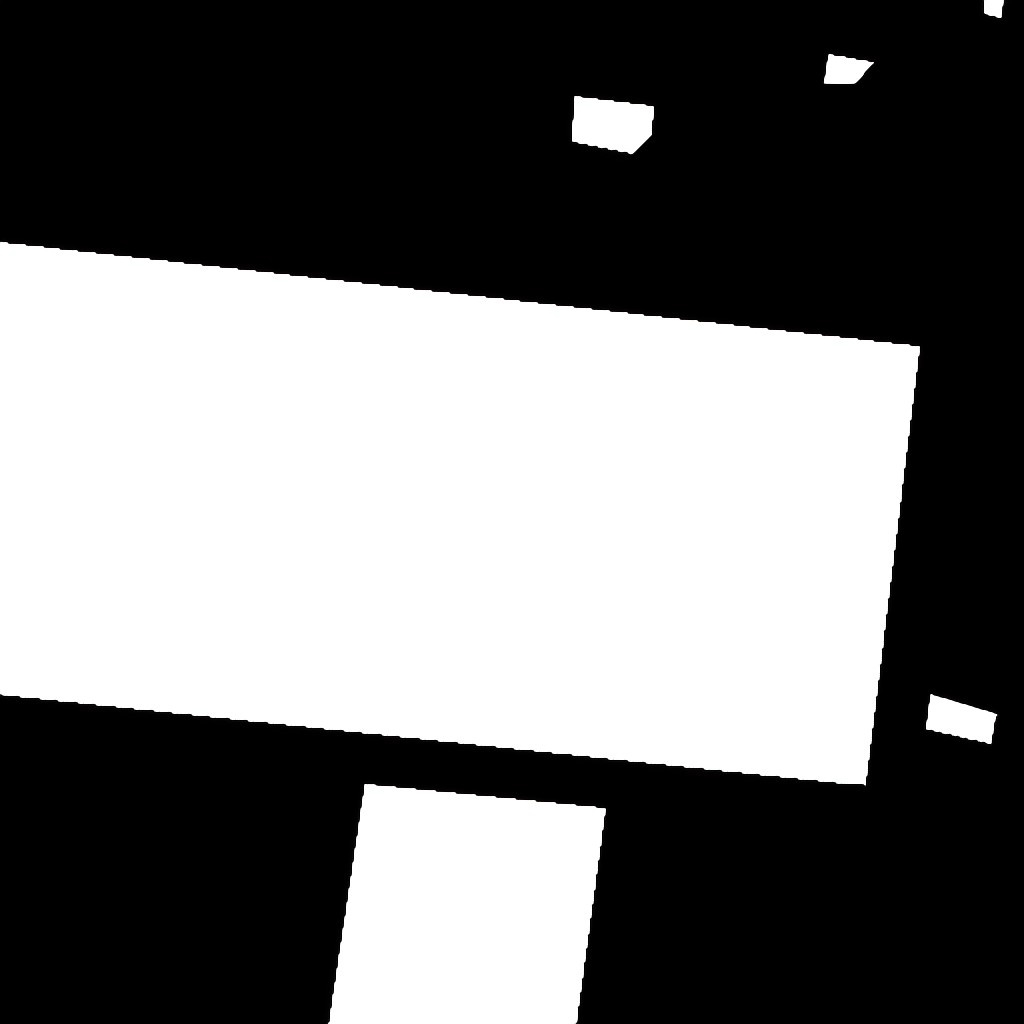}\vspace{2pt}
\includegraphics[width=1\linewidth]{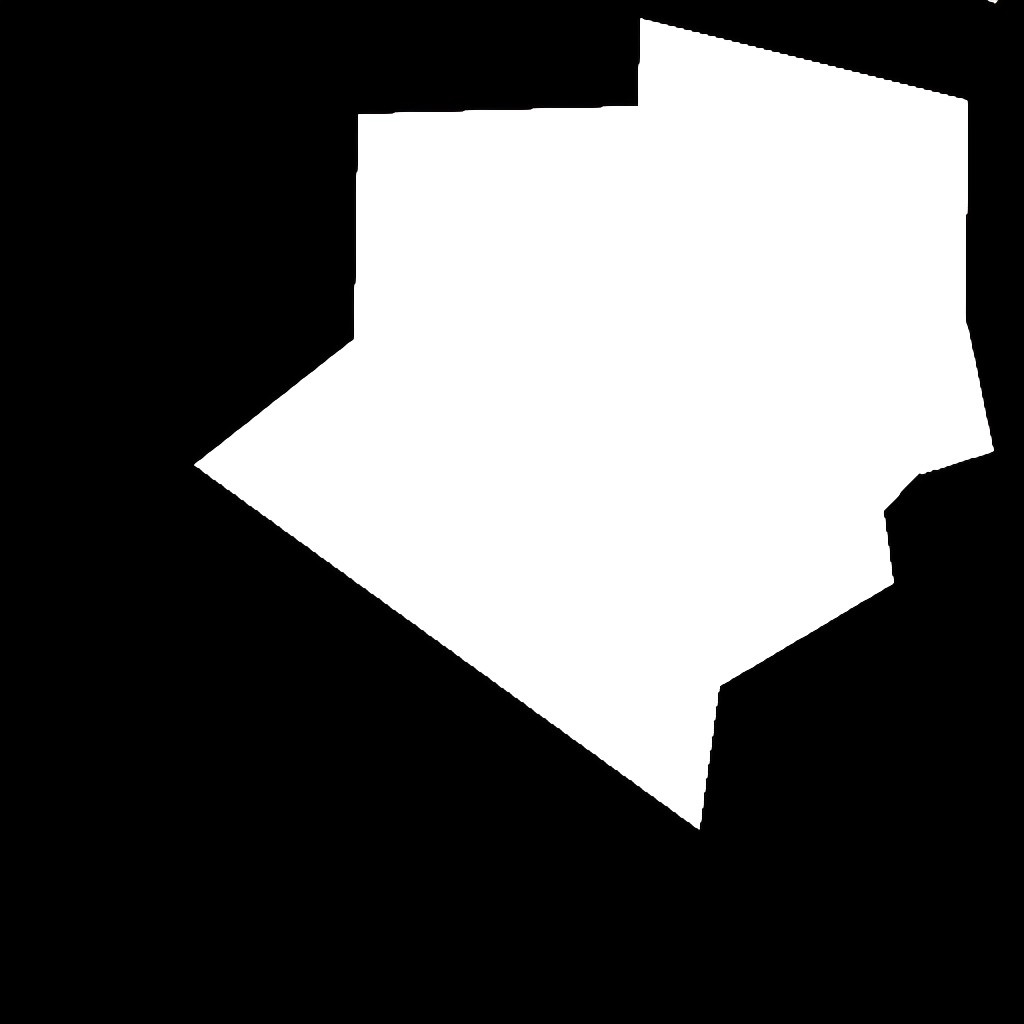}\vspace{2pt}
\includegraphics[width=1\linewidth]{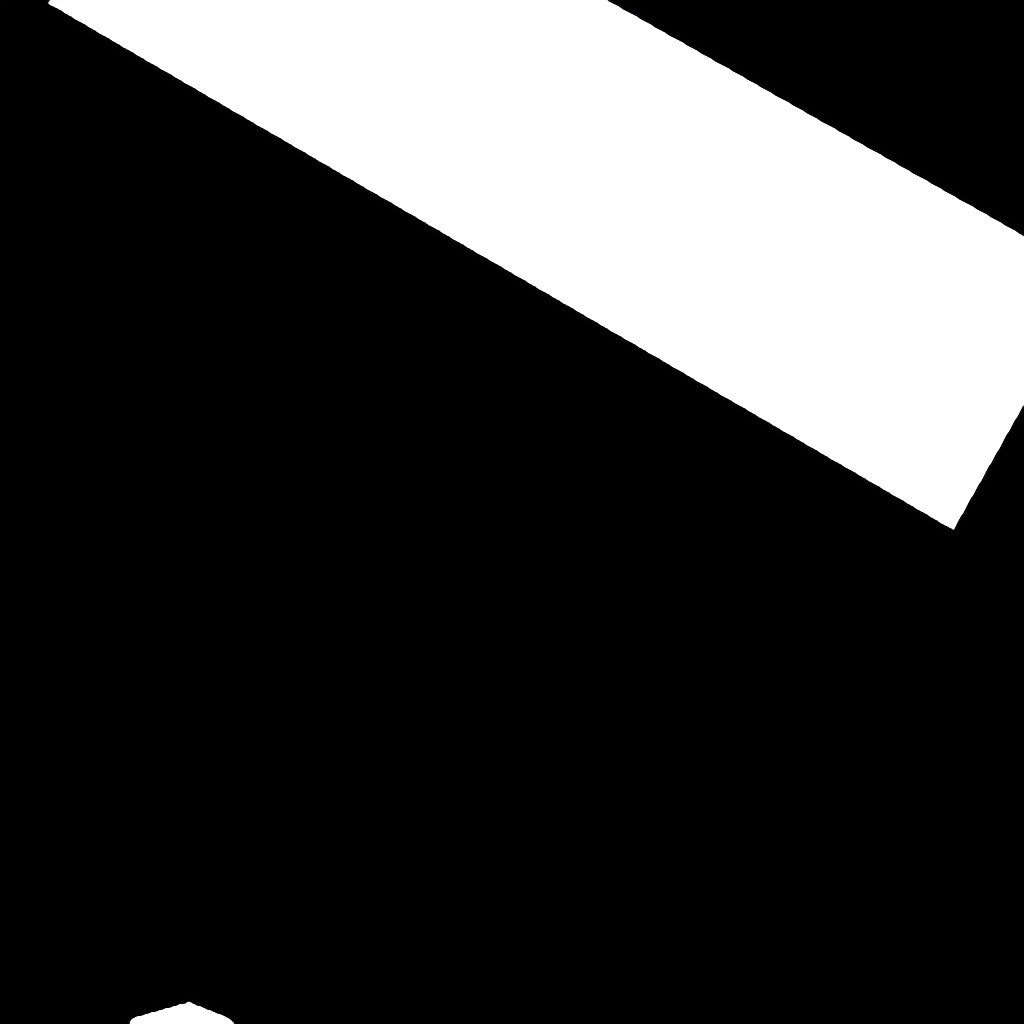}\vspace{2pt}
\includegraphics[width=1\linewidth]{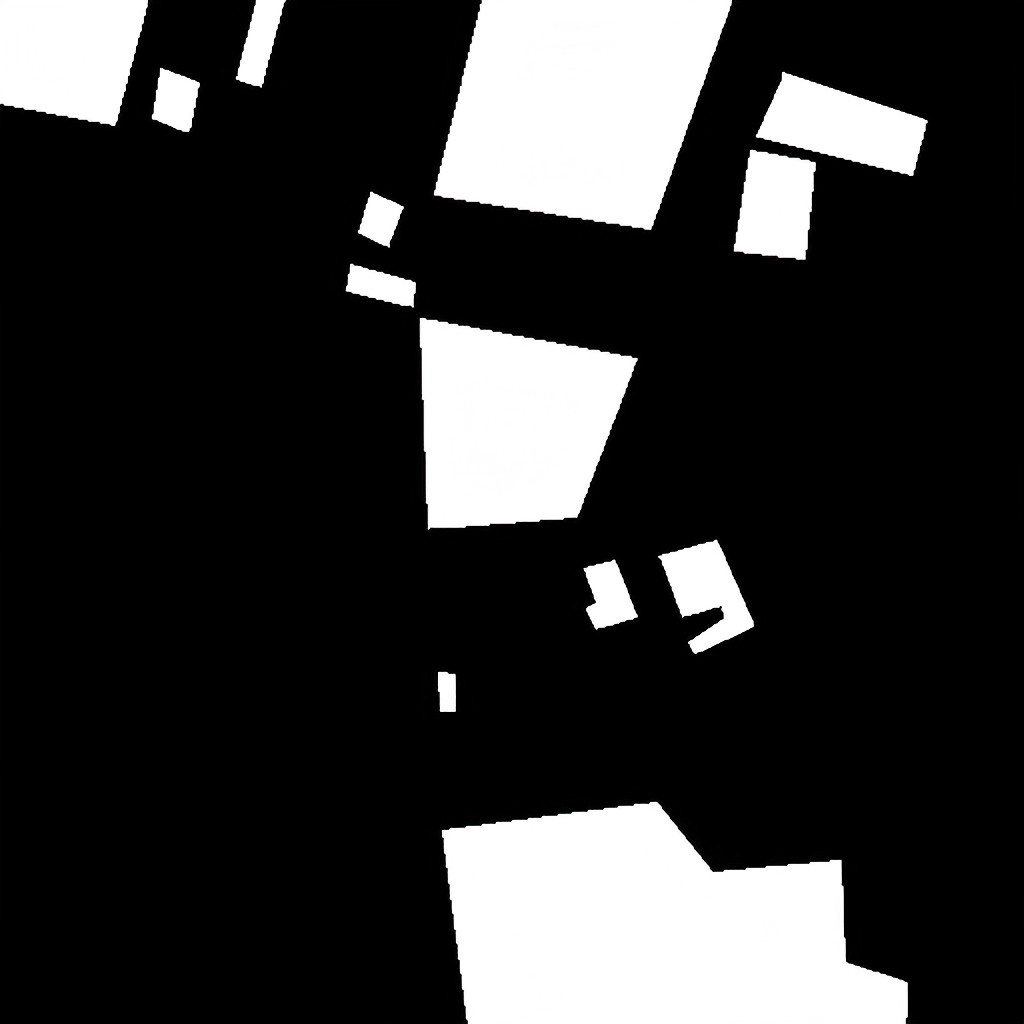}\vspace{2pt}
\end{minipage}}
\subfloat[SNUNet]{
\begin{minipage}[t]{0.096\linewidth}
\includegraphics[width=1\linewidth]{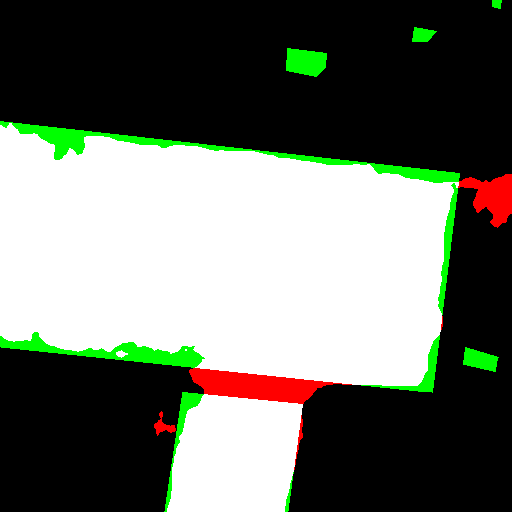}\vspace{2pt}
\includegraphics[width=1\linewidth]{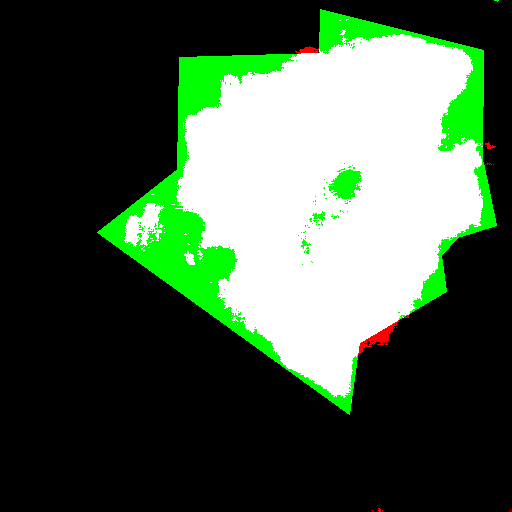}\vspace{2pt}
\includegraphics[width=1\linewidth]{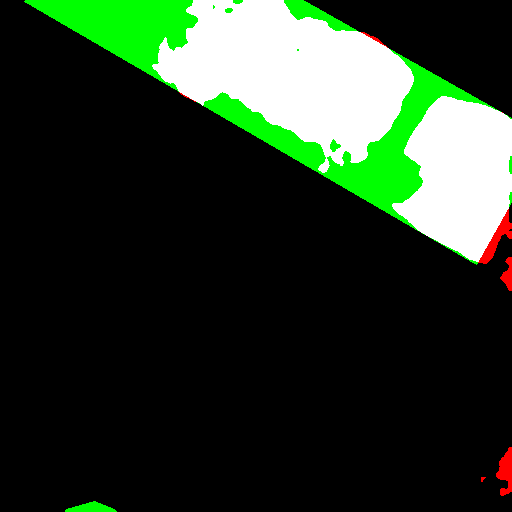}\vspace{2pt}
\includegraphics[width=1\linewidth]{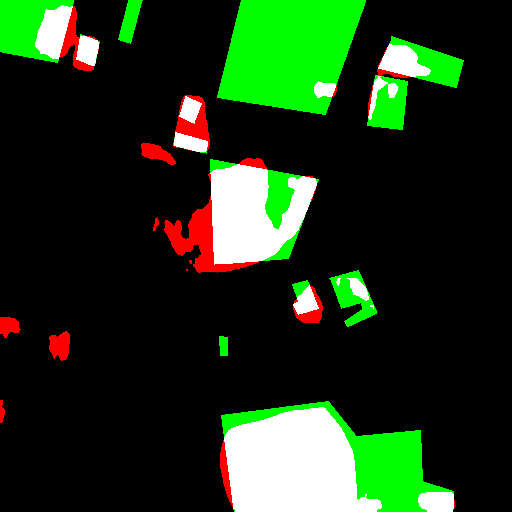}\vspace{2pt}
\end{minipage}}
\subfloat[BIT]{
\begin{minipage}[t]{0.096\linewidth}
\includegraphics[width=1\linewidth]{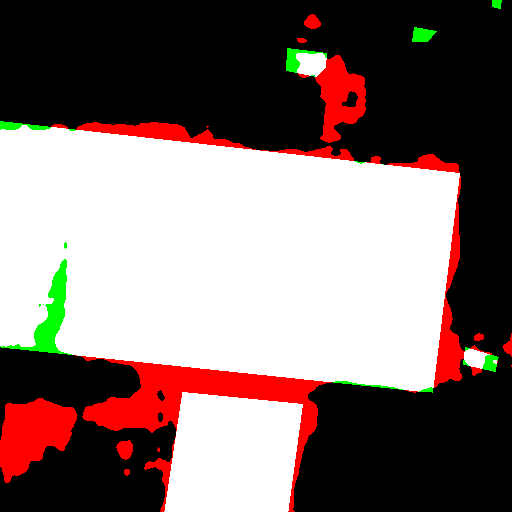}\vspace{2pt}
\includegraphics[width=1\linewidth]{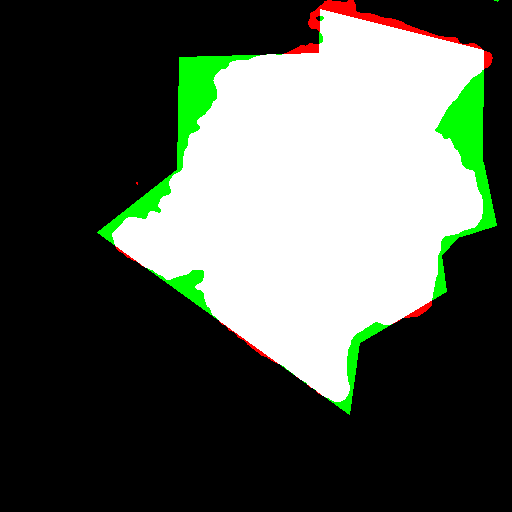}\vspace{2pt}
\includegraphics[width=1\linewidth]{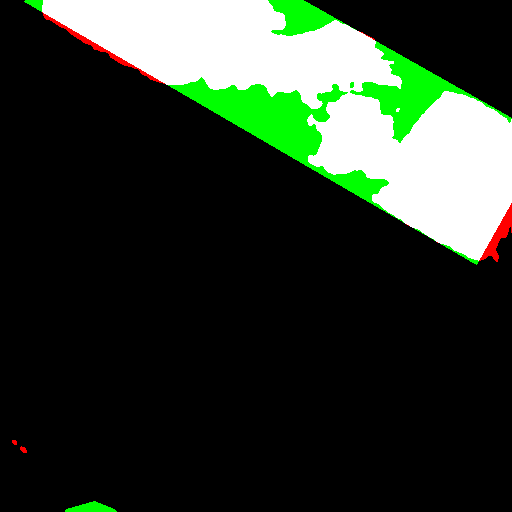}\vspace{2pt}
\includegraphics[width=1\linewidth]{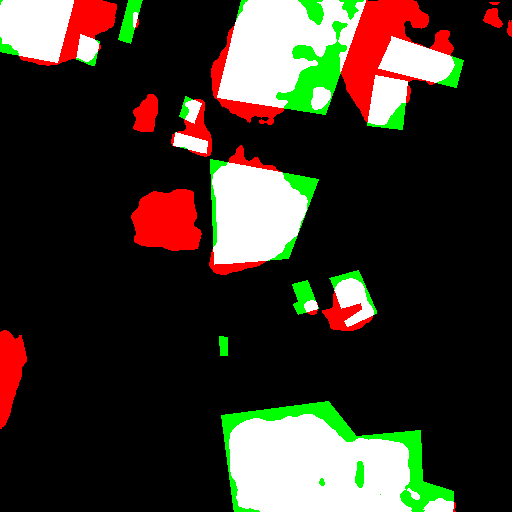}\vspace{2pt}
\end{minipage}}
\subfloat[SARASNet]{
\begin{minipage}[t]{0.096\linewidth}
\includegraphics[width=1\linewidth]{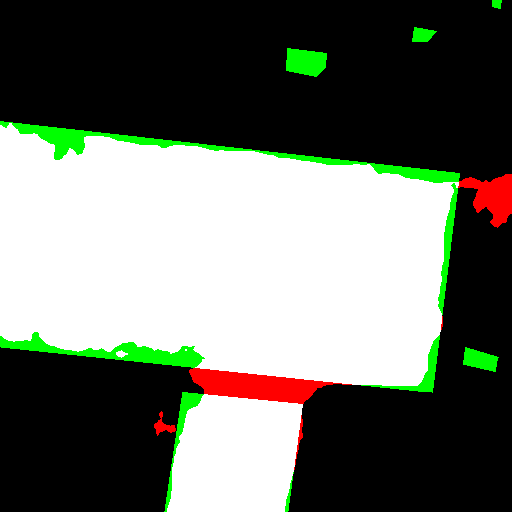}\vspace{2pt}
\includegraphics[width=1\linewidth]{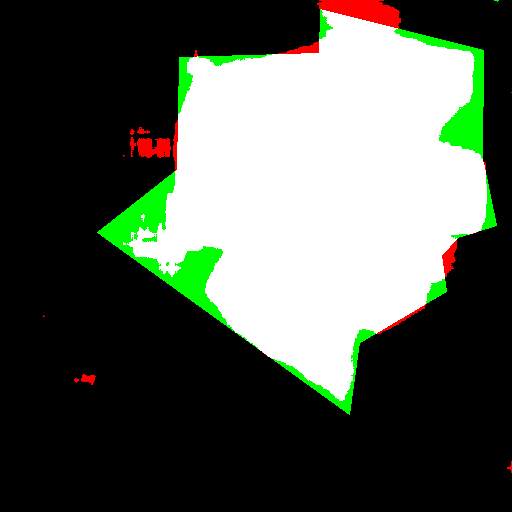}\vspace{2pt}
\includegraphics[width=1\linewidth]{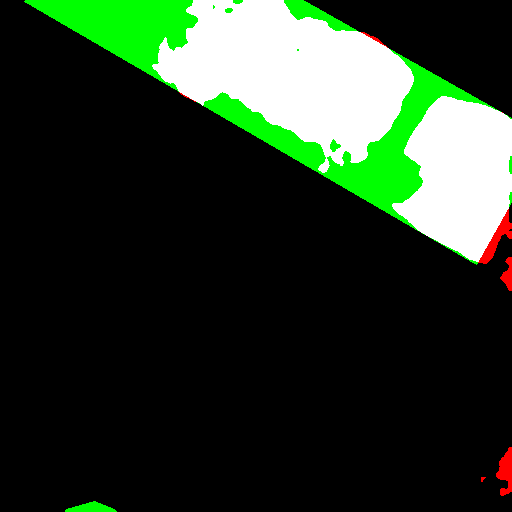}\vspace{2pt}
\includegraphics[width=1\linewidth]{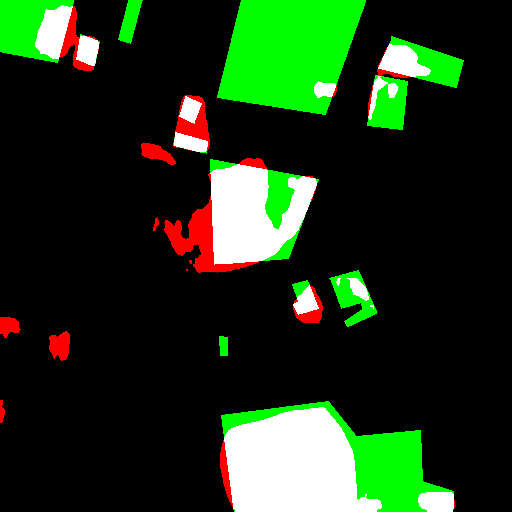}\vspace{2pt}
\end{minipage}}
\subfloat[AFCD3DNet]{
\begin{minipage}[t]{0.096\linewidth}
\includegraphics[width=1\linewidth]{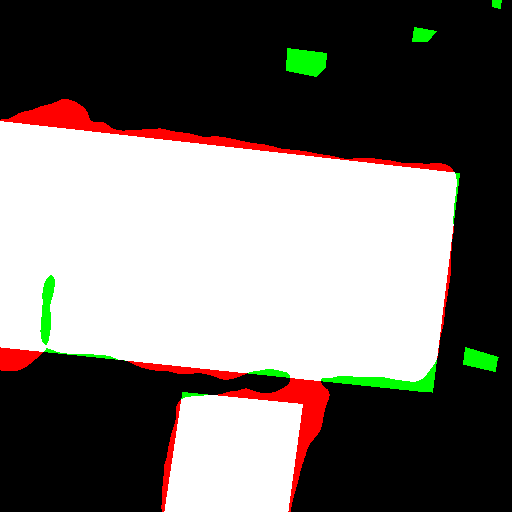}\vspace{2pt}
\includegraphics[width=1\linewidth]{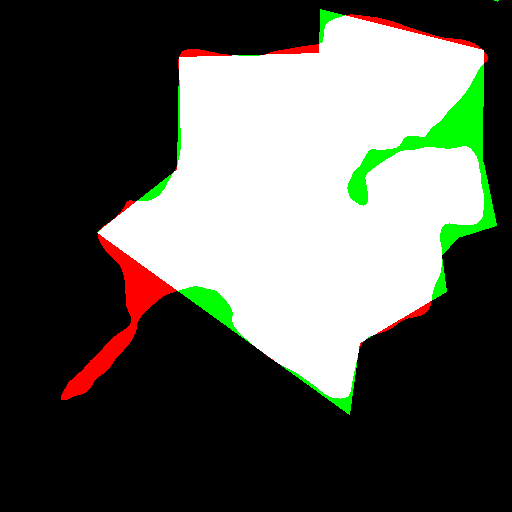}\vspace{2pt}
\includegraphics[width=1\linewidth]{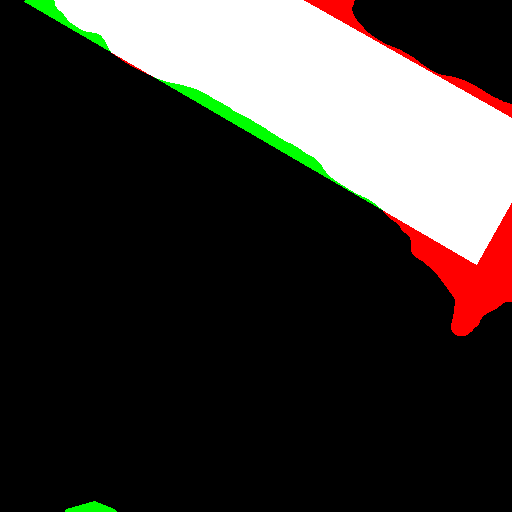}\vspace{2pt}
\includegraphics[width=1\linewidth]{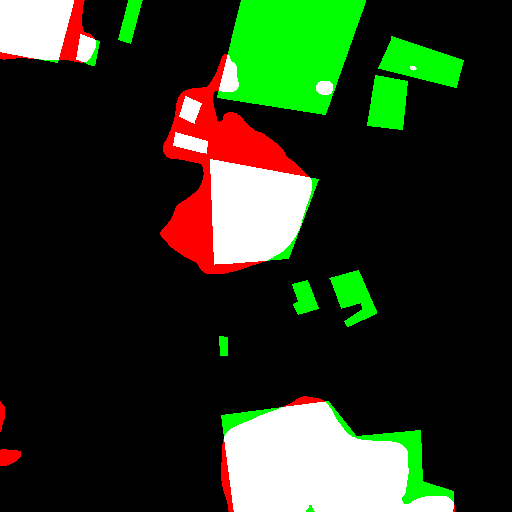}\vspace{2pt}
\end{minipage}}
\subfloat[RSMamba]{
\begin{minipage}[t]{0.096\linewidth}
\includegraphics[width=1\linewidth]{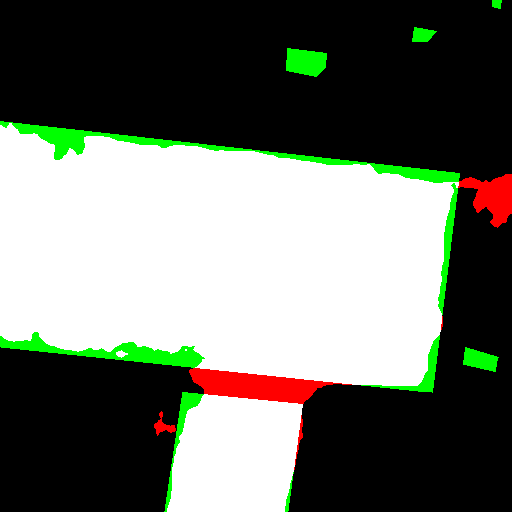}\vspace{2pt}
\includegraphics[width=1\linewidth]{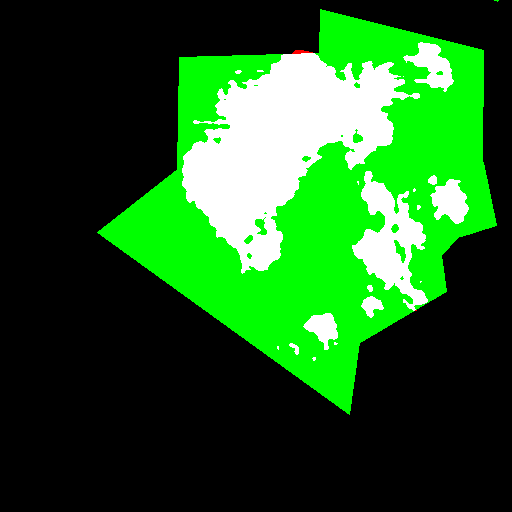}\vspace{2pt}
\includegraphics[width=1\linewidth]{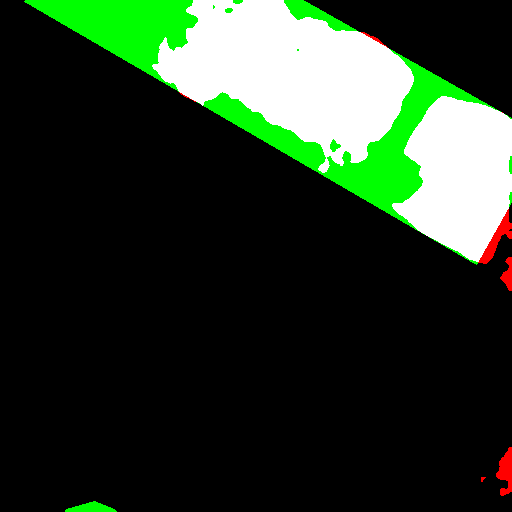}\vspace{2pt}
\includegraphics[width=1\linewidth]{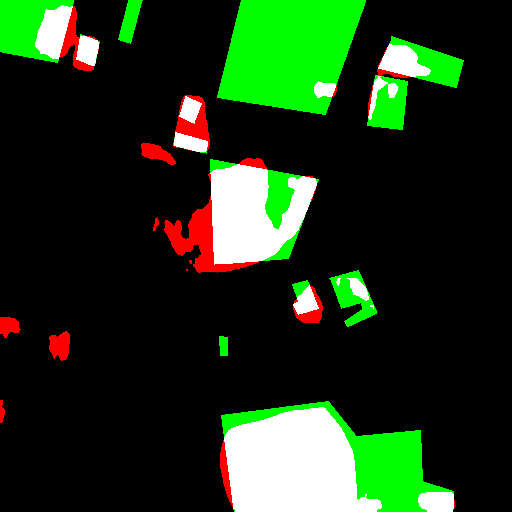}\vspace{2pt}
\end{minipage}}
\subfloat[ChangeMamba]{
\begin{minipage}[t]{0.096\linewidth}
\includegraphics[width=1\linewidth]{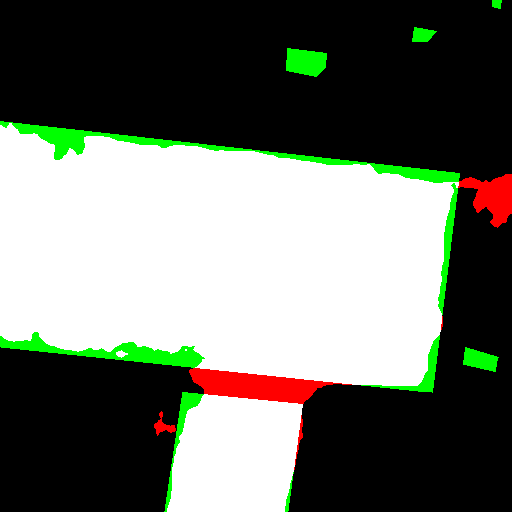}\vspace{2pt}
\includegraphics[width=1\linewidth]{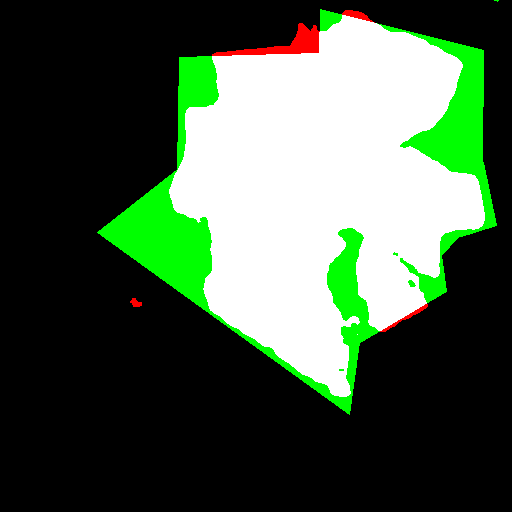}\vspace{2pt}
\includegraphics[width=1\linewidth]{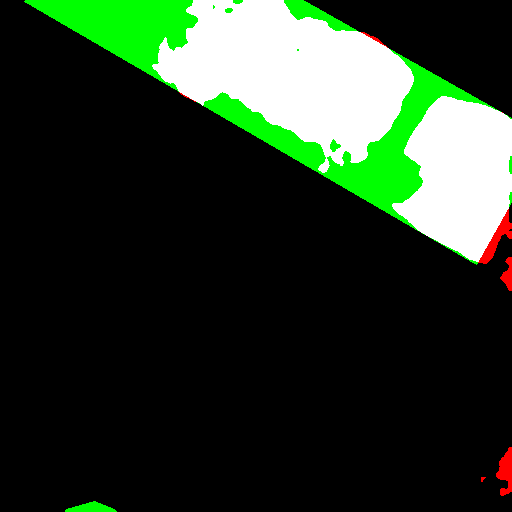}\vspace{2pt}
\includegraphics[width=1\linewidth]{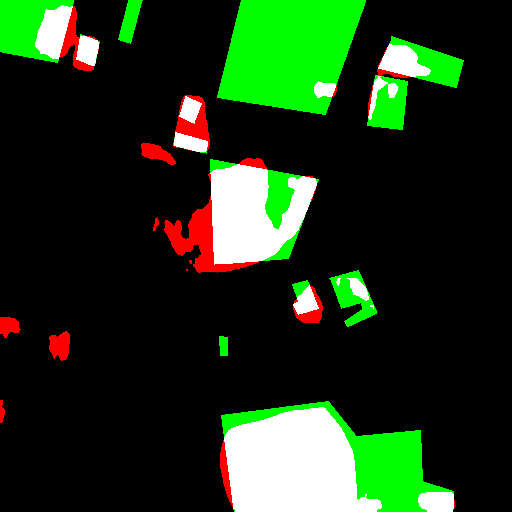}\vspace{2pt}
\end{minipage}}
\subfloat[CDLamba]{
\begin{minipage}[t]{0.096\linewidth}
\includegraphics[width=1\linewidth]{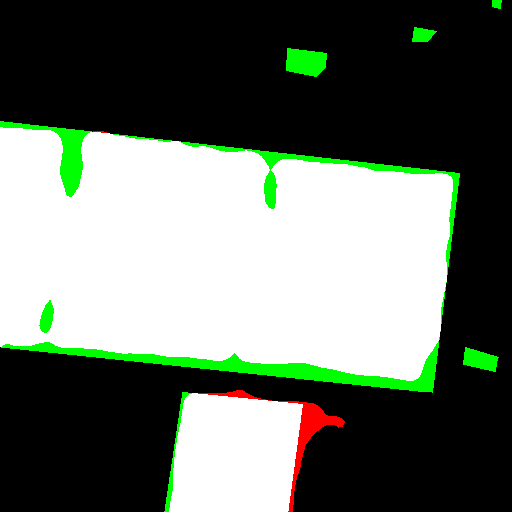}\vspace{2pt}
\includegraphics[width=1\linewidth]{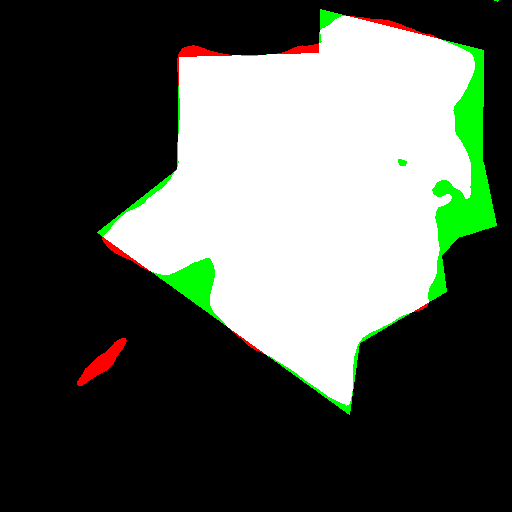}\vspace{2pt}
\includegraphics[width=1\linewidth]{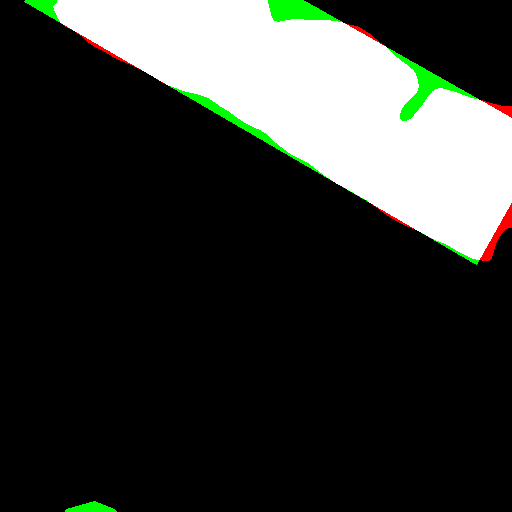}\vspace{2pt}
\includegraphics[width=1\linewidth]{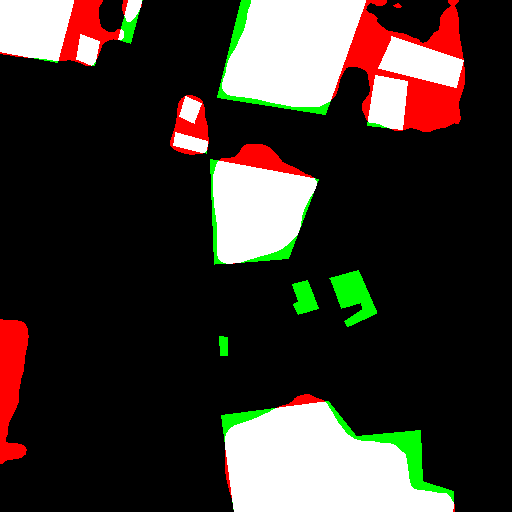}\vspace{2pt}
\end{minipage}}
\caption{Example results output from RSCD methods on test sets from DSIFN-CD dataset. Pixels are colored differently for better visualization (i.e., white for true positive, black for true negative, red for false positive, and green for false negative).}
\label{fig:dsifn}
\end{figure*}

\begin{figure*}[t]
\centering
\captionsetup[subfloat]{labelsep=none,format=plain,labelformat=empty}
\subfloat[$T_1$]{
\begin{minipage}[t]{0.096\linewidth}
\includegraphics[width=1\linewidth]{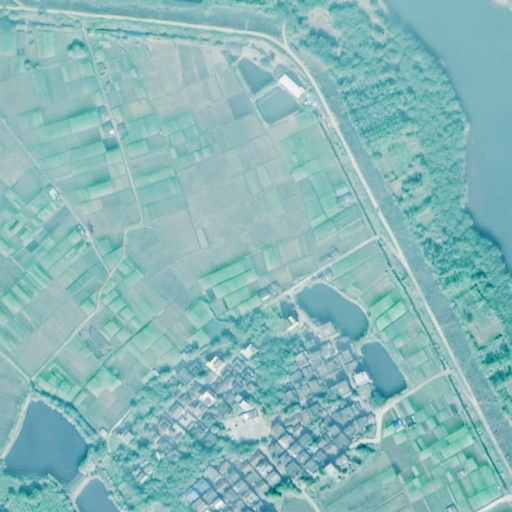}\vspace{2pt}
\includegraphics[width=1\linewidth]{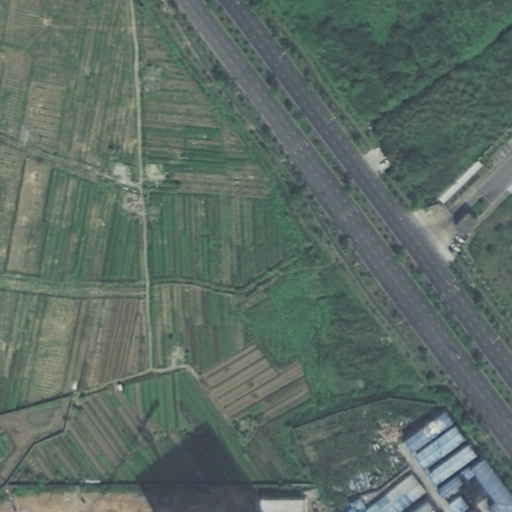}\vspace{2pt}
\includegraphics[width=1\linewidth]{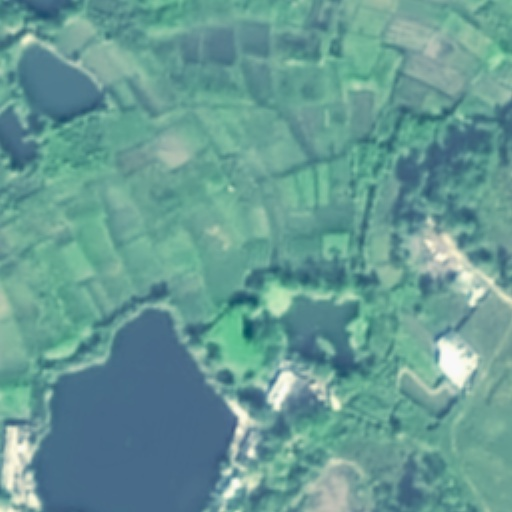}\vspace{2pt}
\includegraphics[width=1\linewidth]{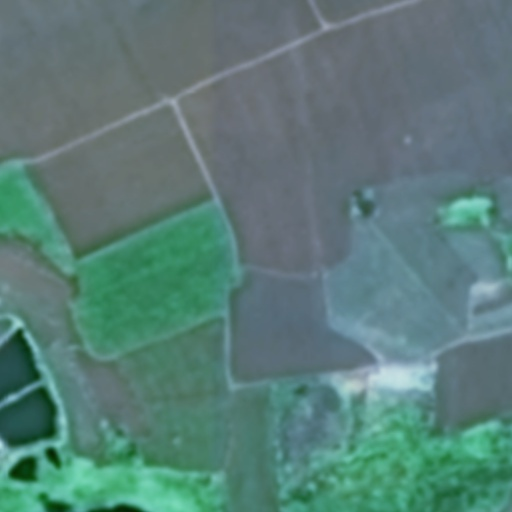}\vspace{2pt}
\end{minipage}}
\subfloat[$T_2$]{
\begin{minipage}[t]{0.096\linewidth}
\includegraphics[width=1\linewidth]{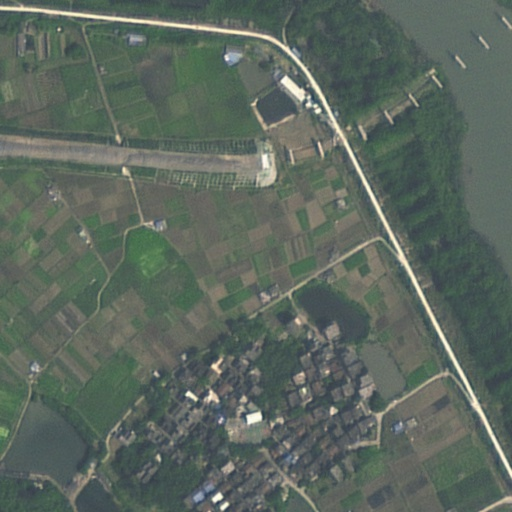}\vspace{2pt}
\includegraphics[width=1\linewidth]{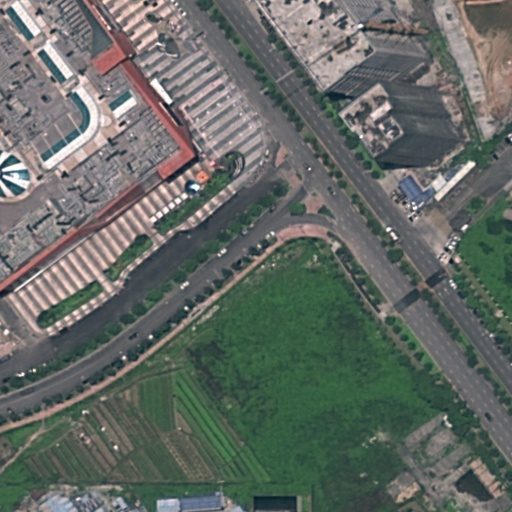}\vspace{2pt}
\includegraphics[width=1\linewidth]{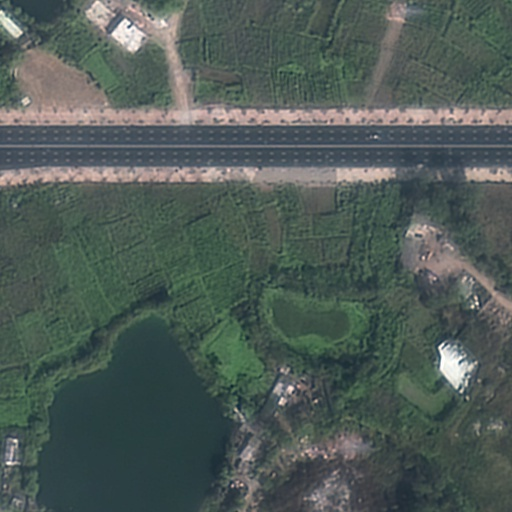}\vspace{2pt}
\includegraphics[width=1\linewidth]{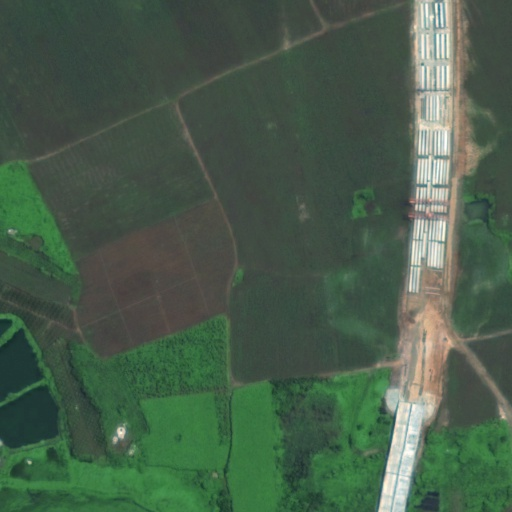}\vspace{2pt}
\end{minipage}}
\subfloat[GT]{
\begin{minipage}[t]{0.096\linewidth}
\includegraphics[width=1\linewidth]{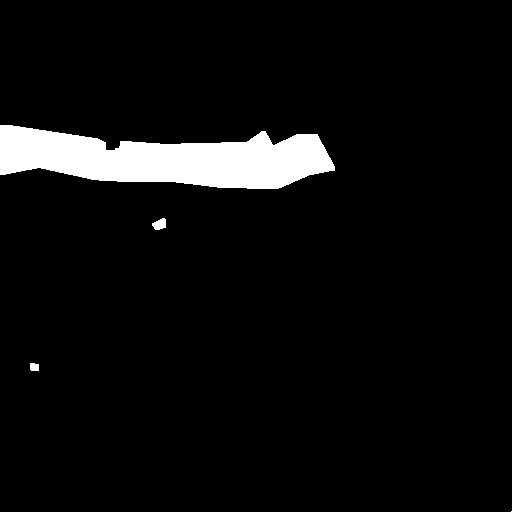}\vspace{2pt}
\includegraphics[width=1\linewidth]{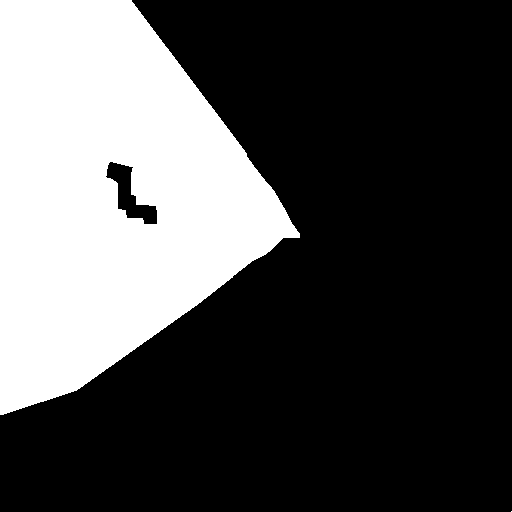}\vspace{2pt}
\includegraphics[width=1\linewidth]{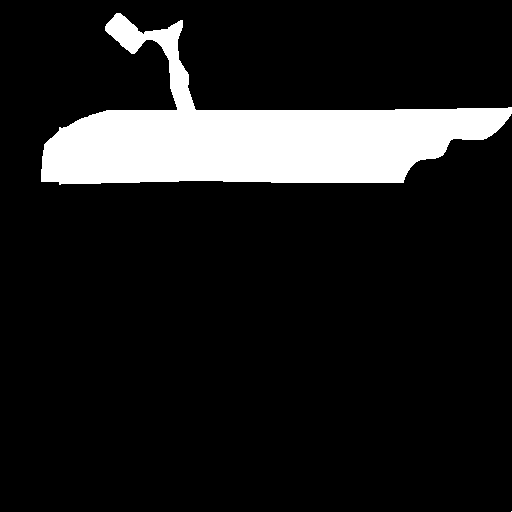}\vspace{2pt}
\includegraphics[width=1\linewidth]{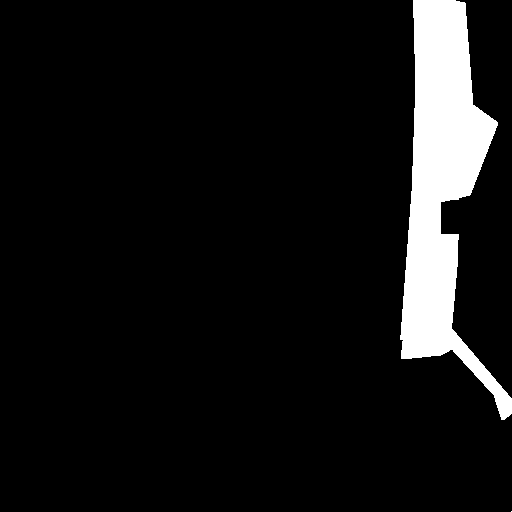}\vspace{2pt}
\end{minipage}}
\subfloat[SNUNet]{
\begin{minipage}[t]{0.096\linewidth}
\includegraphics[width=1\linewidth]{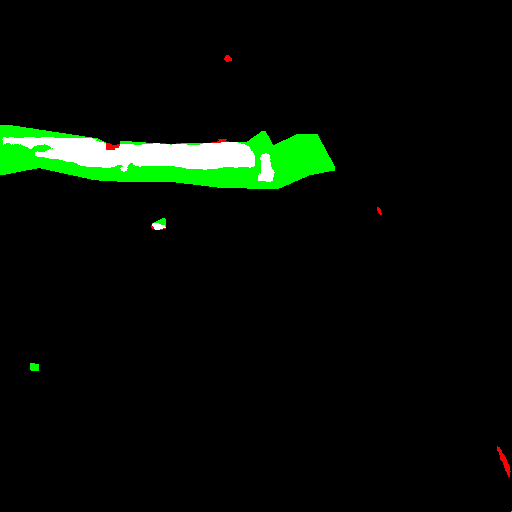}\vspace{2pt}
\includegraphics[width=1\linewidth]{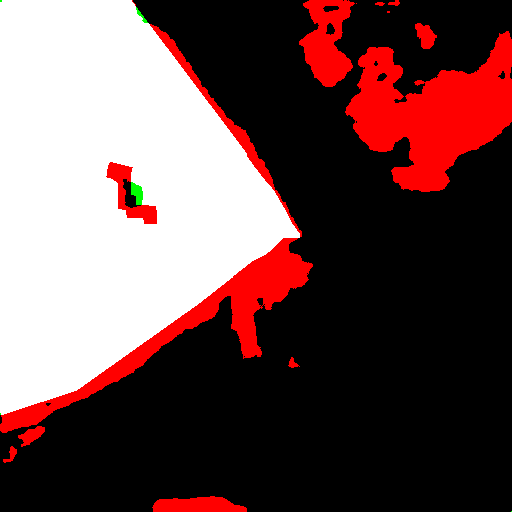}\vspace{2pt}
\includegraphics[width=1\linewidth]{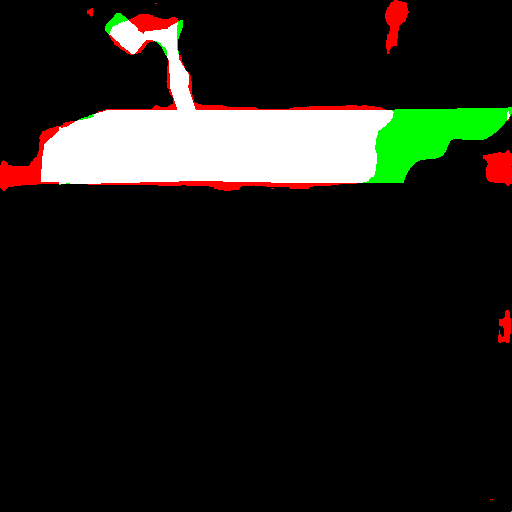}\vspace{2pt}
\includegraphics[width=1\linewidth]{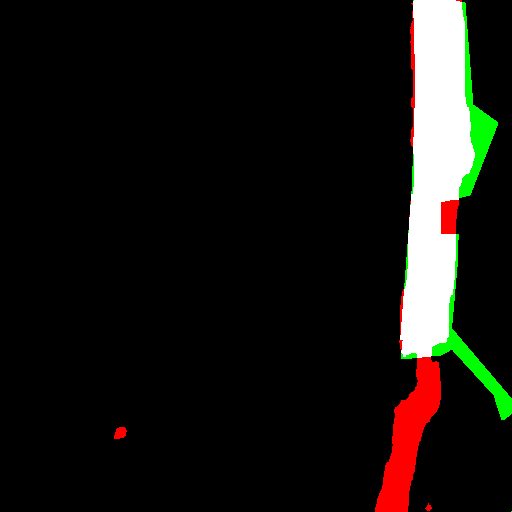}\vspace{2pt}
\end{minipage}}
\subfloat[BIT]{
\begin{minipage}[t]{0.096\linewidth}
\includegraphics[width=1\linewidth]{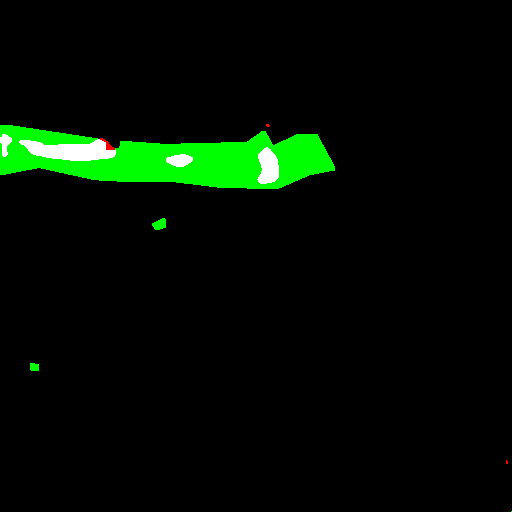}\vspace{2pt}
\includegraphics[width=1\linewidth]{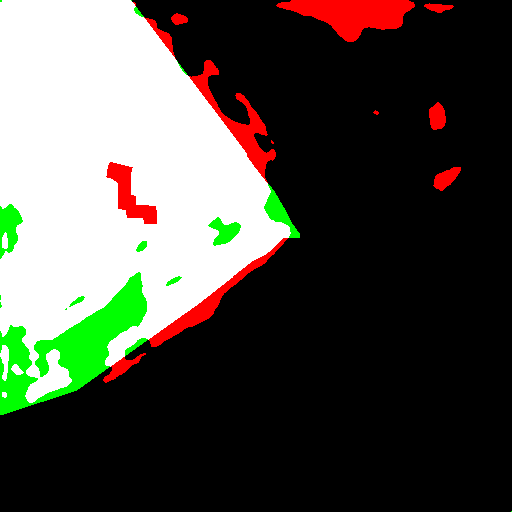}\vspace{2pt}
\includegraphics[width=1\linewidth]{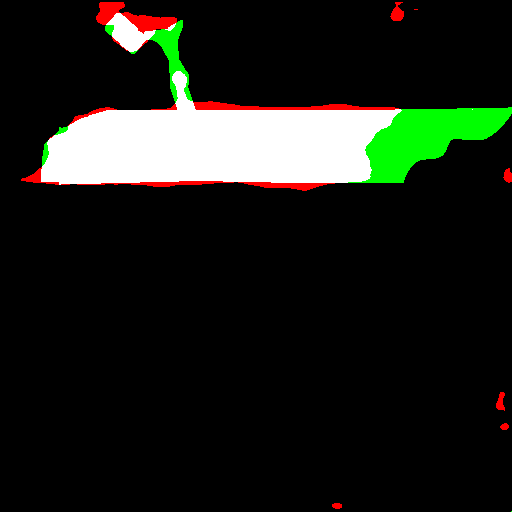}\vspace{2pt}
\includegraphics[width=1\linewidth]{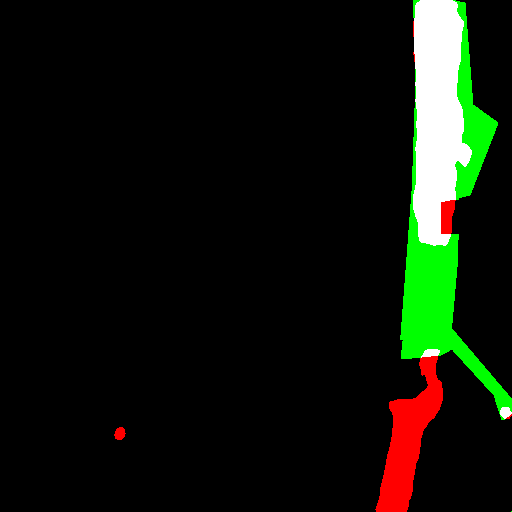}\vspace{2pt}
\end{minipage}}
\subfloat[SARASNet]{
\begin{minipage}[t]{0.096\linewidth}
\includegraphics[width=1\linewidth]{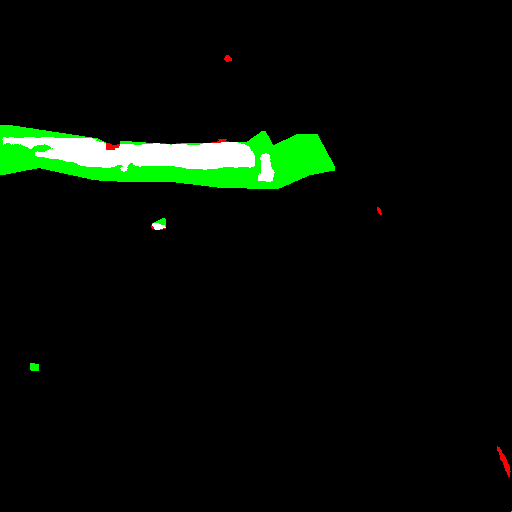}\vspace{2pt}
\includegraphics[width=1\linewidth]{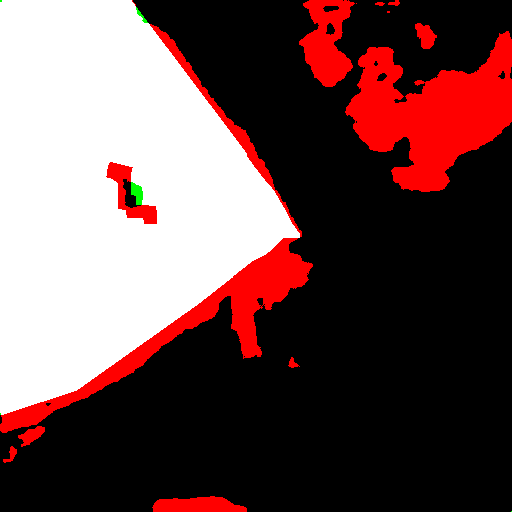}\vspace{2pt}
\includegraphics[width=1\linewidth]{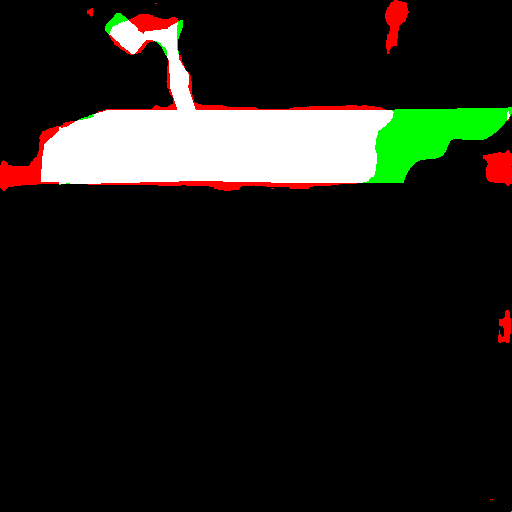}\vspace{2pt}
\includegraphics[width=1\linewidth]{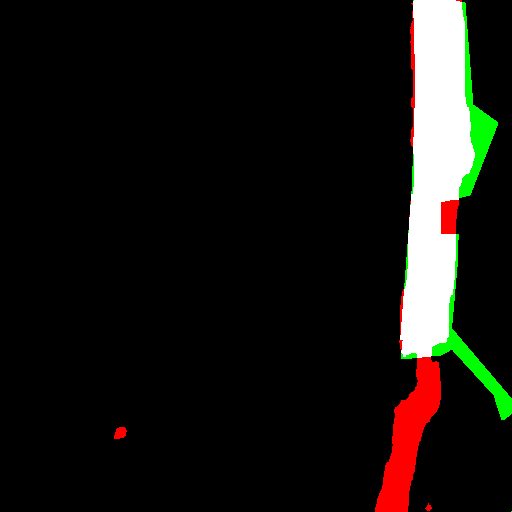}\vspace{2pt}
\end{minipage}}
\subfloat[AFCD3DNet]{
\begin{minipage}[t]{0.096\linewidth}
\includegraphics[width=1\linewidth]{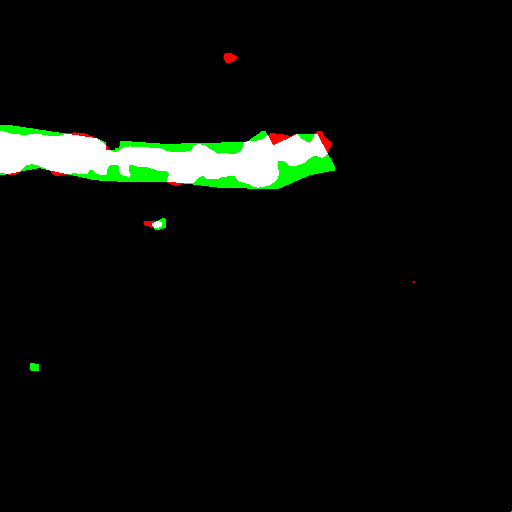}\vspace{2pt}
\includegraphics[width=1\linewidth]{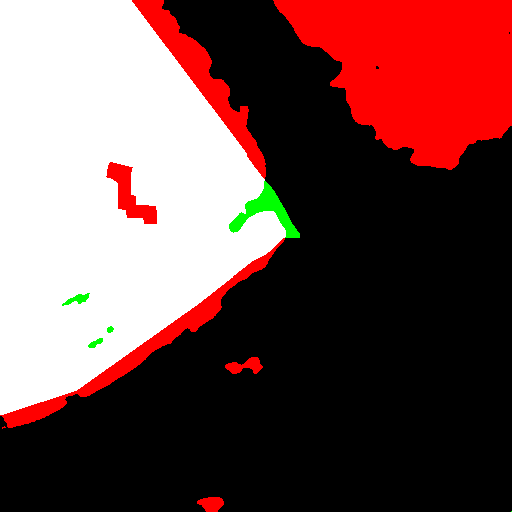}\vspace{2pt}
\includegraphics[width=1\linewidth]{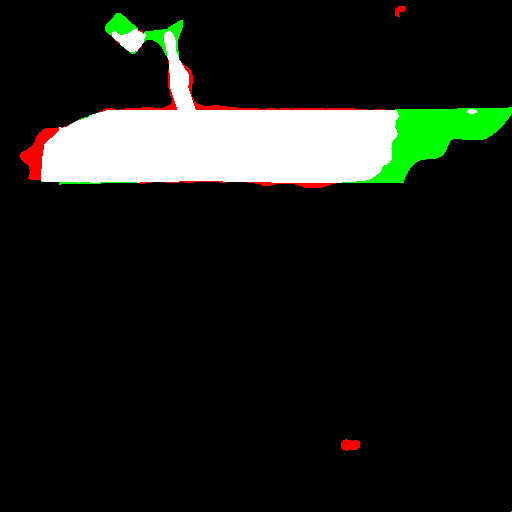}\vspace{2pt}
\includegraphics[width=1\linewidth]{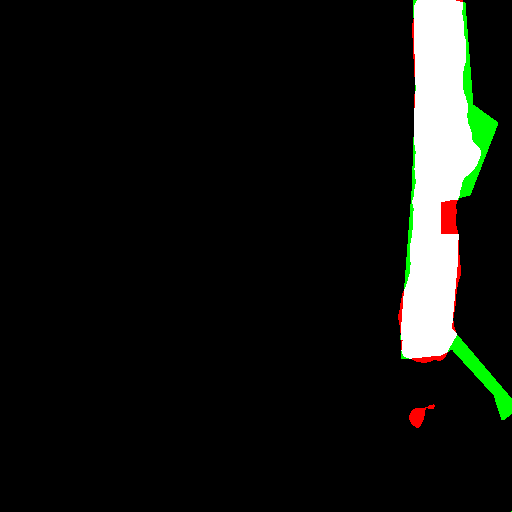}\vspace{2pt}
\end{minipage}}
\subfloat[RSMamba]{
\begin{minipage}[t]{0.096\linewidth}
\includegraphics[width=1\linewidth]{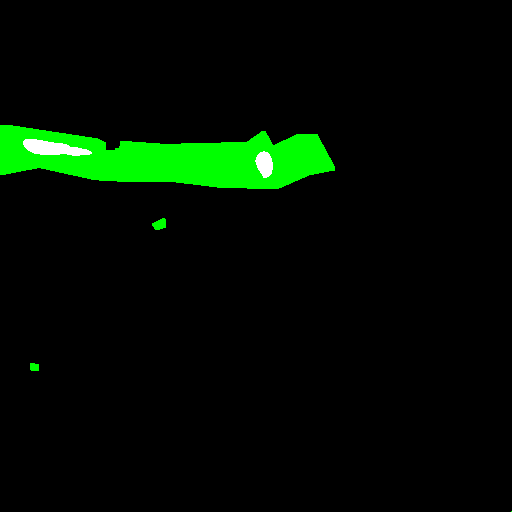}\vspace{2pt}
\includegraphics[width=1\linewidth]{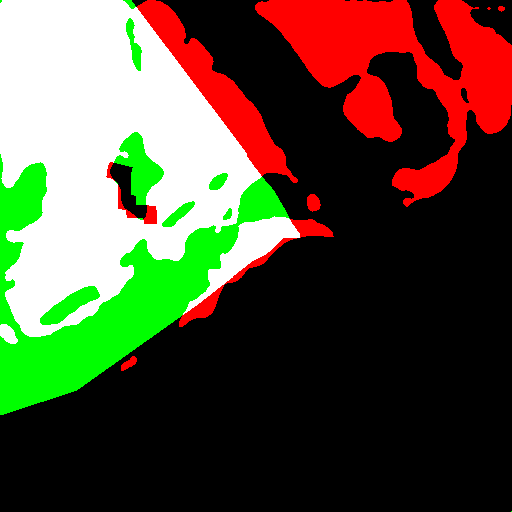}\vspace{2pt}
\includegraphics[width=1\linewidth]{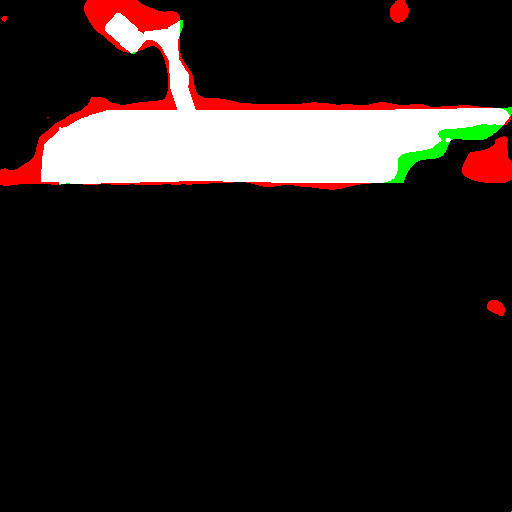}\vspace{2pt}
\includegraphics[width=1\linewidth]{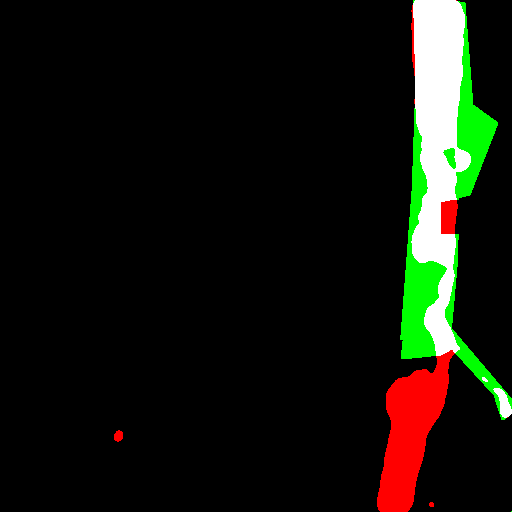}\vspace{2pt}
\end{minipage}}
\subfloat[ChangeMamba]{
\begin{minipage}[t]{0.096\linewidth}
\includegraphics[width=1\linewidth]{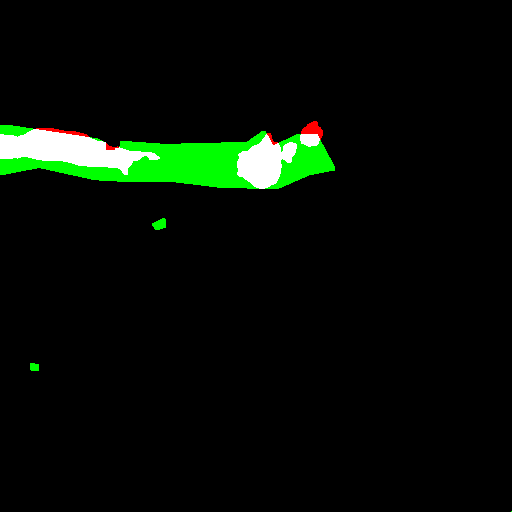}\vspace{2pt}
\includegraphics[width=1\linewidth]{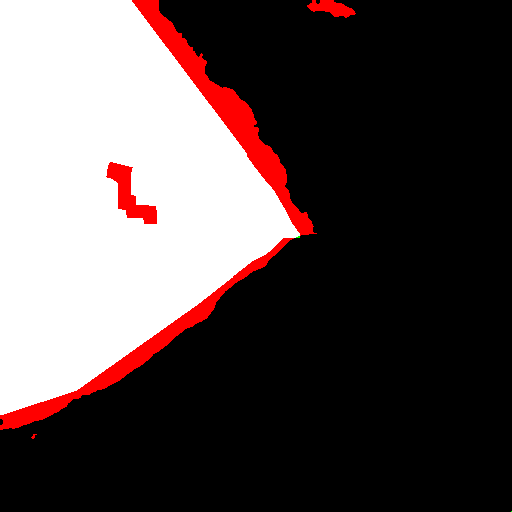}\vspace{2pt}
\includegraphics[width=1\linewidth]{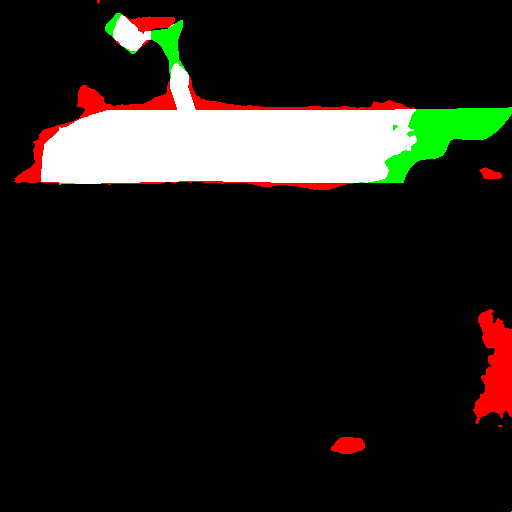}\vspace{2pt}
\includegraphics[width=1\linewidth]{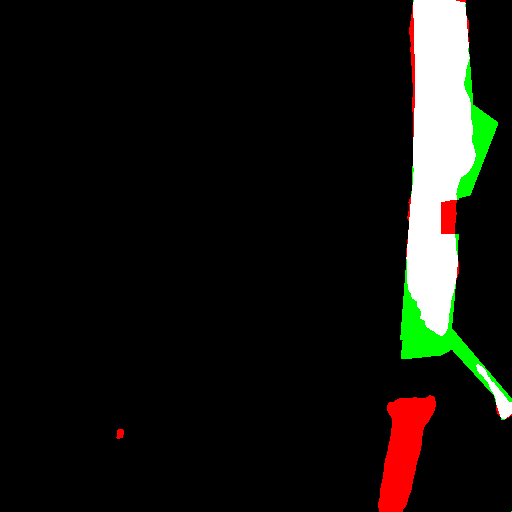}\vspace{2pt}
\end{minipage}}
\subfloat[CDLamba]{
\begin{minipage}[t]{0.096\linewidth}
\includegraphics[width=1\linewidth]{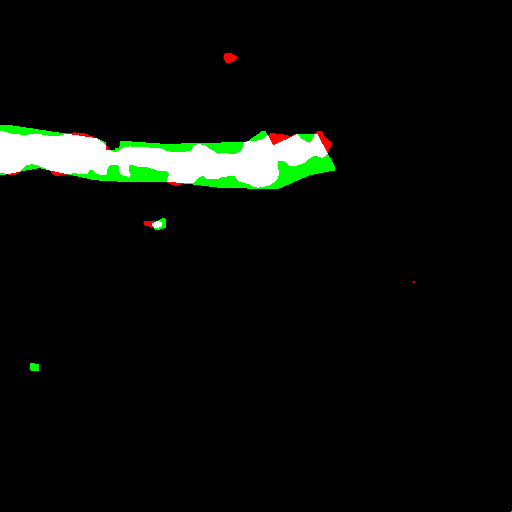}\vspace{2pt}
\includegraphics[width=1\linewidth]{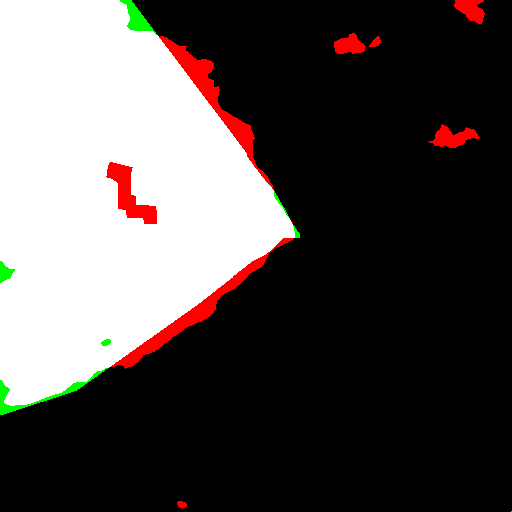}\vspace{2pt}
\includegraphics[width=1\linewidth]{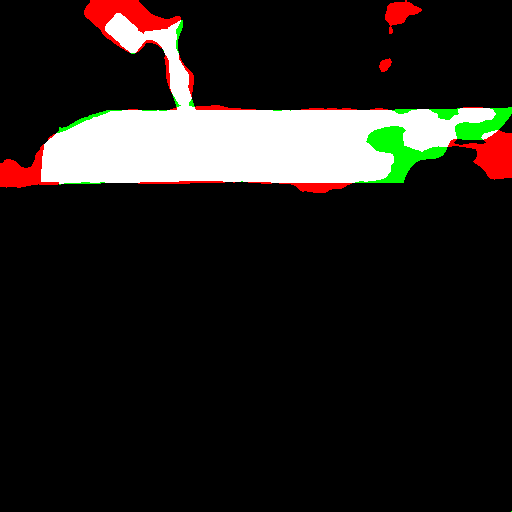}\vspace{2pt}
\includegraphics[width=1\linewidth]{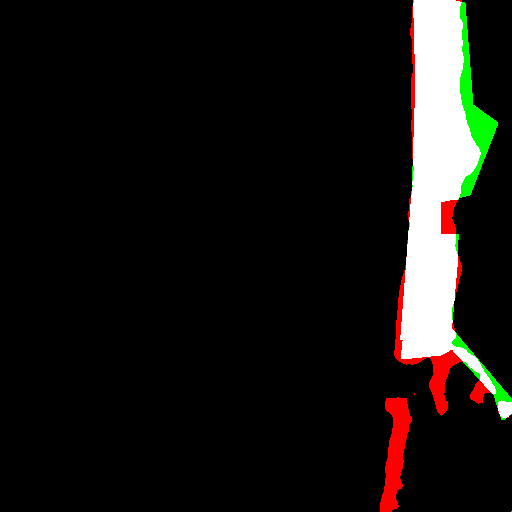}\vspace{2pt}
\end{minipage}}
\caption{Example results output from RSCD methods on test sets from CLCD dataset. Pixels are colored differently for better visualization (i.e., white for true positive, black for true negative, red for false positive, and green for false negative).}
\label{fig:clcd}
\end{figure*}

\subsection{Lightweight Change Detector}
\label{lcd}

Since our CT-LASS modules can modulate feature generation based on bi-temporal feature flow, our CD-Lamba inherently performs robust spatio-temporal context modeling. As a result, the bi-temporal features obtained at each spatial scale exhibit significant distinctions in regions of interest change. In this sense, to avoid introducing additional parameters and computational complexity to the model, we propose a lightweight change detector (LCD) to fuse multi-scale change representations and further derive the final change mask.

Specifically, our LCD takes the bi-temporal features \( \mathcal{F}_{i, 1}^o \) and \( \mathcal{F}_{i, 2}^o \) (\( i \in \{ 1,2,3,4 \} \)) outputted at four different scales as input. Then, the four change representations \( \mathcal{R}_z^i \) are obtained by performing element-wise subtraction between \( \mathcal{F}_{i, 1}^o \) and \( \mathcal{F}_{i, 2}^o \):

\begin{equation}
    \mathcal{R}_z^i = \mathcal{F}_{i, 1}^o \ominus \mathcal{F}_{i, 2}^o,
\end{equation}
where $\ominus$ denotes element-wise subtraction.

Next, these change representations \( \mathcal{R}_z^i \) are upsampled to the same resolution as the shallowest change representation \( \mathcal{R}_z^1 \), which, in our practical implementation, is set to one-fourth of the original image resolution, following \cite{changemamba}. Then, the representations \( \mathcal{R}_z^i \) are concatenated along the channel dimension to obtain the multi-scale change representations, denoted as \( \mathcal{R}_z \). Subsequently, a simple Multilayer Perceptron (MLP) is employed for feature integration $G$. Additionally, we incorporate depth convolution to enhance spatial details \cite{ren2022shunted}. The MLP module here consists of two simple $1 \times 1$ convolutions and the GeLU activation function. This process could be formulated as:
\begin{equation}
\mathcal{R}_z = \mathcal{R}_z^1 \oplus \mathcal{R}_z^2 \oplus \mathcal{R}_z^3 \oplus \mathcal{R}_z^4,
\end{equation}
\begin{equation}
G=\mathrm{MLP}\left( \mathcal{R}_z \right), 
\end{equation}
\begin{equation}
\hat{Y}=\mathrm{MLP}\left( G \oplus \mathrm{DWConv}\left( G \right) \right), 
\end{equation}
where $\oplus$ denotes channel-wise concatenation and $\hat{Y}$ represents the final change mask. 

Consequently, the change detector is performed in a lightweight way, significantly boosting the computational efficiency of the model.

\begin{figure*}[t]
\centering
\captionsetup[subfloat]{labelsep=none,format=plain,labelformat=empty}
\subfloat[$T_1$]{
\begin{minipage}[t]{0.13\linewidth}
\includegraphics[width=1\linewidth]{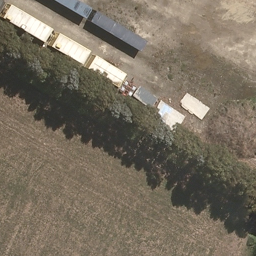}\vspace{2pt}
\includegraphics[width=1\linewidth]{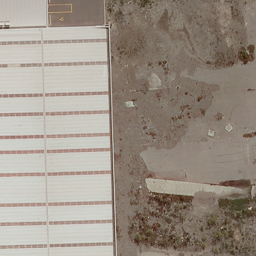}\vspace{2pt}
\includegraphics[width=1\linewidth]{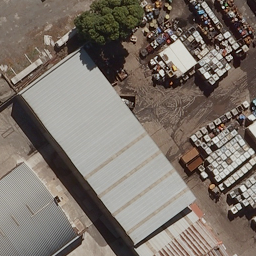}\vspace{2pt}
\includegraphics[width=1\linewidth]{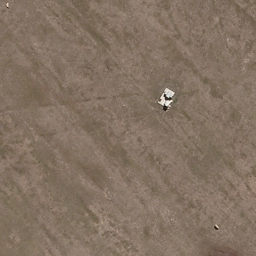}\vspace{2pt}
\end{minipage}}
\subfloat[$T_2$]{
\begin{minipage}[t]{0.13\linewidth}
\includegraphics[width=1\linewidth]{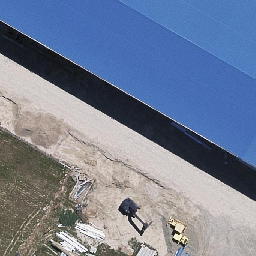}\vspace{2pt}
\includegraphics[width=1\linewidth]{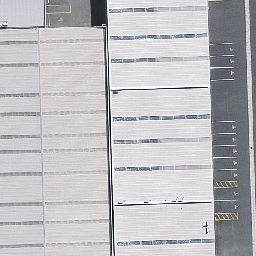}\vspace{2pt}
\includegraphics[width=1\linewidth]{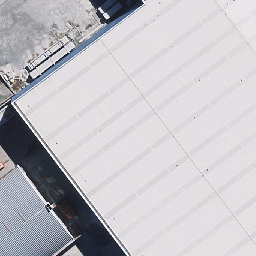}\vspace{2pt}
\includegraphics[width=1\linewidth]{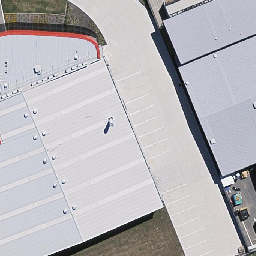}\vspace{2pt}
\end{minipage}}
\subfloat[GT]{
\begin{minipage}[t]{0.13\linewidth}
\includegraphics[width=1\linewidth]{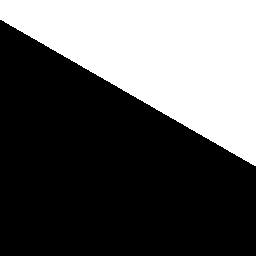}\vspace{2pt}
\includegraphics[width=1\linewidth]{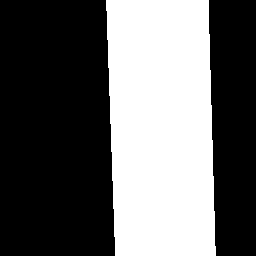}\vspace{2pt}
\includegraphics[width=1\linewidth]{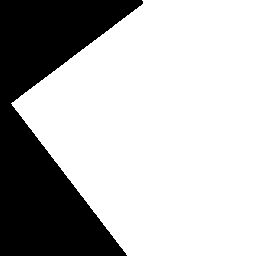}\vspace{2pt}
\includegraphics[width=1\linewidth]{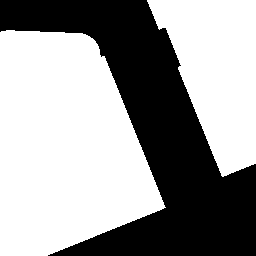}\vspace{2pt}
\end{minipage}}
\subfloat[$R_1$]{
\begin{minipage}[t]{0.13\linewidth}
\includegraphics[width=1\linewidth]{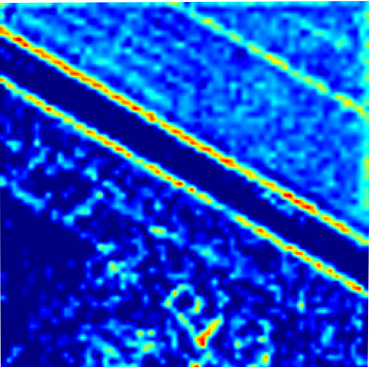}\vspace{2pt}
\includegraphics[width=1\linewidth]{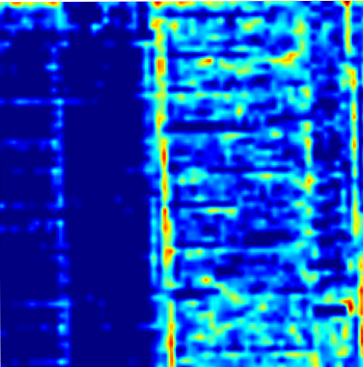}\vspace{2pt}
\includegraphics[width=1\linewidth]{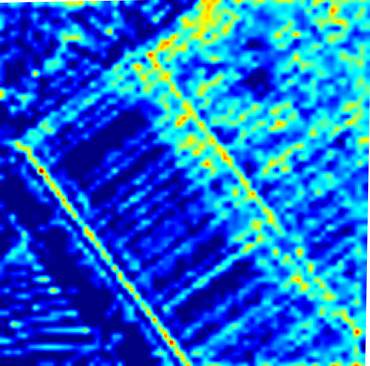}\vspace{2pt}
\includegraphics[width=1\linewidth]{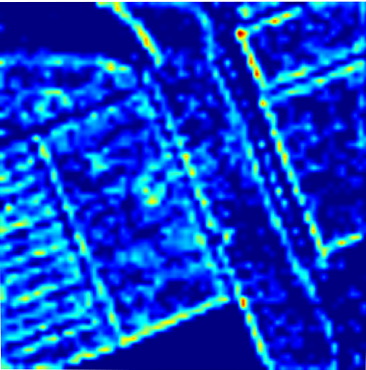}\vspace{2pt}
\end{minipage}}
\subfloat[$R_2$]{
\begin{minipage}[t]{0.13\linewidth}
\includegraphics[width=1\linewidth]{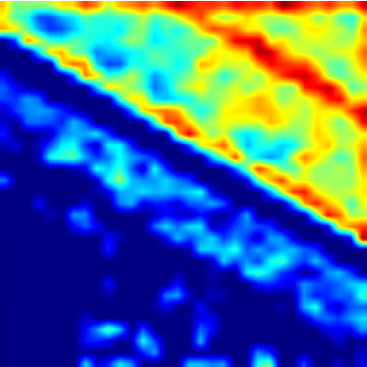}\vspace{2pt}
\includegraphics[width=1\linewidth]{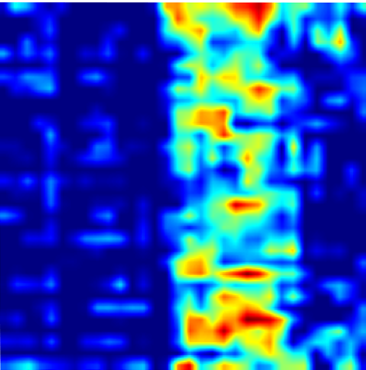}\vspace{2pt}
\includegraphics[width=1\linewidth]{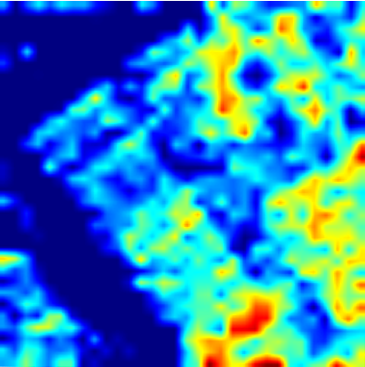}\vspace{2pt}
\includegraphics[width=1\linewidth]{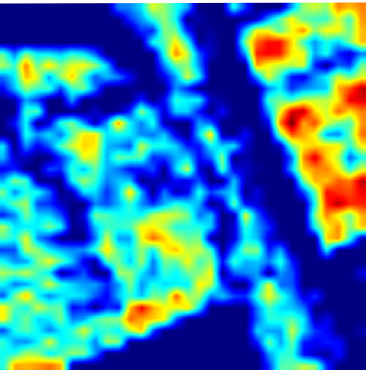}\vspace{2pt}
\end{minipage}}
\subfloat[$R_3$]{
\begin{minipage}[t]{0.13\linewidth}
\includegraphics[width=1\linewidth]{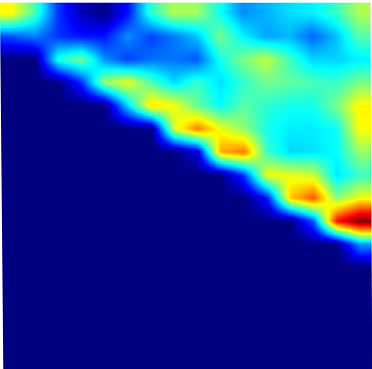}\vspace{2pt}
\includegraphics[width=1\linewidth]{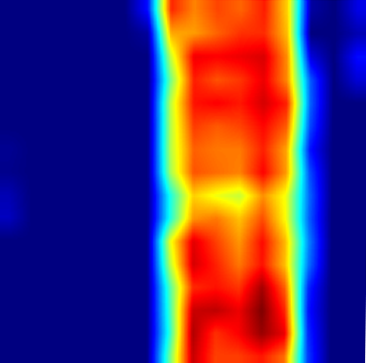}\vspace{2pt}
\includegraphics[width=1\linewidth]{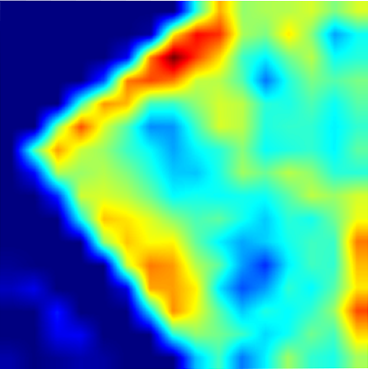}\vspace{2pt}
\includegraphics[width=1\linewidth]{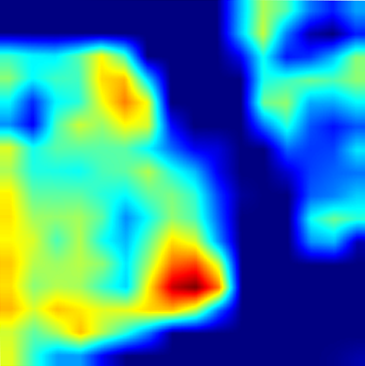}\vspace{2pt}
\end{minipage}}
\subfloat[$R_4$]{
\begin{minipage}[t]{0.13\linewidth}
\includegraphics[width=1\linewidth]{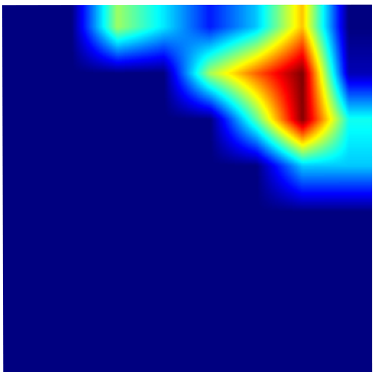}\vspace{2pt}
\includegraphics[width=1.02\linewidth]{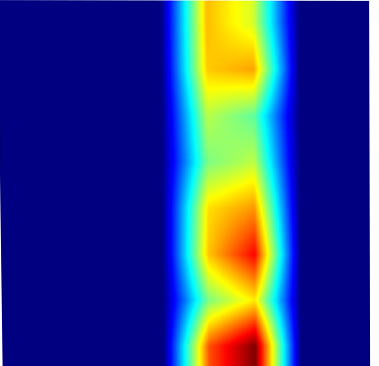}\vspace{2pt}
\includegraphics[width=1.02\linewidth]{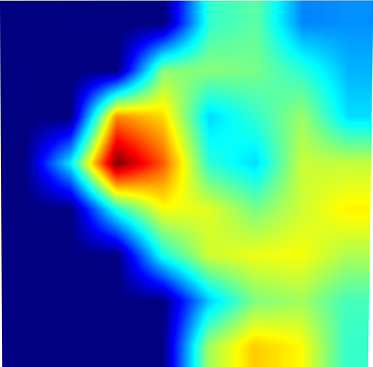}\vspace{2pt}
\includegraphics[width=1.02\linewidth]{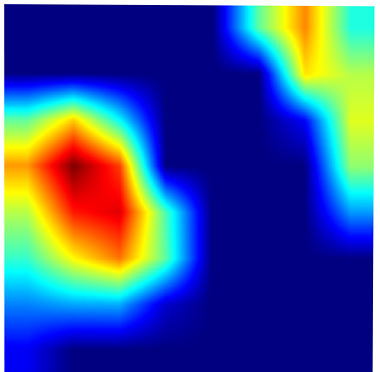}\vspace{2pt}
\end{minipage}}
\caption{Class activation maps for the change category of the features modulated by Cross-Temporal Locally Adaptive State-Space Scan (CT-LASS) module in different stages, which is based on grad-cam \cite{gradcam}. Example images are from the WHU-CD test set. $R_1$, $R_2$, $R_3$, and $R_4$ correspond to 1/4, 1/8, 1/16, and 1/32 resolution of the input image, respectively.}
\label{fig:gradcam_whucd}
\end{figure*}

\subsection{Loss Function}
\label{4.4}

Our CD-Mamba model is trained using a combination of binary cross-entropy loss and Dice loss~\cite{dice} to supervise the learning process of the change mask \( \hat{Y} \). The final loss function consists of both mask loss and classification loss, formulated as: 

\begin{equation}
\mathcal{L} = \lambda_{ce}\mathcal{L}_{ce} + \lambda_{dice}\mathcal{L}_{dice}. 
\end{equation}

The training process follows an end-to-end approach.

\section{Experiments}
\subsection{Experimental Settings}

\subsubsection{Datasets}
We evaluate our approach on four widely used publicly available datasets: WHU-CD~\cite{whu}, SYSU-CD~\cite{sysucd}, DSIFN-CD~\cite{dsifn}, and CLCD~\cite{clcd}, which are detailed below.

\textbf{WHU-CD~\cite{whu}} is a building CD dataset. It contains a pair of $32507 \times 15354$-sized dual-time aerial images with a spatial resolution of 0.075 m. Following previous work~\cite{BIT}, the paper crops the images into $256\times256$-sized blocks and randomly divides them into a training set (6096 images), a validation set (762 images) and a test set (762 images).

\textbf{SYSU-CD~\cite{sysucd}} is a high-resolution bi-temporal change detection dataset that includes 20,000 pairs of orthographic aerial images with a spatial size of $256 \times 256$ and a spatial resolution of 0.5 meters, primarily collected over Hong Kong. The dataset captures a diverse range of land-cover objects, including buildings, vessels, roads, and vegetation, offering a significant challenge for change detection tasks. For data distribution, the dataset is divided into a training set (12,000 images), a validation set (4,000 images), and a test set (4,000 images).

\textbf{DSIFN-CD~\cite{dsifn}} is a high-resolution bi-temporal CD dataset. It contains the change of multiple kinds of land-cover objects, such as roads, buildings, croplands, and water bodies. The paper follows the default cropped samples of size $512\times512$ provided by the authors and the default way to divide them into a training set (3600 images), a validation set (340 images) and a test set (48 images).

\textbf{CLCD~\cite{clcd}} is a cropland change detection dataset containing 600 pairs of remote sensing images of $512\times512$ size with spatial resolution ranging from 0.5 m to 2 m. The paper randomly divides it into a training set (360 images), a validation set (120 images) and a test set (120 images).

\begin{figure*}[t]
\centering
\captionsetup[subfloat]{labelsep=none,format=plain,labelformat=empty}
\subfloat[$T_1$]{
\begin{minipage}[t]{0.13\linewidth}
\includegraphics[width=1\linewidth]{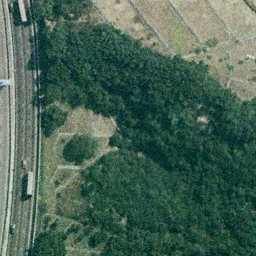}\vspace{2pt}
\includegraphics[width=1\linewidth]{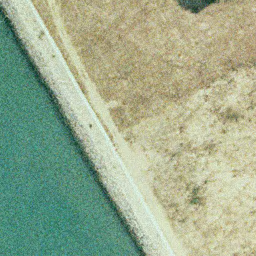}\vspace{2pt}
\includegraphics[width=1\linewidth]{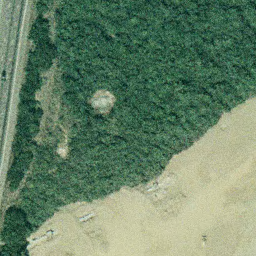}\vspace{2pt}
\includegraphics[width=1\linewidth]{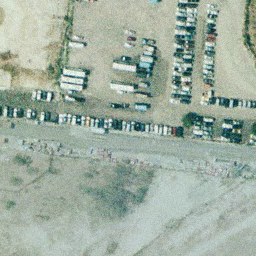}\vspace{2pt}
\end{minipage}}
\subfloat[$T_2$]{
\begin{minipage}[t]{0.13\linewidth}
\includegraphics[width=1\linewidth]{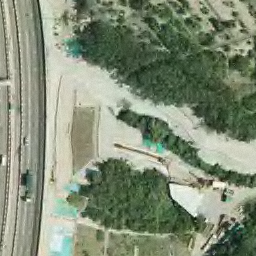}\vspace{2pt}
\includegraphics[width=1\linewidth]{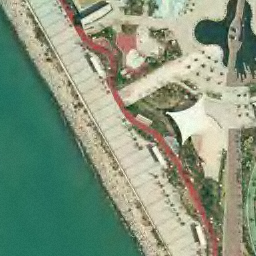}\vspace{2pt}
\includegraphics[width=1\linewidth]{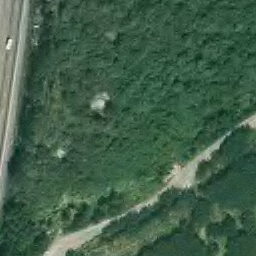}\vspace{2pt}
\includegraphics[width=1\linewidth]{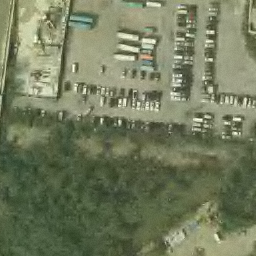}\vspace{2pt}
\end{minipage}}
\subfloat[GT]{
\begin{minipage}[t]{0.13\linewidth}
\includegraphics[width=1\linewidth]{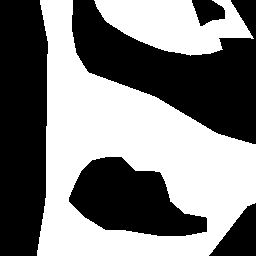}\vspace{2pt}
\includegraphics[width=1\linewidth]{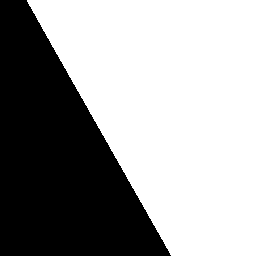}\vspace{2pt}
\includegraphics[width=1\linewidth]{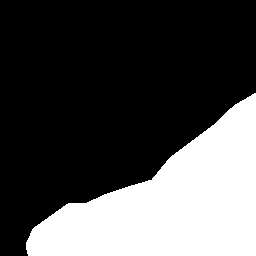}\vspace{2pt}
\includegraphics[width=1\linewidth]{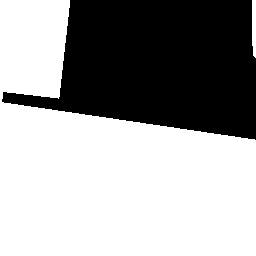}\vspace{2pt}
\end{minipage}}
\subfloat[$R_1$]{
\begin{minipage}[t]{0.13\linewidth}
\includegraphics[width=1\linewidth]{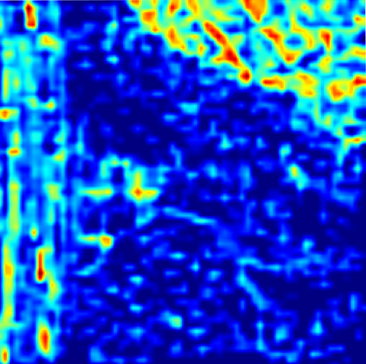}\vspace{2pt}
\includegraphics[width=1\linewidth]{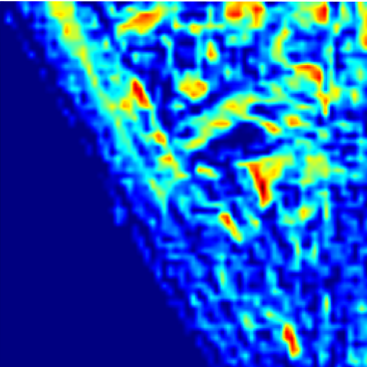}\vspace{2pt}
\includegraphics[width=1\linewidth]{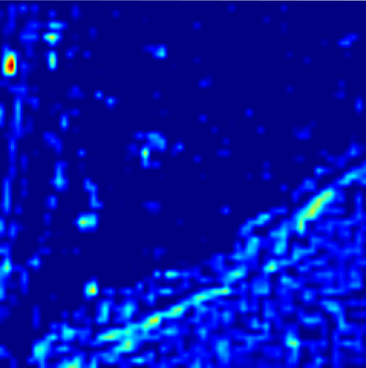}\vspace{2pt}
\includegraphics[width=1\linewidth]{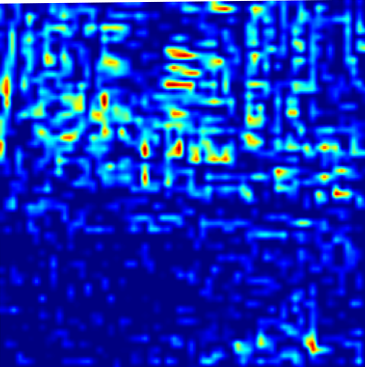}\vspace{2pt}
\end{minipage}}
\subfloat[$R_2$]{
\begin{minipage}[t]{0.13\linewidth}
\includegraphics[width=1\linewidth]{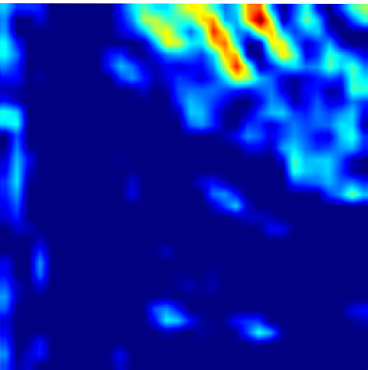}\vspace{2pt}
\includegraphics[width=1\linewidth]{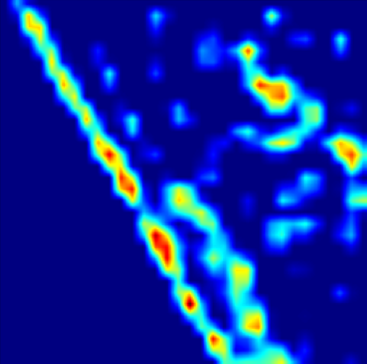}\vspace{2pt}
\includegraphics[width=1\linewidth]{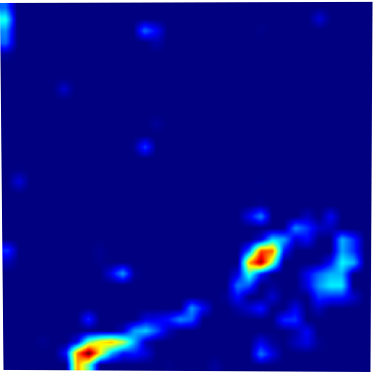}\vspace{2pt}
\includegraphics[width=1\linewidth]{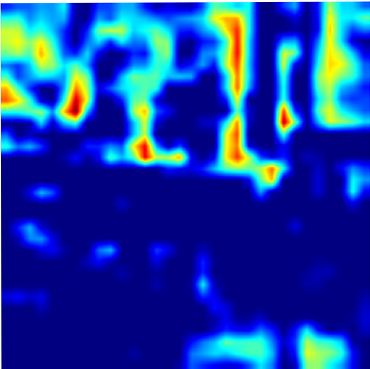}\vspace{2pt}
\end{minipage}}
\subfloat[$R_3$]{
\begin{minipage}[t]{0.13\linewidth}
\includegraphics[width=1\linewidth]{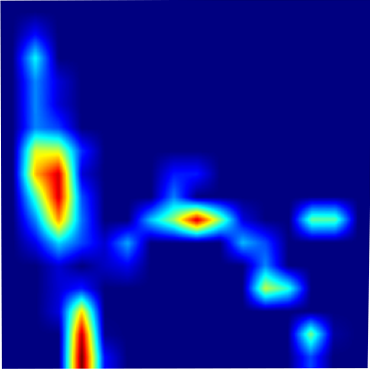}\vspace{2pt}
\includegraphics[width=1\linewidth]{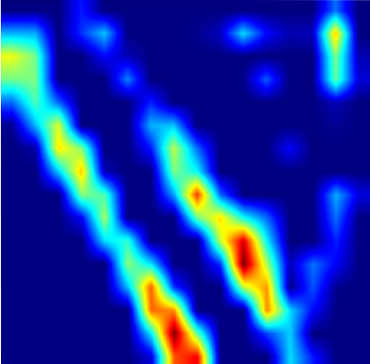}\vspace{2pt}
\includegraphics[width=1\linewidth]{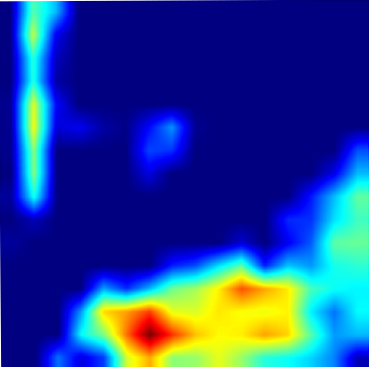}\vspace{2pt}
\includegraphics[width=1\linewidth]{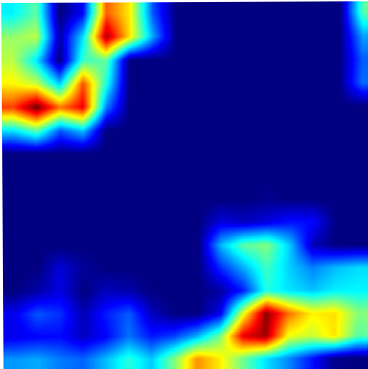}\vspace{2pt}
\end{minipage}}
\subfloat[$R_4$]{
\begin{minipage}[t]{0.13\linewidth}
\includegraphics[width=1\linewidth]{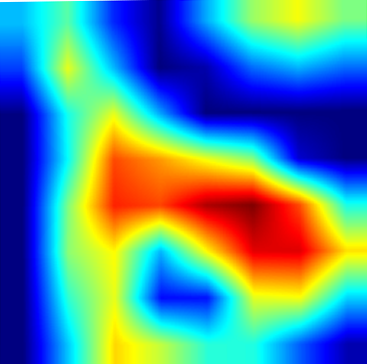}\vspace{2pt}
\includegraphics[width=1\linewidth]{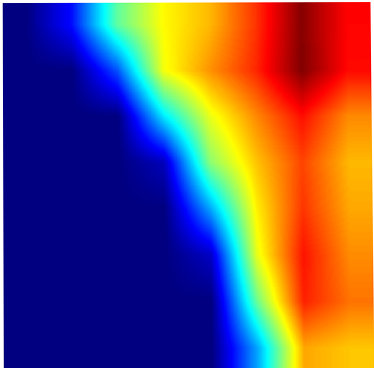}\vspace{2pt}
\includegraphics[width=1\linewidth]{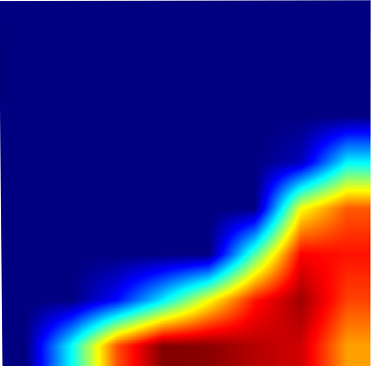}\vspace{2pt}
\includegraphics[width=1\linewidth]{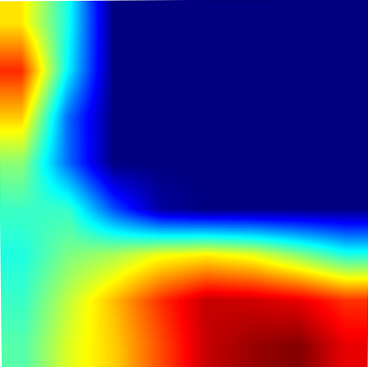}\vspace{2pt}
\end{minipage}}
\caption{Class activation maps for the change category of the features modulated by Cross-Temporal Locally Adaptive State-Space Scan (CT-LASS) module in different stages, which is based on grad-cam \cite{gradcam}. Example images are from the SYSU-CD test set. $R_1$, $R_2$, $R_3$, and $R_4$ correspond to 1/4, 1/8, 1/16, and 1/32 resolution of the input image, respectively.}
\label{fig:gradcam_sysucd}
\end{figure*}

\subsubsection{Evaluation Metrics}

We measure and report Precision (Pre.), Recall (Rec.), Intersection over Union (IoU), and Overall Accuracy (OA) of our model and competitors in terms of their predicted change categories on the test set of the employed datasets, respecitvely, which are defined as:
\begin{equation}
    \mathrm{Pre}=\frac{\mathrm{TP}}{\mathrm{TP}+\mathrm{FP}},
\end{equation}
\begin{equation}
    \mathrm{Rec}=\frac{\mathrm{TP}}{\mathrm{TP}+\mathrm{FN}},
\end{equation}
\begin{equation}
    \mathrm{IoU}=\frac{\mathrm{TP}}{\mathrm{TP}+\mathrm{FN}+\mathrm{FP}},
\end{equation}
\begin{equation}
    \mathrm{OA}=\frac{\mathrm{TP}+\mathrm{TN}}{\mathrm{TP}+\mathrm{TN}+\mathrm{FN}+\mathrm{FP}},
\end{equation}
where TP, TN, FP, and FN represent the number of true positive, true negative, false positive, and false negative, respectively.

We use the F1-score (F1) of the change category as the primary evaluation metric within the test set. The F1-score is derived from the precision and recall of the test set, calculated using the harmonic mean of these two metrics as follows:
\begin{equation}
    \mathrm{F1}=\frac{2 \times (\mathrm{Rec} \times \mathrm{Pre})}{\mathrm{Rec} + \mathrm{Pre}}.
\end{equation}

\subsubsection{Implementation Details}

We implement our CD-Lamba using Python based on PyTorch library, where a workstation with two NVIDIA GTX A6000 graphics cards and four NVIDIA GTX A5000 graphics cards (192~GB GPU memory in total) is employed. We initiated training with a learning rate of 0.01, employing the SGD optimizer which is configured with a momentum of 0.9 and a weight decay of 0.0005. During the training period, we employ a poly learning rate decay strategy with the power of $0.9$. The batch size is set to 8 for WHU-CD and SYSU-CD, while 4 for DSIFN-CD and CLCD. Additionally, we enhanced the training process by applying data augmentation techniques such as image flipping and blurring, aiming to improve model generalization and robustness.

\subsection{Main Results}
We compare the results with several state-of-the-art methods. CNN-based methods include FC-EF \cite{fc-siam}, FC-Siam-Di, FC-Siam-Conc, IFNet \cite{ifnet}, DTCDSCN \cite{dtcdscn}, SNUNet \cite{snunet}, ChangeStar(FarSeg \cite{farseg}) \cite{changestar}, LGPNet \cite{lgpnet}, USSFC-Net \cite{USSFCNet}, and AFCF3D-Net \cite{AFCF3DNet}. Transformer-based methods include DMATNet \cite{dmatnet}, BIT \cite{BIT}, ChangeFormer \cite{changeformer}, SARASNet \cite{SARASNet}. SSM-based methods include RS-Mamba \cite{rsmamba} and ChangeMamba \cite{changemamba}.

\begin{figure*}[t]
\centering
\captionsetup[subfloat]{labelsep=none,format=plain,labelformat=empty}
\subfloat[$T_1$]{
\begin{minipage}[t]{0.13\linewidth}
\includegraphics[width=1\linewidth]{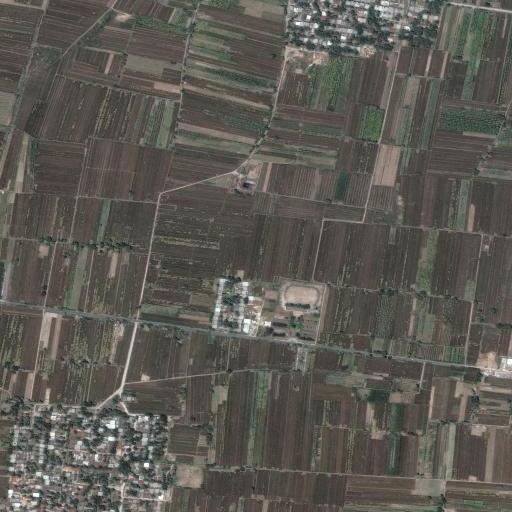}\vspace{2pt}
\includegraphics[width=1\linewidth]{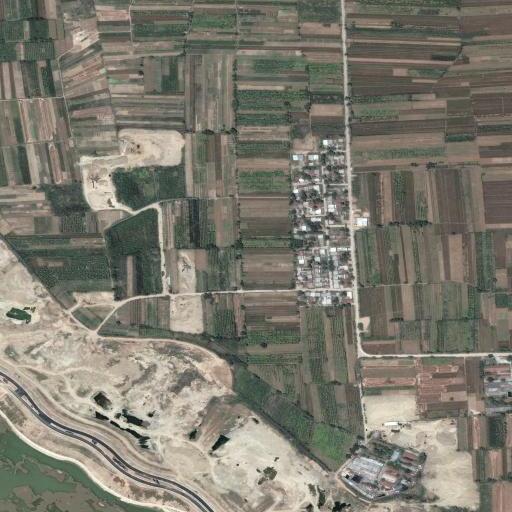}\vspace{2pt}
\includegraphics[width=1\linewidth]{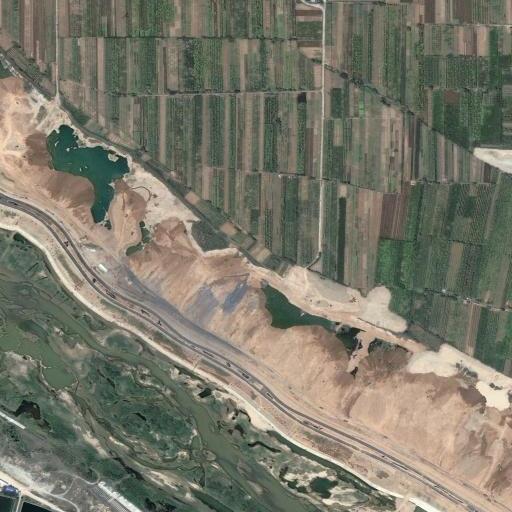}\vspace{2pt}
\includegraphics[width=1\linewidth]{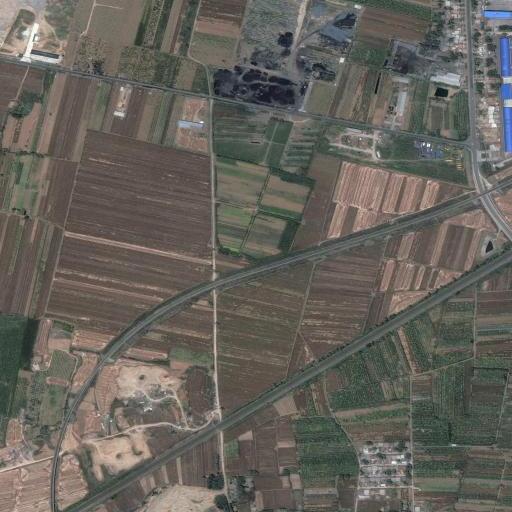}\vspace{2pt}
\end{minipage}}
\subfloat[$T_2$]{
\begin{minipage}[t]{0.13\linewidth}
\includegraphics[width=1\linewidth]{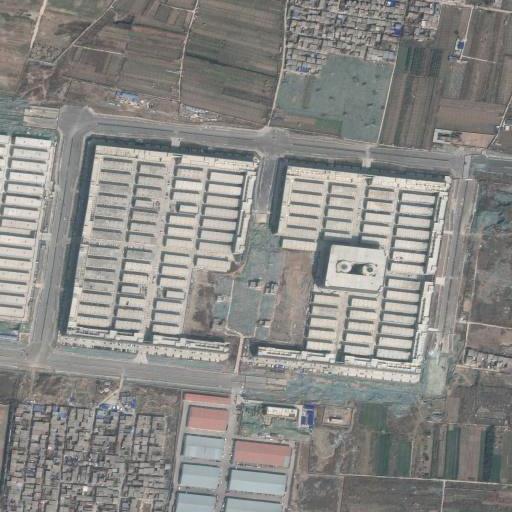}\vspace{2pt}
\includegraphics[width=1\linewidth]{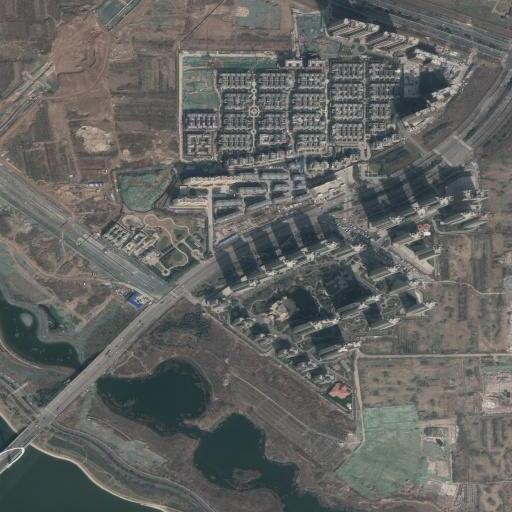}\vspace{2pt}
\includegraphics[width=1\linewidth]{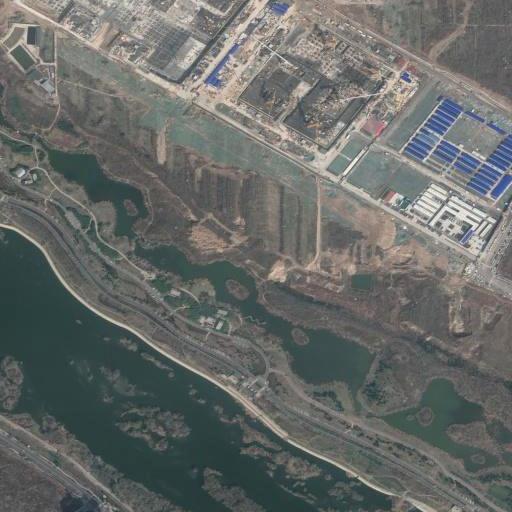}\vspace{2pt}
\includegraphics[width=1\linewidth]{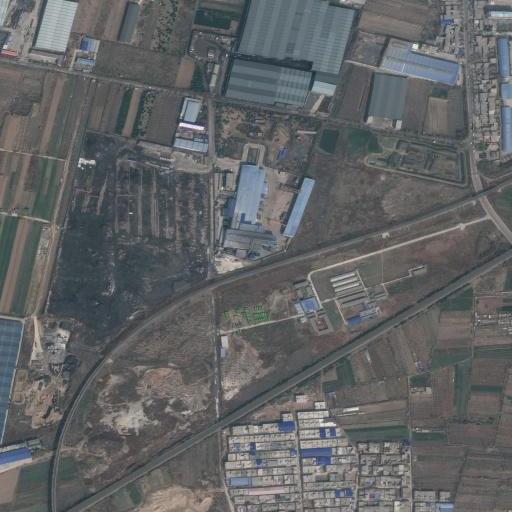}\vspace{2pt}
\end{minipage}}
\subfloat[GT]{
\begin{minipage}[t]{0.13\linewidth}
\includegraphics[width=1\linewidth]{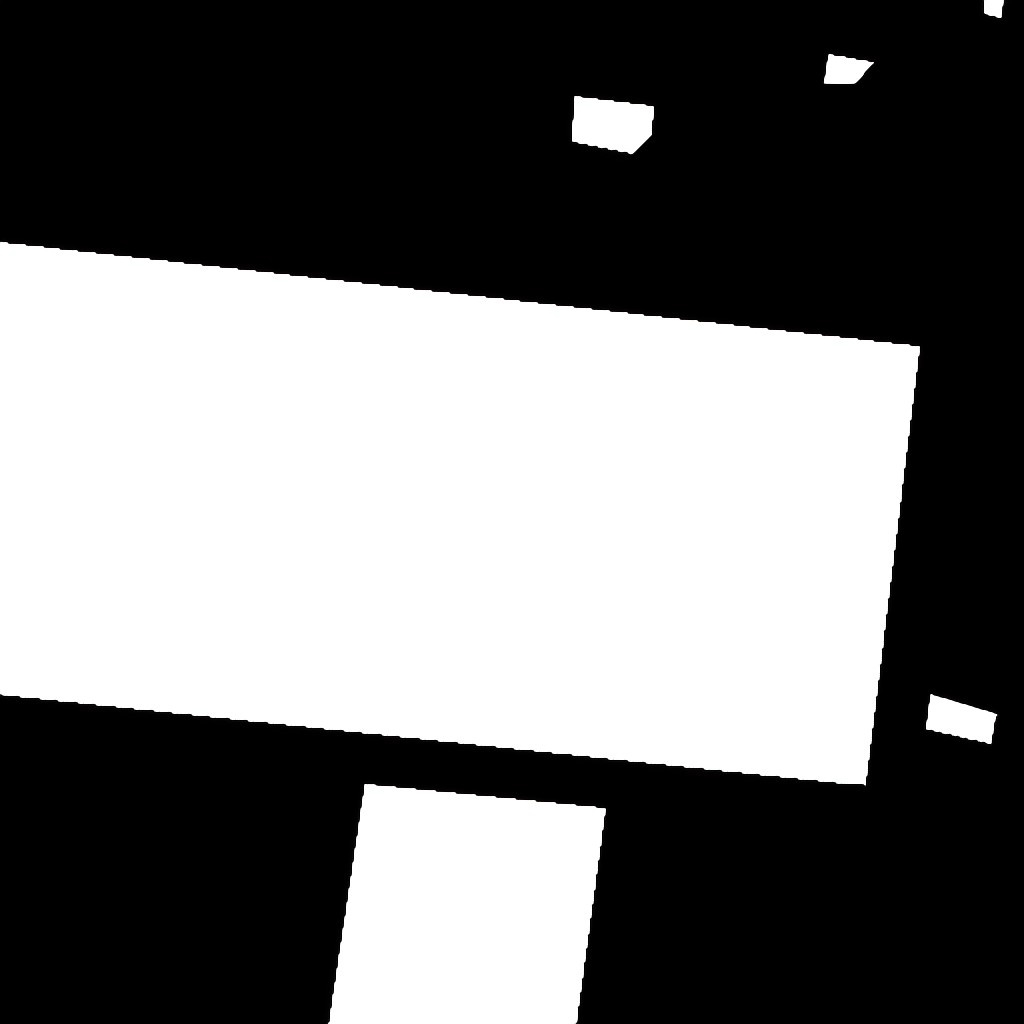}\vspace{2pt}
\includegraphics[width=1\linewidth]{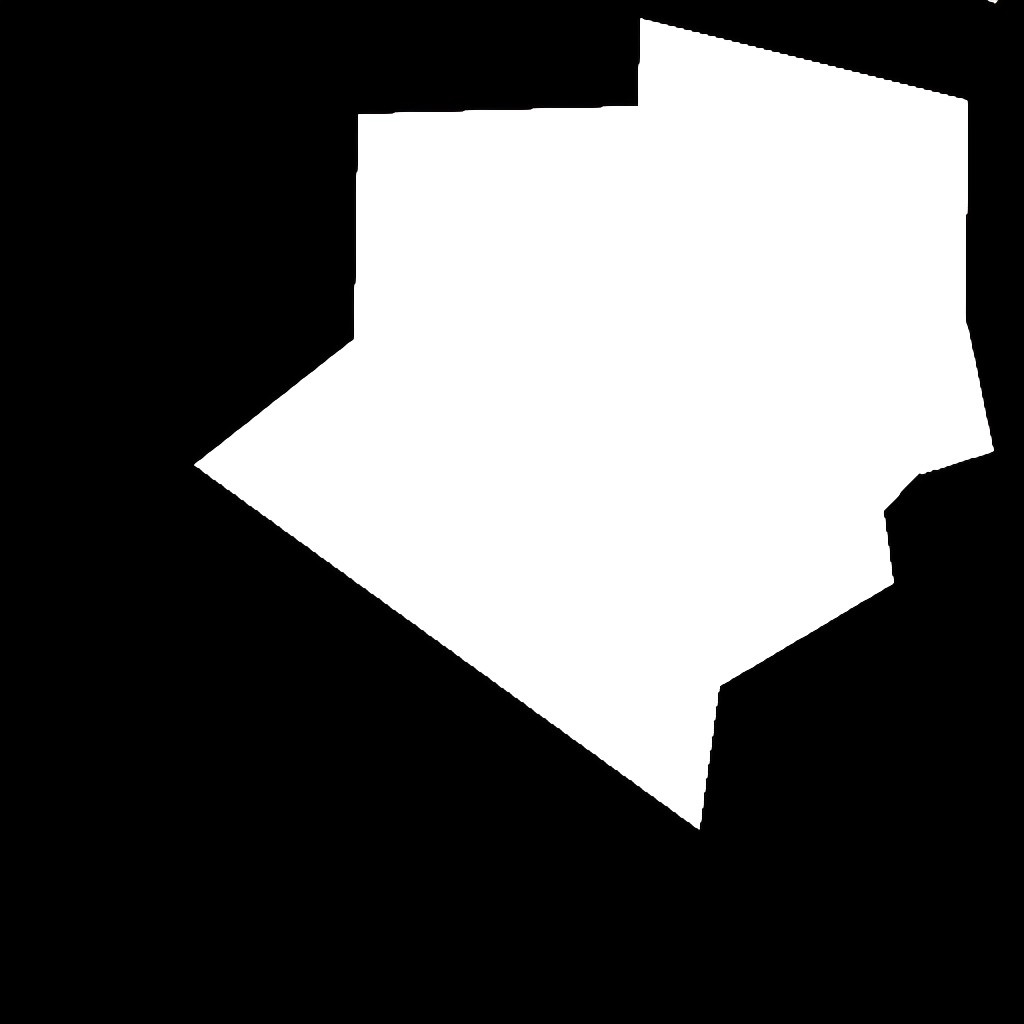}\vspace{2pt}
\includegraphics[width=1\linewidth]{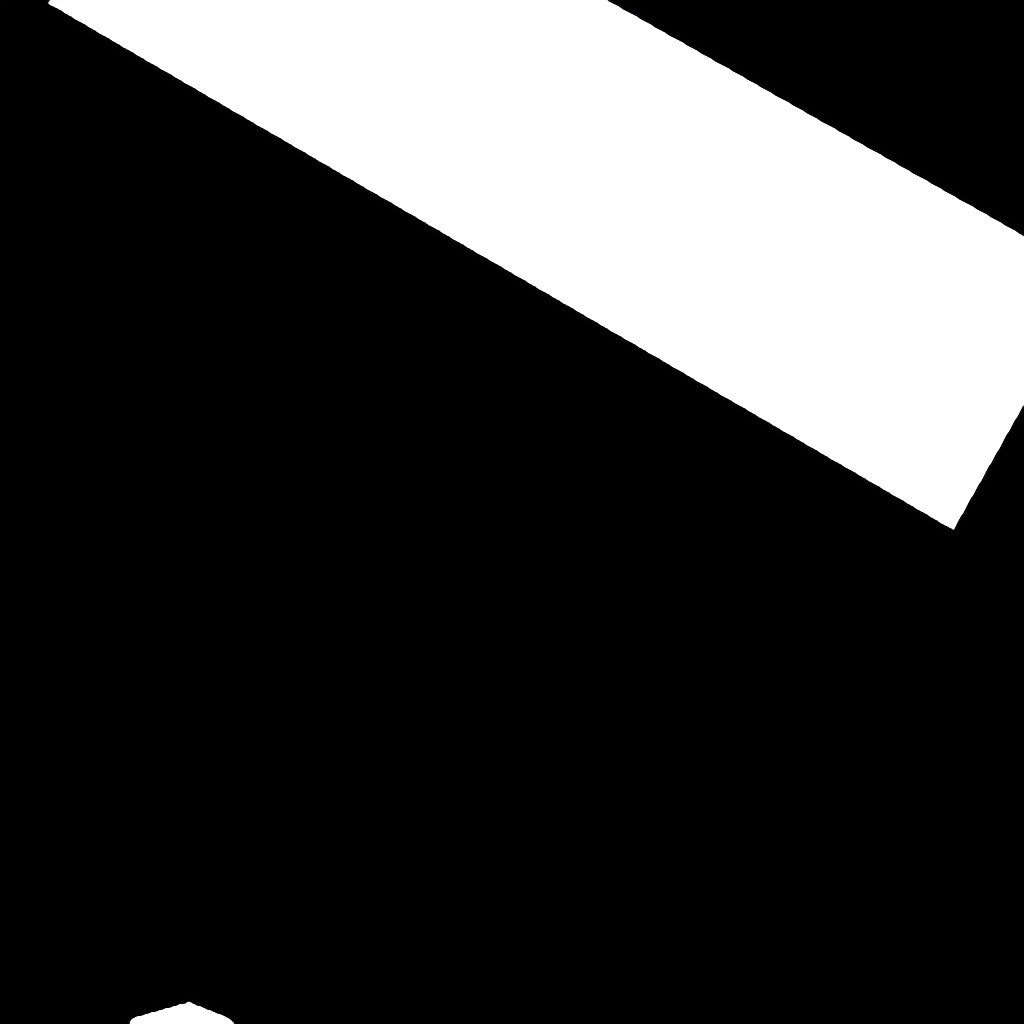}\vspace{2pt}
\includegraphics[width=1\linewidth]{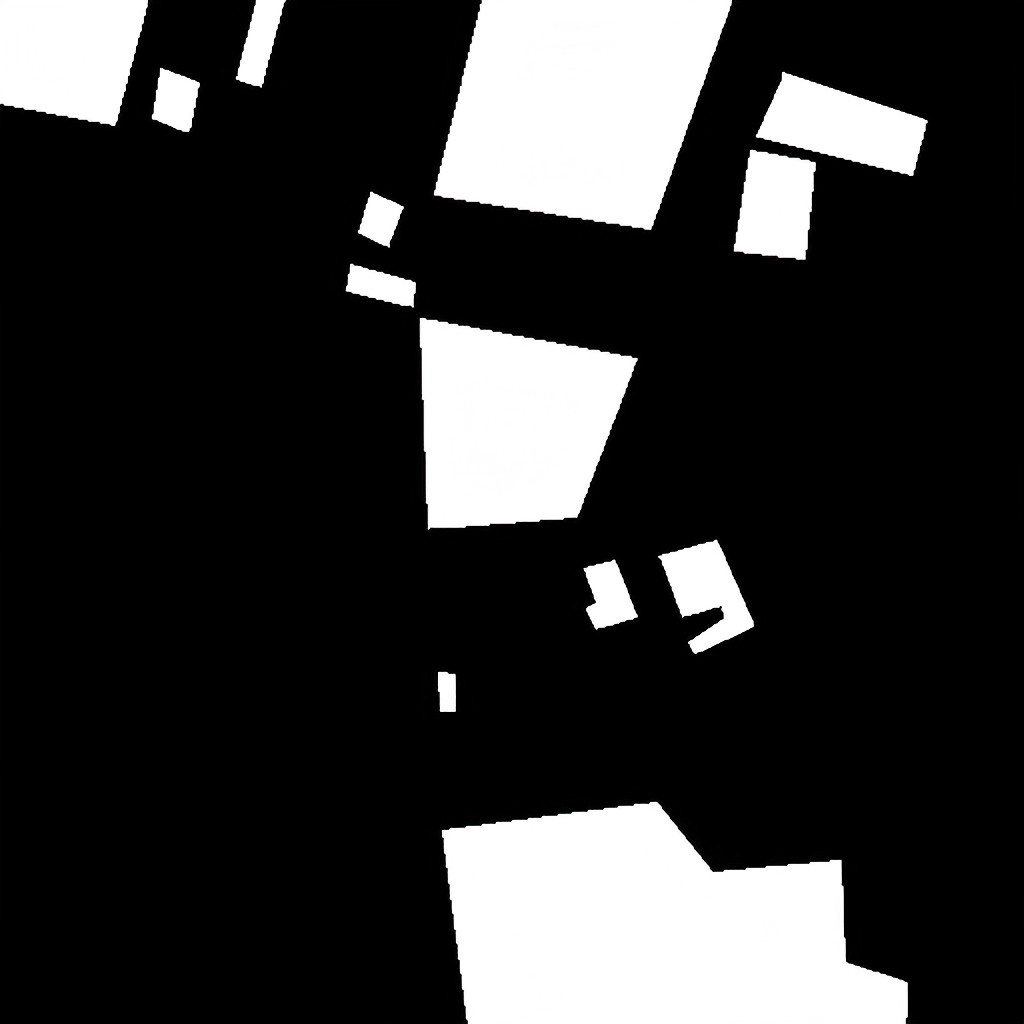}\vspace{2pt}
\end{minipage}}
\subfloat[$R_1$]{
\begin{minipage}[t]{0.13\linewidth}
\includegraphics[width=1\linewidth]{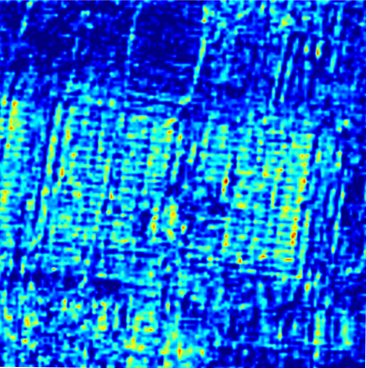}\vspace{2pt}
\includegraphics[width=1\linewidth]{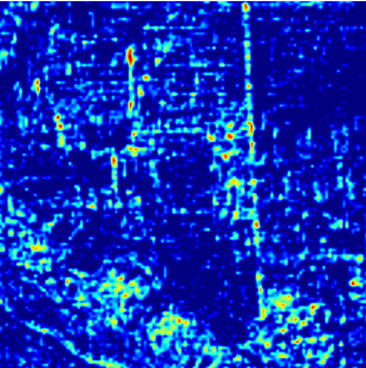}\vspace{2pt}
\includegraphics[width=1\linewidth]{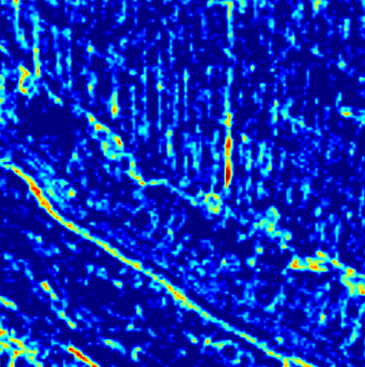}\vspace{2pt}
\includegraphics[width=1\linewidth]{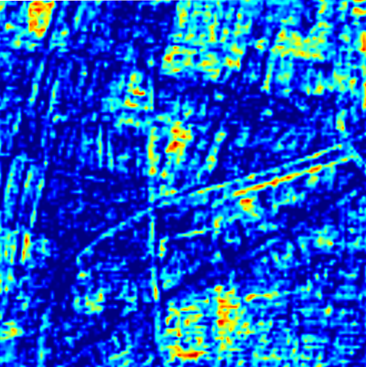}\vspace{2pt}
\end{minipage}}
\subfloat[$R_2$]{
\begin{minipage}[t]{0.13\linewidth}
\includegraphics[width=1\linewidth]{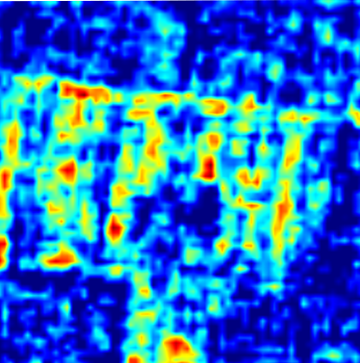}\vspace{2pt}
\includegraphics[width=1\linewidth]{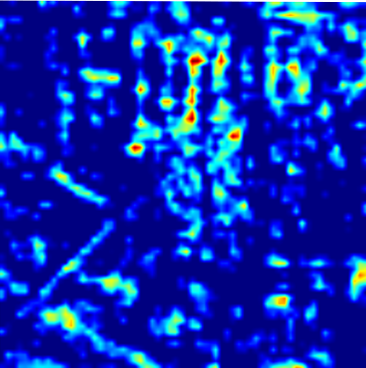}\vspace{2pt}
\includegraphics[width=1\linewidth]{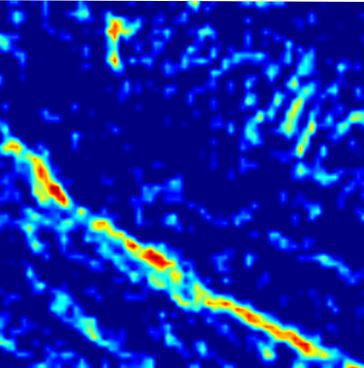}\vspace{2pt}
\includegraphics[width=1\linewidth]{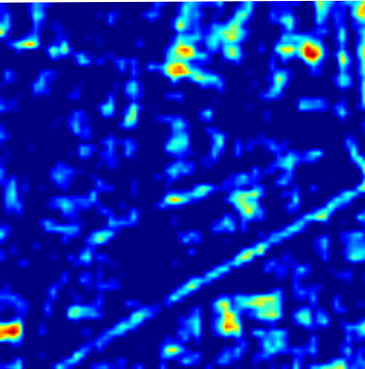}\vspace{2pt}
\end{minipage}}
\subfloat[$R_3$]{
\begin{minipage}[t]{0.13\linewidth}
\includegraphics[width=1.02\linewidth]{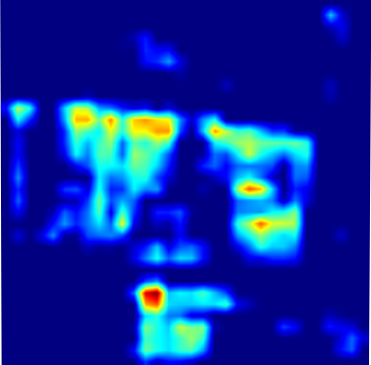}\vspace{2pt}
\includegraphics[width=1.02\linewidth]{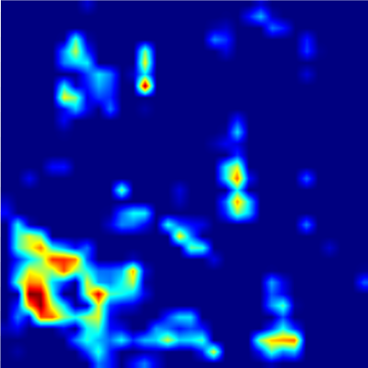}\vspace{2pt}
\includegraphics[width=1.02\linewidth]{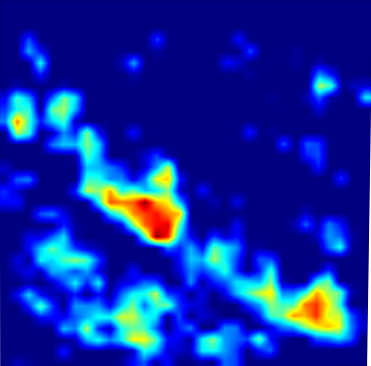}\vspace{2pt}
\includegraphics[width=1.02\linewidth]{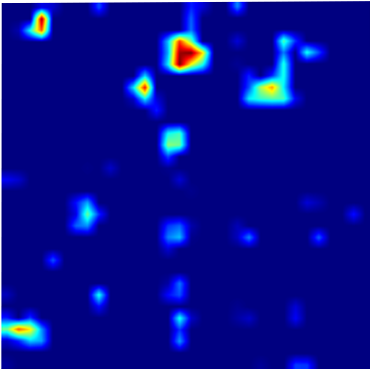}\vspace{2pt}
\end{minipage}}
\subfloat[$R_4$]{
\begin{minipage}[t]{0.13\linewidth}
\includegraphics[width=1\linewidth]{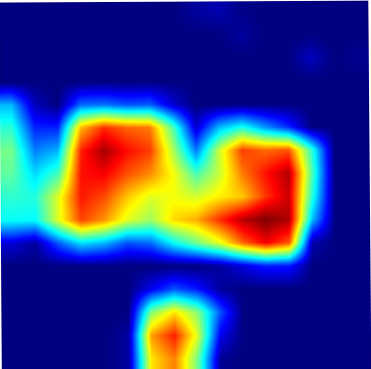}\vspace{2pt}
\includegraphics[width=1\linewidth]{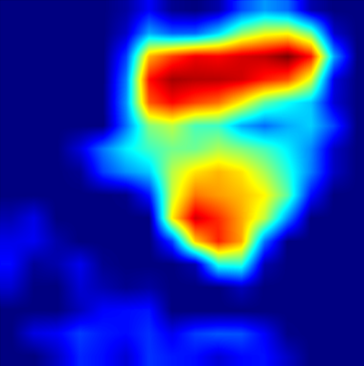}\vspace{2pt}
\includegraphics[width=1\linewidth]{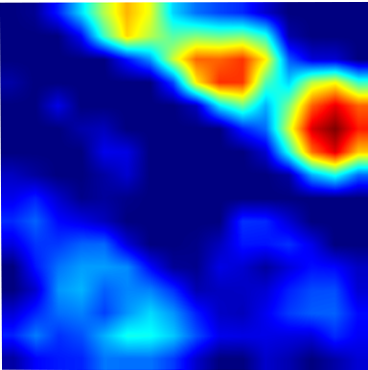}\vspace{2pt}
\includegraphics[width=1\linewidth]{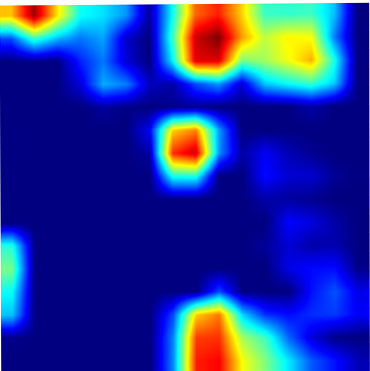}\vspace{2pt}
\end{minipage}}
\caption{Class activation maps for the change category of the features modulated by Cross-Temporal Locally Adaptive State-Space Scan (CT-LASS) module in different stages, which is based on grad-cam \cite{gradcam}. Example images are from the DSIFN-CD test set. $R_1$, $R_2$, $R_3$, and $R_4$ correspond to 1/4, 1/8, 1/16, and 1/32 resolution of the input image, respectively.}
\label{fig:gradcam_dsifn}
\end{figure*}

\subsubsection{Qualitative Analysis}
The experimental results are shown in Tables~\ref{table:2} and~\ref{table:3}. It can be observed that the proposed CD-Lamba achieves state-of-the-art performance on the five change detection datasets. Specifically, CD-Lamba achieves 2.43\%, 3.28\%, 5.75\% and 8.06\% F1 metrics enhancement on WHU-CD, SYSU-CD, DSIFN-CD, and CLCD datasets, respectively, compared to the recent SSM-based method ChangeMamba. From this, it can be seen that CD-Lamba can achieve more significant improvements in more complex scenarios, more diverse object distributions, and richer variations (e.g., CLCD and DSIFN-CD datasets). In addition, CD-Lamba only has 28.74M parameters and 15.26G Flops, which requires only 59.18\% of parameter requirements and 39.65\% of computational consumption compared to ChangeMamba. Compared to other models such as AFCF3D-Net, SARASNet and ChangeFormer, CD-Lamba improves more significantly and has a better balance between performance and efficiency.

\subsubsection{Qualitative Visualization}
We also perform visual comparisons on the four RSCD datasets. As shown in Fig. \ref{fig:whucd} (WHU-CD), Fig. \ref{fig:sysucd} (SYSU-CD), Fig. \ref{fig:dsifn} (DSIFN-CD), and Fig. \ref{fig:clcd} (CLCD), the proposed CD-Lamba demonstrates better visualization performance compared to previous state-of-the-art methods such as SNUNet, BIT, SARASNet, AFCF3DNet, RS-Mamba, and ChangeMamba.
Specifically, our CD-Lamba appears to have fewer false alarms (e.g., the first row of Fig. \ref{fig:whucd}, the fourth row of Fig. \ref{fig:sysucd}, and the first row of Fig. \ref{fig:clcd}) and omission alarms (e.g., the third row of Fig. \ref{fig:sysucd} and the second row of Fig. \ref{fig:clcd}). In addition, CD-Lamba outputs masks with sharper boundaries (e.g., the fourth row of Fig. \ref{fig:whucd} and the third row of Fig. \ref{fig:clcd}) as well as more complete topological shapes (e.g., the second row of Fig. \ref{fig:whucd}, the fourth row of Fig. \ref{fig:dsifn}, and the fourth row of Fig. \ref{fig:clcd}). It is also worth noting that CD-Lamba, due to its ability to enhance spatial locality details, can identify changes that other SSM-based methods (e.g., RS-Mamba and ChangeMamba) completely fail to detect (e.g., the first and the third rows of Fig. \ref{fig:whucd}). These visual comparisons further validate the capabilities of our CD-Lamba in effectively enhancing changes of interest and capturing detailed regions of change across a variety of complex scenarios.

\begin{figure*}[t]
\centering
\captionsetup[subfloat]{labelsep=none,format=plain,labelformat=empty}
\subfloat[$T_1$]{
\begin{minipage}[t]{0.13\linewidth}
\includegraphics[width=1\linewidth]{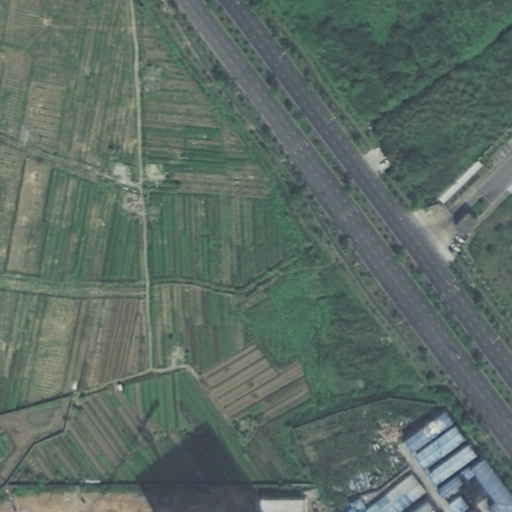}\vspace{2pt}
\includegraphics[width=1\linewidth]{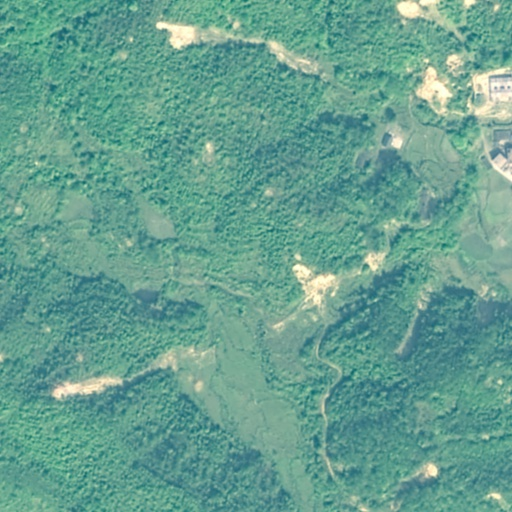}\vspace{2pt}
\includegraphics[width=1\linewidth]{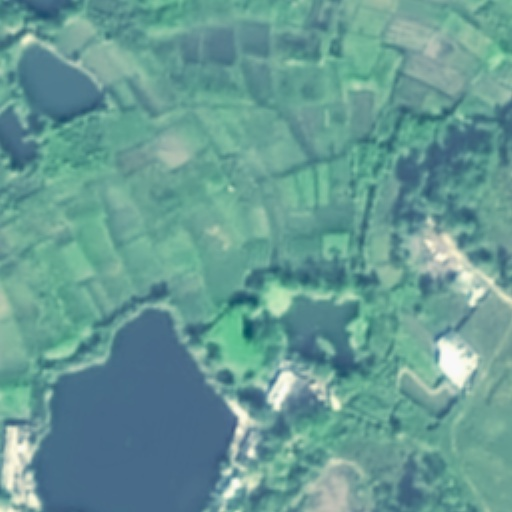}\vspace{2pt}
\includegraphics[width=1\linewidth]{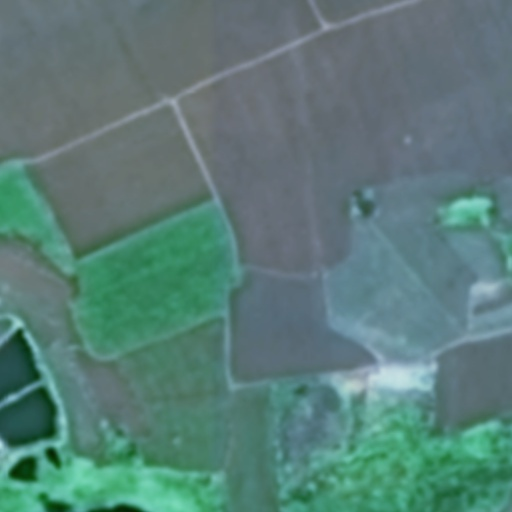}\vspace{2pt}
\end{minipage}}
\subfloat[$T_2$]{
\begin{minipage}[t]{0.13\linewidth}
\includegraphics[width=1\linewidth]{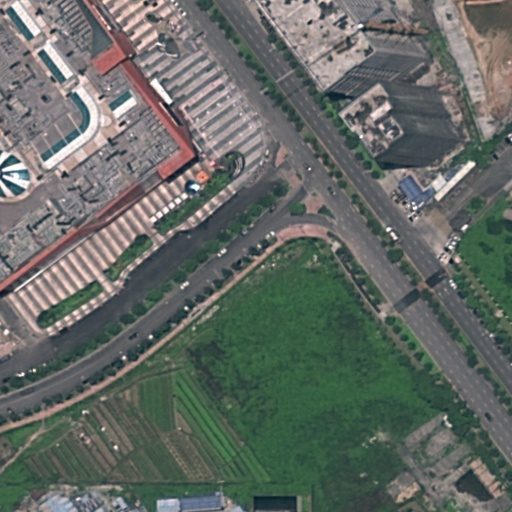}\vspace{2pt}
\includegraphics[width=1\linewidth]{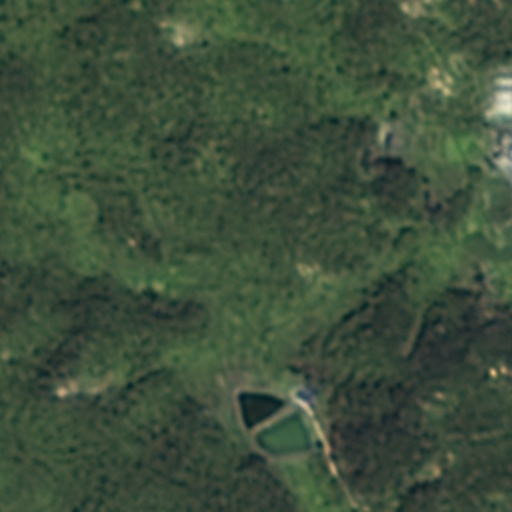}\vspace{2pt}
\includegraphics[width=1\linewidth]{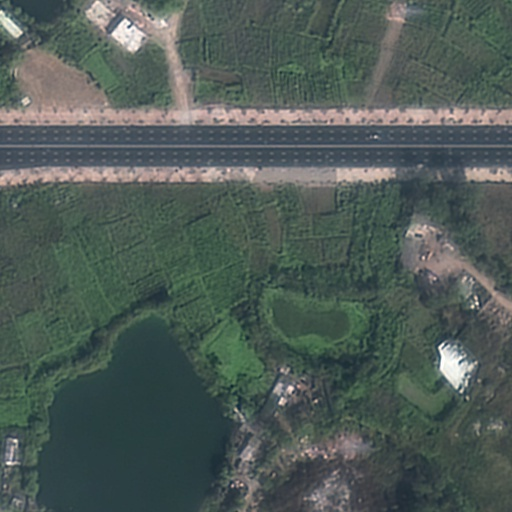}\vspace{2pt}
\includegraphics[width=1\linewidth]{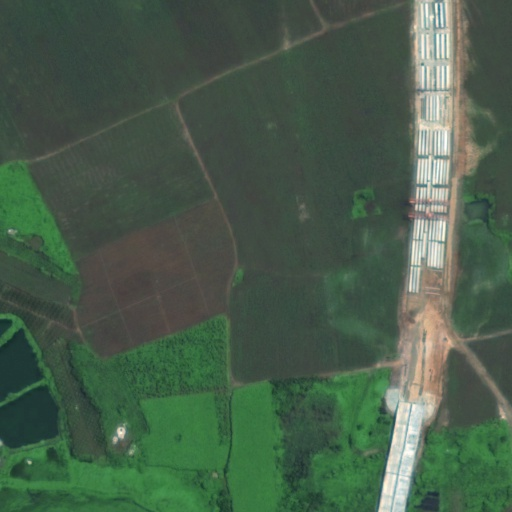}\vspace{2pt}
\end{minipage}}
\subfloat[GT]{
\begin{minipage}[t]{0.13\linewidth}
\includegraphics[width=1\linewidth]{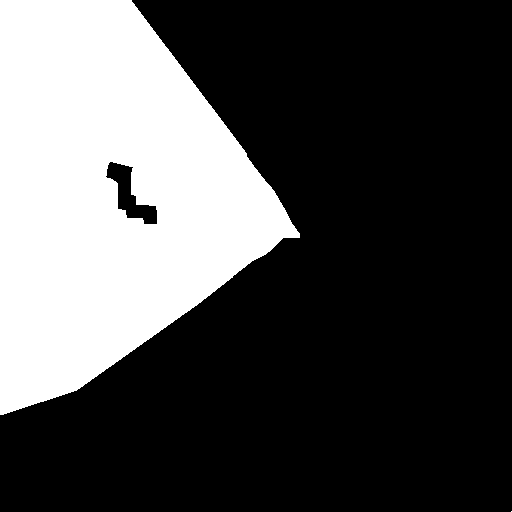}\vspace{2pt}
\includegraphics[width=1\linewidth]{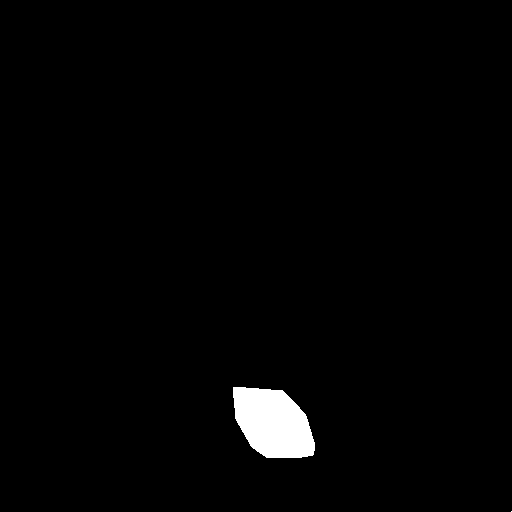}\vspace{2pt}
\includegraphics[width=1\linewidth]{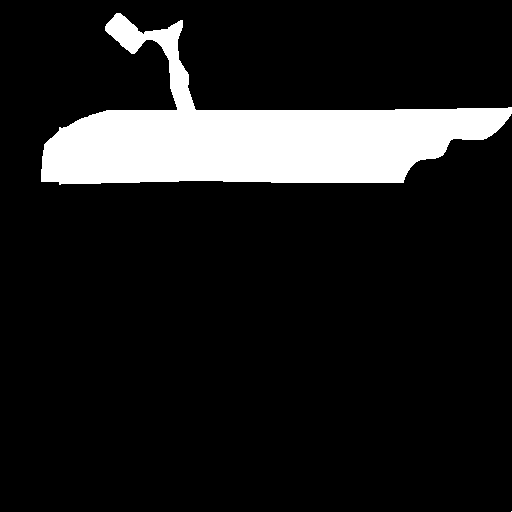}\vspace{2pt}
\includegraphics[width=1\linewidth]{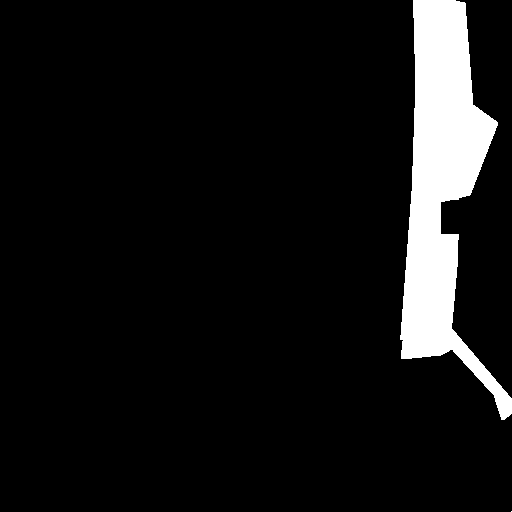}\vspace{2pt}
\end{minipage}}
\subfloat[$R_1$]{
\begin{minipage}[t]{0.13\linewidth}
\includegraphics[width=1\linewidth]{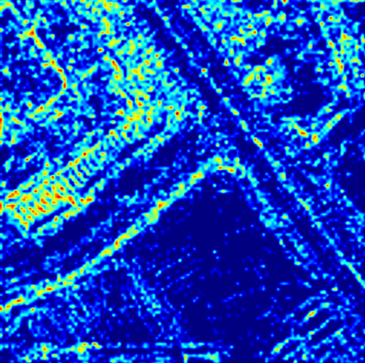}\vspace{2pt}
\includegraphics[width=1\linewidth]{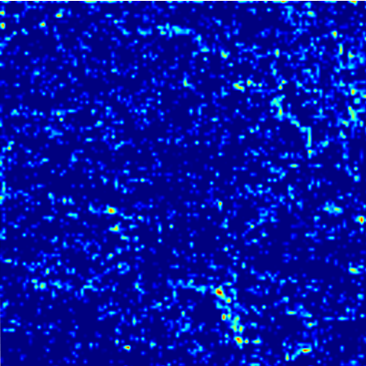}\vspace{2pt}
\includegraphics[width=1\linewidth]{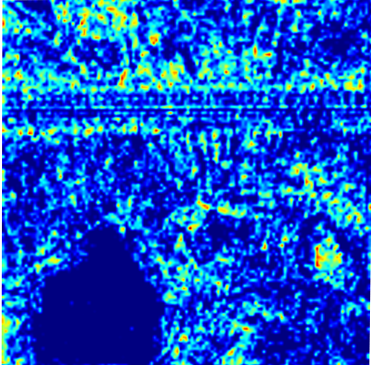}\vspace{2pt}
\includegraphics[width=1\linewidth]{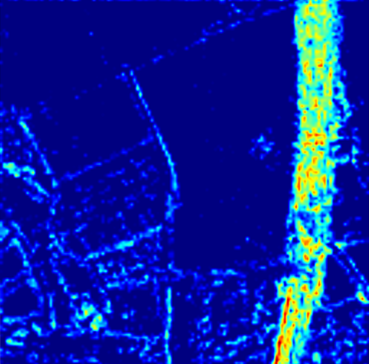}\vspace{2pt}
\end{minipage}}
\subfloat[$R_2$]{
\begin{minipage}[t]{0.13\linewidth}
\includegraphics[width=1\linewidth]{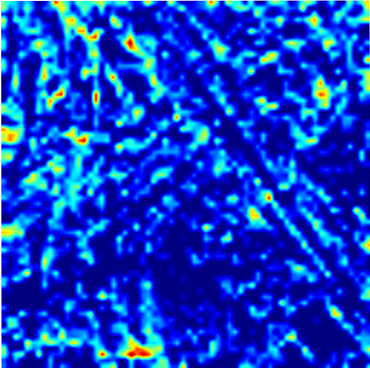}\vspace{2pt}
\includegraphics[width=1\linewidth]{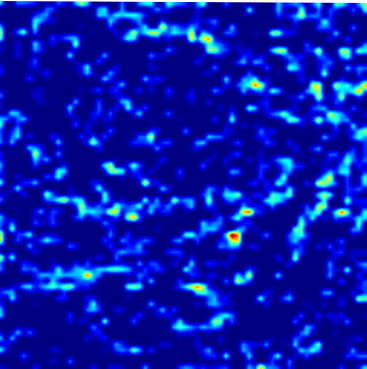}\vspace{2pt}
\includegraphics[width=0.98\linewidth]{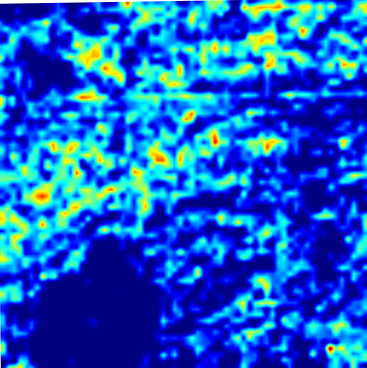}\vspace{2pt}
\includegraphics[width=0.98\linewidth]{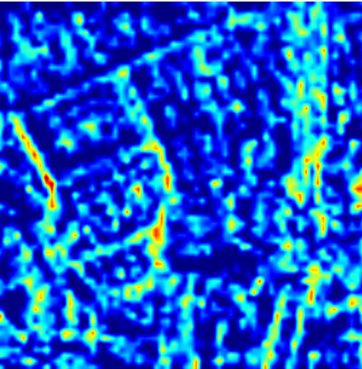}\vspace{2pt}
\end{minipage}}
\subfloat[$R_3$]{
\begin{minipage}[t]{0.13\linewidth}
\includegraphics[width=1\linewidth]{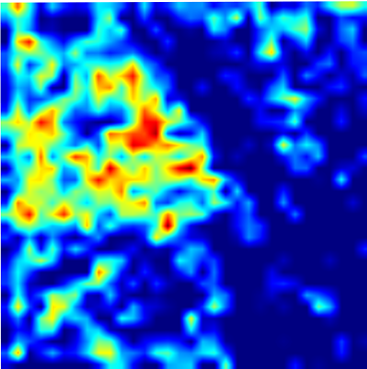}\vspace{2pt}
\includegraphics[width=1\linewidth]{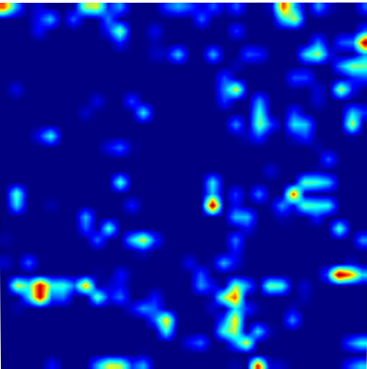}\vspace{2pt}
\includegraphics[width=0.98\linewidth]{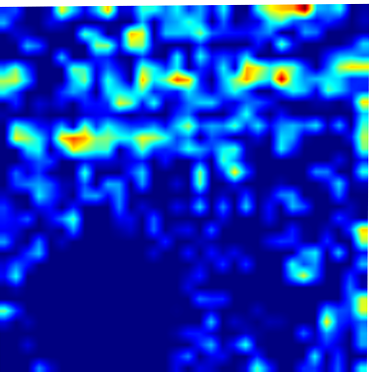}\vspace{2pt}
\includegraphics[width=0.98\linewidth]{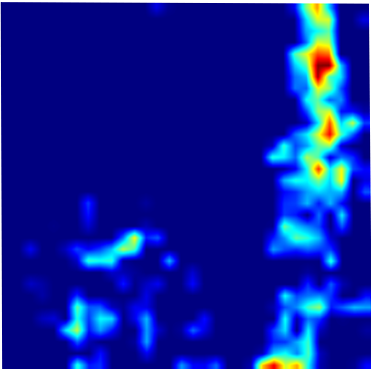}\vspace{2pt}
\end{minipage}}
\subfloat[$R_4$]{
\begin{minipage}[t]{0.13\linewidth}
\includegraphics[width=1\linewidth]{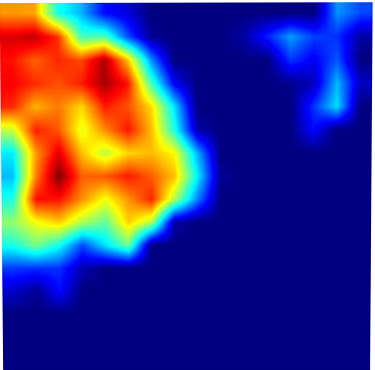}\vspace{2pt}
\includegraphics[width=1\linewidth]{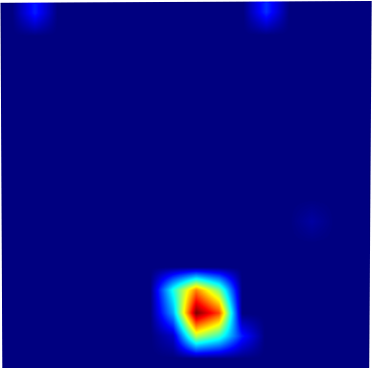}\vspace{2pt}
\includegraphics[width=1\linewidth]{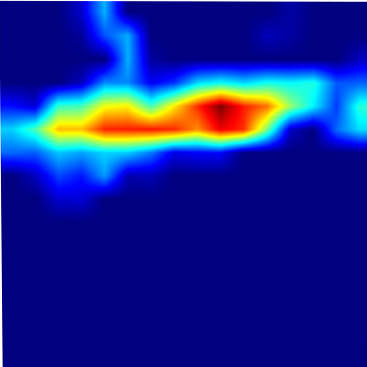}\vspace{2pt}
\includegraphics[width=1\linewidth]{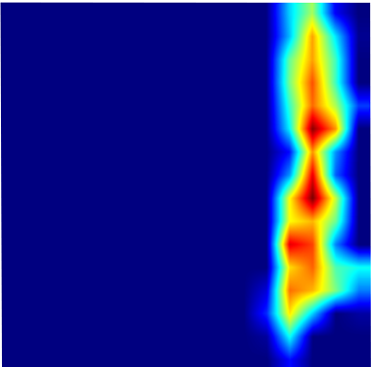}\vspace{2pt}
\end{minipage}}
\caption{Class activation maps for the change category of the features modulated by Cross-Temporal Locally Adaptive State-Space Scan (CT-LASS) module in different stages, which is based on grad-cam \cite{gradcam}. Example images are from the CLCD test set. $R_1$, $R_2$, $R_3$, and $R_4$ correspond to 1/4, 1/8, 1/16, and 1/32 resolution of the input image, respectively.}
\label{fig:gradcam_clcd}
\end{figure*}

To explore the validity of the model, we additionally visualize the activation maps of the features output by Siamese backbone of four stages modulated by the Cross-Temporal State-space Scan (CTSS) blocks on the CLCD dataset, which is implemented based on Grad-CAM. As shown in Fig. \ref{fig:gradcam_whucd} (WHU-CD), Fig. \ref{fig:gradcam_sysucd} (SYSU-CD), Fig. \ref{fig:gradcam_dsifn} (DSIFN-CD), and Fig. \ref{fig:gradcam_clcd} (CLCD), the activation values in the change region gradually increase as the depth of the model deepens, which proves that the CT-LASS module is effective in enhancing the semantic feature differences in the change region.

\subsection{Ablation Study}
A series of ablation experiments are conducted on the CLCD dataset to achieve the optimal model structure design, as follows.

\setlength{\tabcolsep}{6pt}
\begin{table}[t]
	\begin{center}
		\caption{
		Ablation experiments on the scan strategy on CLCD dataset.
		}
		\label{table:4}
            \begin{tabular}{l||ccccc}
		\Xhline{1.2pt}
            \rowcolor{mygray}
		     Scan strategy &F1 &Pre. &Rec. &IoU  & OA\\		
                \hline \hline
                VMamba & 76.46& 79.77 & 73.43& 61.90& 96.64\\
                LocalMamba  & 77.56& \bf83.37& 72.50& 63.34&\bf96.88\\
                \hline
                CT-LASS   & \bf78.06 &79.20& \bf76.96 &\bf64.02 &96.78                                     \\
   			\hline
		\end{tabular}
	\end{center}
\end{table}

\setlength{\tabcolsep}{6pt}
\begin{table}[t]
	\begin{center}
		\caption{
		Ablation experiments on the Cross-Temporal State-Space Scan (CTSS) strategy on CLCD dataset.
		}
		\label{table:7}
            \begin{tabular}{l||ccccc}
		\Xhline{1.2pt}
            \rowcolor{mygray}
		    \makecell{Bi-temporal \\ fusion strategy}  &F1 &Pre. &Rec. &IoU  & OA\\	
                \hline \hline
               CDS & 77.04& \bf83.61& 71.42&62.65 & \bf96.83\\
               RRS   & 76.73& 79.20& 74.41&62.24 & 96.64\\
               \hline
               CTSS & \bf78.06 &79.20& \bf76.96 &\bf64.02 &96.78     \\
   			\hline
		\end{tabular}
	\end{center}
\end{table}

\subsubsection{Ablation Study on the Scan Strategy} 
We conduct ablation analysis on the CLCD dataset to demonstrate that CT-LASS modules effectively enable global spatio-temporal context modeling while enhancing locality to further improve performance. Specifically, we retain the backbone and LCD of CD-Lambda and replace CT-LASS with SS2D in VMamba and LocalMamba for comparison. To adapt to the RSCD task. We also retain the CTSS cross-fusion strategy to integrate bi-temporal features. The variables are designed using two classical approaches: directly flattening for sequential scanning and constructing fixed windows for individual scanning. As shown in Table \ref{table:4}, CT-LASS achieves superior performance with an F1 score of 78.06, outperforming other variables.

\subsubsection{Ablation Study on the Component of Cross-Temporal Locally Adaptive State-Space Scan Module} 
We conduct preliminary experiments on certain variables in the key component of CD-Lamba, namely CT-LASS, to determine the most appropriate values.

First, we test the value of Top-$k$. This step aims to roughly identify regions with strong locality. Given a total of 16 windows, where regions with changes typically occupy a smaller proportion, we experiment with five different values for $k$: \{4, 5, 6, 7, 8\}. As shown in Table. \ref{table:5}, setting $k$ to 6 results in the highest F1 score of 78.06, indicating optimal model performance.

Next, we test the value of i in WSP (Window Shift Policy). We compare three configurations: (1) setting i to \{0\}, which excludes any shift strategy; (2) setting i to \{0, 1\}, which incorporates only the shift strategy from Swin Transformer; and (3) our proposed full configuration, where i is set to \{0, 1, 2, 3, 4\}, enabling windows to fully interact with neighboring windows. As shown in Table. \ref{table:6}, our full WSP configuration achieves the best model performance.

\subsubsection{Ablation Study on the Cross-Temporal State-Space Scan Strategy} 
\begin{figure}[t]
  \centering
  \includegraphics[width=1\linewidth]{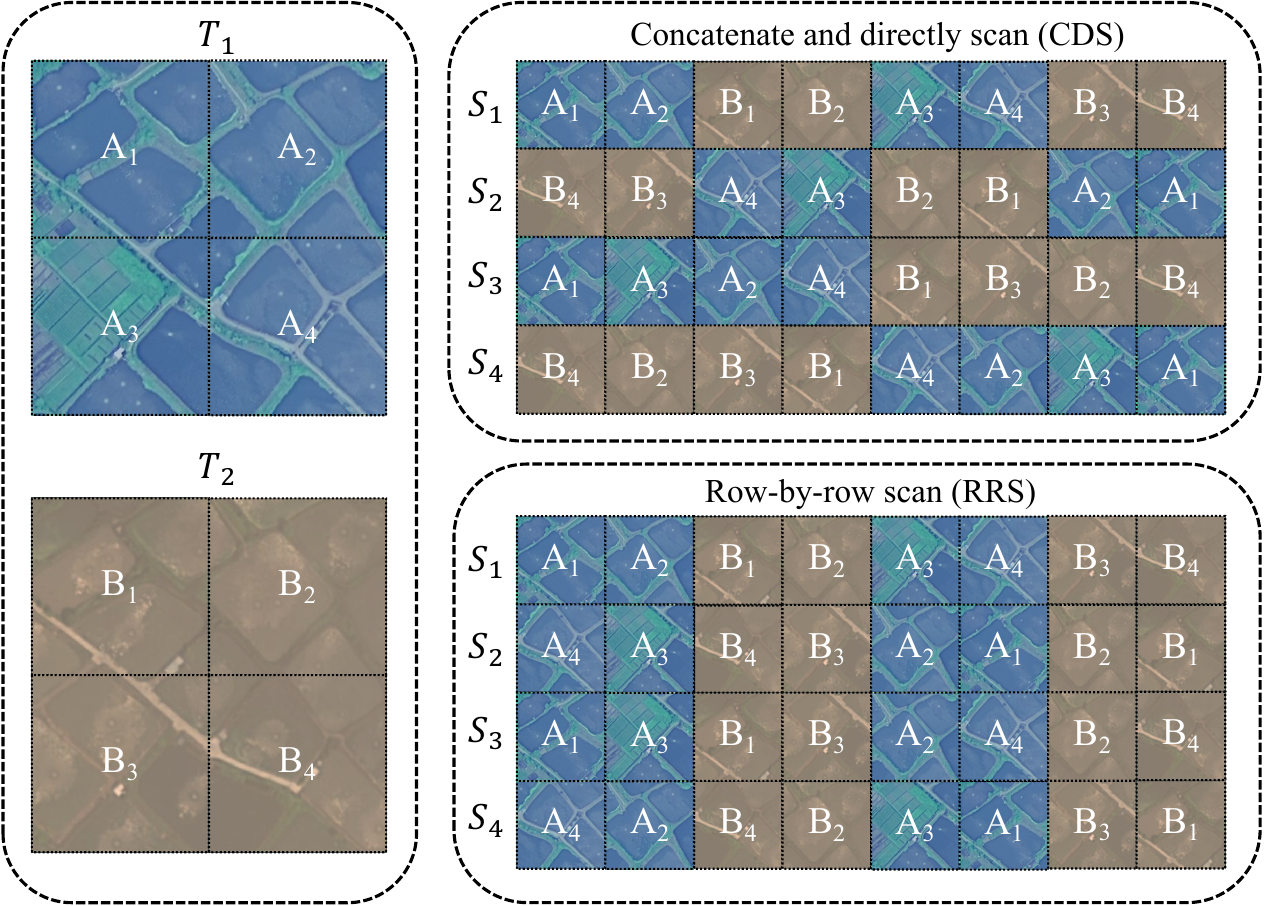}
  \caption{Different formations for bi-temporal selective scan strategy. Given bi-temporal images $T_1$ and $T_2$, we compare performance with two additional scan strategies: 1) Concatenate and directly scan (CDS); 2) Row-by-row scan (RRS).}
  \label{fig:ablation2}
\end{figure}

We explore the impact of different bi-temporal scan strategies on the model performance, as shown in Table \ref{table:7}. We first try to splice along the width dimension and then scan in all four directions, as described by CDS strategy in Fig. \ref{fig:ablation2}. This scheme by CDS achieved an F1 value of 77.04. Sequentially, another process, depicted by the RRS strategy in Fig. \ref{fig:ablation2}, involves scanning one row (or column) from image $\mathcal{T}_1$, then scanning one row (or column) from image $\mathcal{T}_2$, alternating between images $\mathcal{T}_1$ and $\mathcal{T}_2$ for the scanning process. This scheme by RRS achieved an F1 value of 76.73. Obviously, our pixel-by-pixel scan strategy has proven to be the most effective, yielding an F1 score of 78.06. 

\setlength{\tabcolsep}{6pt}
\begin{table}[t]
	\begin{center}
		\caption{
		Ablation experiments on the choice of Top-$k$ on CLCD dataset.
		}
		\label{table:5}
            \begin{tabular}{l||ccccc}
		\Xhline{1.2pt}
            \rowcolor{mygray}
		     Top-$k$ &F1 &Pre. &Rec. &IoU  & OA\\		
                \hline \hline
                Top-4 & 77.13 & 82.04 &  72.77& 62.77& 96.79 \\
                Top-5  &77.11 & 80.61& 73.89& 62.74& 96.74\\
                Top-6 & \bf78.06 &79.20& \bf76.96 &\bf64.02 &96.78     \\
                Top-7  & 77.16& 82.20& 72.71& 62.82&\bf96.80 \\
                Top-8 & 76.62& \bf82.36& 71.63& 62.10&96.75  \\
   			\hline
		\end{tabular}
	\end{center}
\end{table}

\setlength{\tabcolsep}{6pt}
\begin{table}[t]
	\begin{center}
		\caption{
		Ablation experiments on the Window Shifting and Perception (WSP) mechanism on CLCD dataset.
		}
		\label{table:6}
            \begin{tabular}{l||ccccc}
		\Xhline{1.2pt}
            \rowcolor{mygray}
		     Values of $i$ &F1 &Pre. &Rec. &IoU  & OA\\	
                \hline \hline
               \{0\} & 76.82 &\bf81.18 & 72.91& 62.37& 96.73\\
               \{0,1\}   & 77.31& 80.53& 74.33& 63.01& 96.75\\
               \{0,1,2,3,4\} & \bf78.06 &79.20& \bf76.96 &\bf64.02 &\bf96.78     \\
   			\hline
		\end{tabular}
	\end{center}
\end{table}

\section{Conclusion}
Mamba has recently brought a new perspective to the field of RSCD with its powerful global perception capabilities and linear computational complexity. However, current SSM-based RSCD methods, despite altering scanning directions and introducing cross-temporal strategies, commonly adopt a straightforward approach of flattening images into sequences before applying Mamba. This practice significantly disrupts the inherent locality of the change regions. These limitations hinder the performance improvements of existing RSCD methods.
In this paper, we propose CD-Lamba, which efficiently models spatio-temporal context with enhanced locality. Specifically, the Locally Adaptive State-Space Scan strategy is designed to enhance locality while maintaining global perception, the Cross-Temporal State-Space Scan strategy facilitates bi-temporal feature fusion, and the Window Shifting and Perception mechanism improves interactions across segmented windows. These strategies are integrated into a multi-scale Cross-Temporal Locally Adaptive State-Space Scan module, which effectively highlights changes and refines features.
The proposed CD-Lamba achieves state-of-the-art performance on four RSCD datasets, striking a better balance between accuracy and efficiency. However, CD-Lamba currently has a gap in distinguishing between actual changes of interest and pseudo changes when obtaining locality windows. It still tends to classify some regions with a high likelihood of change as locality windows. In future work, we plan to introduce appropriate state transfer mechanisms and knowledge distillation for autonomous window learning, as we believe this will further advance the development of SSMs for RSCD tasks.

\section{Acknowledgements}
This work is supported by the Public Welfare Science and Technology Plan of Ningbo City (2022S125) and the Key Research and Development Plan of Zhejiang Province (2021C01031).
\bibliographystyle{elsarticle-num}

\bibliography{main.bib}



\end{document}